\documentclass[10pt,twocolumn,letterpaper]{article}

\usepackage[pagenumbers]{cvpr} %

\usepackage{graphicx}
\usepackage{amsmath}
\usepackage{amssymb}
\usepackage{booktabs}

\usepackage{algorithm}
\usepackage{algpseudocode}

\usepackage{multirow}

\usepackage{soul}
\usepackage[percent]{overpic}

\usepackage{paralist}

\usepackage{xcolor}
\usepackage{colortbl}

\definecolor{yellow}{rgb}{1,1, 0.6}
\definecolor{lightyellow}{rgb}{1,1, 0.8}
\definecolor{orange}{rgb}{1, 0.8, 0.6}
\definecolor{tabred}{rgb}{1, 0.6, 0.6}

\usepackage[pagebackref,breaklinks,colorlinks]{hyperref}

\usepackage[capitalize]{cleveref}
\crefname{section}{Sec.}{Secs.}
\Crefname{section}{Section}{Sections}
\Crefname{table}{Table}{Tables}
\crefname{table}{Tab.}{Tabs.}

\def\cvprPaperID{4307} %
\def\confName{CVPR}
\def\confYear{2022}

\newcommand{\sg}{\mathrm{sg}}

\begin{document}

\title{NeRF in the Dark: High Dynamic Range View Synthesis from Noisy Raw Images}

\author{
\hspace{-5pt}
Ben Mildenhall
\;\,
Peter Hedman
\;\,
Ricardo Martin-Brualla
\;\,
Pratul P. Srinivasan
\;\,
Jonathan T. Barron
\\
\vspace{2mm}
{Google Research}
}
\maketitle

\begin{abstract}

Neural Radiance Fields (NeRF) is a technique for high quality novel view synthesis from a collection of posed input images. Like most view synthesis methods, NeRF uses tonemapped low dynamic range (LDR) as input; these images have been processed by a lossy camera pipeline that smooths detail, clips highlights, and distorts the simple noise distribution of raw sensor data. We modify NeRF to instead train directly on linear raw images, preserving the scene's full dynamic range. By rendering raw output images from the resulting NeRF, we can perform novel high dynamic range (HDR) view synthesis tasks. In addition to changing the camera viewpoint, we can manipulate focus, exposure, and tonemapping after the fact. Although a single raw image appears significantly more noisy than a postprocessed one, we show that NeRF is highly robust to the zero-mean distribution of raw noise. When optimized over many noisy raw inputs (25-200), NeRF produces a scene representation so accurate that its rendered novel views outperform dedicated single and multi-image deep raw denoisers run on the same wide baseline input images. As a result, our method, which we call \emph{RawNeRF}, can reconstruct scenes from extremely noisy images captured in near-darkness.

\end{abstract}

\section{Introduction}
\label{sec:intro}

View synthesis methods, such as neural radiance fields (NeRF)~\cite{mildenhall2020nerf}, typically use tonemapped low dynamic range (LDR) images as input and directly reconstruct and render new views of a scene in LDR space. This poses no issues for scenes that are well-lit and do not contain large brightness variations, since they can be captured with minimal noise using a single fixed camera exposure setting. However, this precludes many common capture scenarios: images taken at nighttime or in any but the brightest indoor spaces will have poor signal-to-noise ratios, and scenes with regions of both daylight and shadow have extreme contrast ratios that require high dynamic range (HDR) to represent accurately.

\begin{figure}[]
    \centering
    \includegraphics[width=\columnwidth]{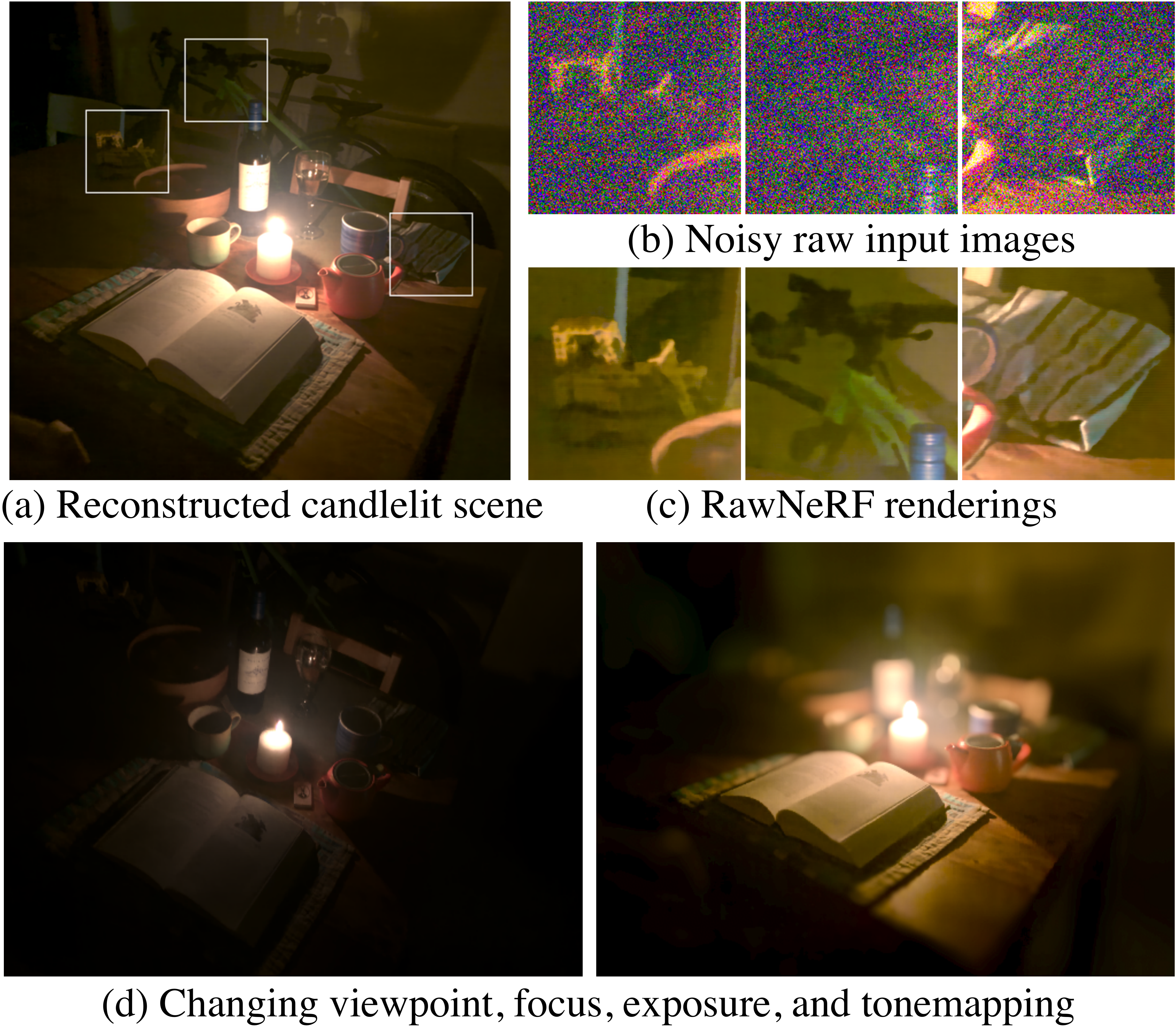}  
    \caption{By jointly optimizing a single scene representation over many input images, NeRF is surprisingly robust to high levels of image noise. We exploit this fact to train RawNeRF directly on completely unprocessed HDR linear raw images. In this nighttime scene lit only by a single candle (a), RawNeRF can extract details from the noisy raw data that would have been destroyed by postprocessing (b, c). RawNeRF recovers full HDR color information, enabling HDR view synthesis tasks such as changing focus and exposure for rendered novel views. The resulting renderings can be retouched like any raw photograph: here we show (d, left) a dark all-in-focus exposure with a simple global tonemap and (d, right) a brighter, synthetically refocused exposure postprocessed by HDRNet~\cite{hdrnet}.
    See our supplementary video for more results.
    }
    \label{fig:teaser}
\end{figure}

\begin{figure*}
    \centering
    \includegraphics[width=\textwidth]{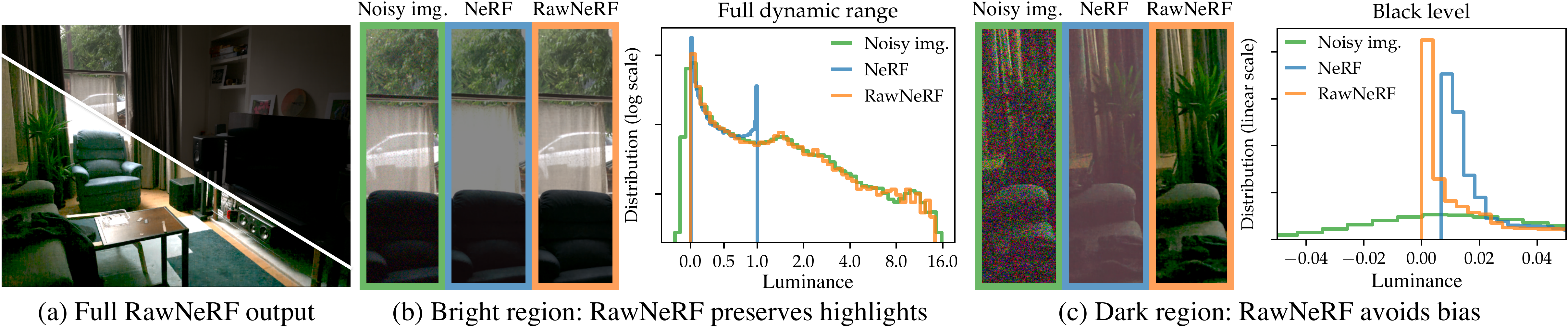}
    \caption{
    Failure modes of NeRF on a daytime indoor scene.
    (a) Here we show two exposures ($24\times$ apart) of a full RawNeRF output rendering, both passed through a global tonemapping curve.
    Training NeRF with postprocessed LDR images, as done in prior work, (b) prevents it from recovering bright highlights clipped above at 1, resulting in the missing car outside the window, and (c) corrupts the per-pixel noise distribution such that NeRF recovers incorrect colors due to the nonlinear tonemap and clipping below at 0, particularly in dark regions around the plant and sofa. In contrast, RawNeRF trains directly on HDR linear raw images and correctly recovers the radiance distribution in both extremely bright and extremely dark parts of the scene.
    }
    \label{fig:raw_vs_srgb}
\end{figure*}

Our method, RawNeRF, modifies NeRF to reconstruct the scene in linear HDR color space by supervising directly on noisy raw input images.
This bypasses the lossy postprocessing that cameras apply to compress dynamic range and smooth out noise in order to produce visually palatable 8-bit JPEGs.
By preserving the full dynamic range of the raw inputs, RawNeRF enables various novel HDR view synthesis tasks. We can modify the exposure level and tonemapping algorithm applied to rendered outputs and even create synthetically refocused images with accurately rendered bokeh effects around out-of-focus light sources. 

Beyond these view synthesis applications, we show that training directly on raw data effectively turns RawNeRF into
a multi-image denoiser capable of reconstructing scenes captured in near-darkness (Figure~\ref{fig:teaser}). 
The standard camera postprocessing pipeline (\eg, HDR+~\cite{hdrplus}) corrupts the simple noise distribution of raw data, introducing significant bias in order to reduce variance and produce an acceptable output image. 
Feeding these images into NeRF thus produces a biased reconstruction with incorrect colors, particularly in the darkest regions of the scene (see Figure~\ref{fig:raw_vs_srgb} for an example). 
We instead exploit NeRF's ability to reduce variance by aggregating information across frames, demonstrating that it is possible for RawNeRF to produce a clean reconstruction from many noisy raw inputs. 

Unlike typical video or burst image denoising methods, RawNeRF assumes a static scene and expects camera poses as input. Provided with these extra constraints, RawNeRF is able to make use of 3D multiview consistency to average information across nearly \emph{all} of the input frames at once. Since our captured scenes each contain 25-200 input images, this means RawNeRF can remove more noise than feed-forward single or multi-image denoising networks that only make use of 1-5 input images for each output. 

In summary, we make the following contributions:
\begin{compactenum}
    \item We propose a method for training RawNeRF directly on raw images
    that can handle high dynamic range scenes as well as noisy inputs captured in the dark.
    \item We show that RawNeRF outperforms NeRF on noisy real and synthetic datasets and is a competitive multi-image denoiser for wide-baseline static scenes.
    \item We showcase novel view synthesis applications made possible by our linear HDR scene representation (varying exposure, tonemapping, and focus).
\end{compactenum}

\section{Related Work}
\label{sec:relatedwork}

RawNeRF combines concepts from several areas of research. We build upon NeRF as a baseline for high quality view synthesis, bring in ideas from low level image processing to optimize NeRF directly on noisy raw data, and take inspiration from uses of HDR in computer graphics and computational photography to showcase new applications made possible by an HDR scene reconstruction. We briefly cover relevant prior work across each of these areas.

\subsection{Novel view synthesis}

Novel view synthesis is the task of using a set of input images and their camera poses to reconstruct a scene representation capable of rendering novel views. When the input images are densely sampled, it is possible to use direct interpolation in pixel space for view synthesis~\cite{levoy96lightfields,cohen96lumigraph}. A more feasible capture scenario is to capture more widely spaced inputs and use a ``proxy'' geometry (\eg, a reconstructed triangle mesh) to reproject and combine colors from the input images, using either a heuristic~\cite{buehler01unstructlumigraph} or learned~\cite{hedman2018deepblending,riegler2020fvs,riegler2021svs} blending function.

Recent work on applying deep learning to view synthesis has focused on volumetric rather than mesh-based scene representations~\cite{flynn2016deepstereo,zhou18stereomag,lombardi2019neuralvolumes}. 
NeRF~\cite{mildenhall2020nerf} directly optimizes a \emph{neural} volumetric scene representation to match all input images using gradient descent on a rendering loss. 
Various extensions have improved NeRF's robustness to varying lighting conditions~\cite{martinbrualla2020nerfw} or added supervision with  depth~\cite{wei2021nerfingmvs,jeong2021scnerf,kangle2021dsnerf}, time-of-flight data~\cite{attal2021torf}, or semantic segmentation labels~\cite{zhi2021semanticnerf}. As of yet, no approach has extended NeRF to work with high dynamic range color data.
Some previous view synthesis methods trained using LDR data jointly solve for per-image scaling factors to account for inconsistent lighting or miscalibration between cameras~\cite{lombardi2019neuralvolumes,kopanas2021pbr}.
ADOP~\cite{ruckert2021adop} supervises with LDR images and solves for exposure through a differentiable tonemapping step to approximately recover HDR, but does not focus on robustness to noise or supervision with raw data.

\subsection{Denoising}

Early neural denoising approaches mostly focused on denoising sRGB images synthetically corrupted with additive white Gaussian noise~\cite{zhang2017beyond}. In 2017, Pl\"otz and Roth~\cite{plotz2017cvpr} established a real raw image denoising benchmark, which showed that these deep denoisers failed to generalize beyond the synthetic data used during training and were outperformed by standard non-learned methods, such as BM3D~\cite{bm3d}. Subsequent work on both single~\cite{chen2018cvpr,brooks2019cvpr} and multi-image~\cite{kpn,godard2018burst,chen2019iccv,rvidenet} denoising demonstrated the benefits of training networks to operate directly on noisy raw input data. Modern cellphone camera pipelines perform a robust averaging of multiple noisy input frames in the raw domain~\cite{hdrplus}, though they typically cannot afford to employ deep networks due to speed and power limitations. 

Another line of research investigated whether denoisers could be trained using \emph{only} noisy data when no corresponding clean ground truth exists. Noise2Noise~\cite{lehtinen2018} demonstrated this was possible given a dataset of pairs of independent noisy observations of the same image, an insight Ehret \etal~\cite{ehret2019f2f} applied to denoise videos by aligning consecutive noisy frames.
Various followups to Noise2Noise proposed modified network architectures allowing supervision with a dataset of single noisy images~\cite{krull2019noise2void,batson2019noise2self,laine2019blindspot}. 
Sheth \etal~\cite{udvd} showed that this paradigm could be applied to train a denoiser using a \emph{single} noisy video, including an application to raw video data. Similarly, RawNeRF is optimized over a single set of images to both denoise and recover the 3D structure of the captured scene.

\subsection{Applications of raw and HDR image data}

\paragraph{Computational photography}
The value of working directly with raw data has long been noted by digital photographers due to the fact that its preservation of dynamic range allows for maximum postprocessing flexibility, letting users modify exposure, white balance, and tonemapping after the fact. Many works have tried to automate this process by using heuristics or machine learning to map directly from raw data to postprocessed LDR images~\cite{mit5k,chen2018cvpr,hdrnet,hu2018whitebox}. 

Another line of work focuses on recovering HDR images from LDR inputs. This concept was pioneered by Debevec and Malik~\cite{debevec1997}, who used a stack of aligned LDR images taken at different exposures to recover and invert the camera's nonlinear response curve. 
Current approaches apply machine learning to produce HDR outputs from single~\cite{ldr2hdrcnn} or multiple misaligned~\cite{kalantari2017hdr} LDR inputs, either recovering or hallucinating detail in clipped highlights.

\paragraph{Synthetic defocus} Many modern cellphones include a postprocessing option to add synthetic defocus blur after capture~\cite{wadhwa2018defocus}. Though it is possible to accurately simulate defocus using a thin-lens model~\cite{cook1984distributed} or real multi-element camera lens~\cite{lenstracing} using ray tracing, most machine learning models use a much faster approximate rendering model, predicting a depth map and applying a depth-varying blur kernel to each discretized depth layer~\cite{barron2015stereo,srinivasan2018aperture}. Performing this blur in HDR space is critical to achieving the correct appearance of defocused bright highlights (known as ``bokeh''), as demonstrated by Zhang \etal~\cite{zhang2019defocus}.

\section{Noisy Raw Input Data}
\label{sec:isp}

NeRF~\cite{mildenhall2020nerf} takes postprocessed low dynamic range (LDR) sRGB color space images as input. This works well when using clean, noise-free images with minimal constrast. However, all real images contain some level of noise, and each step in the camera postprocessing pipeline corrupts this distribution in a certain way.
Here we briefly describe the simplified pipeline stages relevant to our method.

\paragraph{Raw camera measurements}

When capturing an image, the number of photons hitting a pixel on the camera sensor is converted to an electrical charge, which is recorded as a high bit-depth digital signal (typically 10 to 14 bits). These values are offset by a ``black level'' to allow for negative measurements due to noise. After black level subtraction, the signal is a noisy measurement $y_i$ of a quantity $x_i$ proportional to the expected number of photons arriving while the shutter is open. This noise results from both the physical fact that photon arrivals are a Poisson process (``shot'' noise) and noise in the readout circuitry that converts the analog electrical signal to a digital value (``read'' noise). The combined shot and read noise distribution can be well modeled as a Gaussian whose variance is an affine function of its mean~\cite{foi2008noisemodel}; importantly, this implies that the distribution of the error $y_i - x_i$ is zero mean.

\paragraph{Color filter demosaicking} Color cameras contain a Bayer color filter array in front of the image sensor such that each pixel's spectral response curve measures either red, green or blue light. The pixel color values are typically arranged in $2\times 2$ squares containing two green pixels, one red, and one blue pixel (known as a Bayer pattern), resulting in ``mosaicked'' data. To generate a full-resolution color image, the missing color channels are interpolated using a demosaicking algorithm~\cite{li2008demosaic}. 
This interpolation correlates noise spatially, and the checkerboard pattern of the mosaic leads to different noise levels in alternating pixels.

\paragraph{Color correction and white balance} 
The spectral response curves for each color filter element vary between different cameras, and a color correction matrix is used to convert the image from this camera-specific color space to a standardized color space. Additionally, because human perception is robust to the color tint imparted by different light sources, cameras attempt to account for this tint (\ie, make white surfaces appear RGB-neutral white) by scaling each color channel by an estimated white balance coefficient. These two steps are typically combined into a single linear $3\times 3$ matrix transform, which further correlates the noise between color channels.

\paragraph{Gamma compression and tonemapping}

Humans are able to discern smaller relative differences in dark regions compared to bright regions of an image. 
This fact is exploited by sRGB gamma compression, which optimizes the final image encoding by clipping values outside $[0,1]$ and applying a nonlinear curve to the signal that dedicates more bits to dark regions at the cost of compressing bright highlights.
In addition to gamma compression, tonemapping algorithms can be used to better preserve contrast in high dynamic range scenes (where the bright regions are several orders of magnitude brighter than the darkest) when the image is quantized to 8 bits~\cite{debevec1997,hdrplus}.

In a slight abuse of terminology, we will refer both of these steps jointly as ``tonemapping'' in the rest of the paper, indicating the process by which linear HDR values are mapped to nonlinear LDR space for visualization. We will refer to signals before tonemapping as high dynamic range (HDR) and signals after as low dynamic range (LDR).
Of all postprocessing operations, tonemapping has the most drastic effect on the noise distribution: clipping completely discards information in the brightest and darkest regions, and after the non-linear tonemapping curve the noise is no longer guaranteed to be Gaussian or even zero mean.

\section{RawNeRF}
\label{sec:method}

A neural radiance field (NeRF)~\cite{mildenhall2020nerf} is a neural network based scene representation that is optimized to reproduce the appearance of a set of input images with known camera poses. The resulting reconstruction can then be used to render novel views from previously unobserved poses. NeRF's multilayer perceptron (MLP) network takes 3D position and 2D viewing direction as input and outputs volume density and color. To render each pixel in an output image, NeRF uses volume rendering to combine the colors and densities from many points sampled along the corresponding 3D ray.

Standard NeRF takes clean, low dynamic range (LDR) sRGB color space images with values in the range $[0, 1]$ as input. Converting raw HDR images to LDR images (\eg, using the pipeline described in Section~\ref{sec:isp}) has two significant consequences:
\begin{enumerate}
    \item Detail in bright areas is lost when values are clipped from above at one, and detail across the image is compressed by the tonemapping curve and subsequent quantization to 8 bits.
    \item The per-pixel noise distribution becomes biased (no longer zero-mean) after passing through a nonlinear tonemapping curve and being clipped from below at zero.
\end{enumerate}

\begin{figure}
    \centering
    \includegraphics[width=\columnwidth]{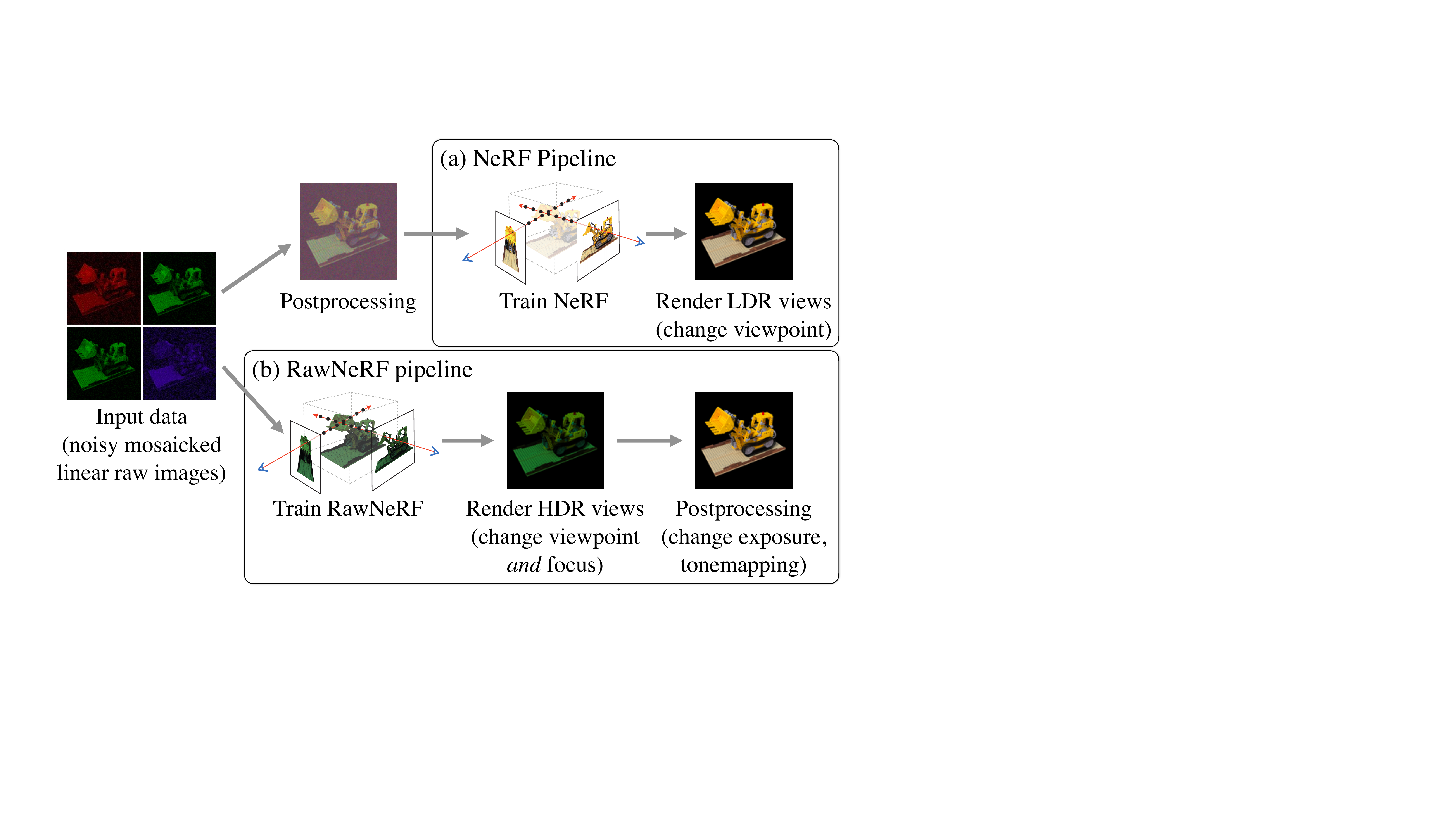}
    \caption{
    The standard NeRF training pipeline (a) takes in LDR images that have been sent through a camera processing pipeline, reconstructing the scene and rendering new views in LDR color space. As such, its renderings are effectively already postprocessed and cannot be significantly retouched. In contrast, our method RawNeRF (b) modifies NeRF to train directly on linear raw HDR input data. The resulting scene representation produces novel views that can be edited like any raw photograph.
    }
    \label{fig:pipeline}
\end{figure}

The goal of RawNeRF is to make use of this information rather than discarding it, optimizing NeRF directly on linear raw input data in HDR color space (Figure~\ref{fig:pipeline}).
In Section~\ref{sec:results}, we will show that reconstructing NeRF in raw space makes it much more robust to noisy inputs and allows for novel HDR view synthesis applications. First, we detail the changes required to make NeRF work with raw data.

\subsection{Loss function}

\begin{figure}
    \centering
    \includegraphics[width=\columnwidth]{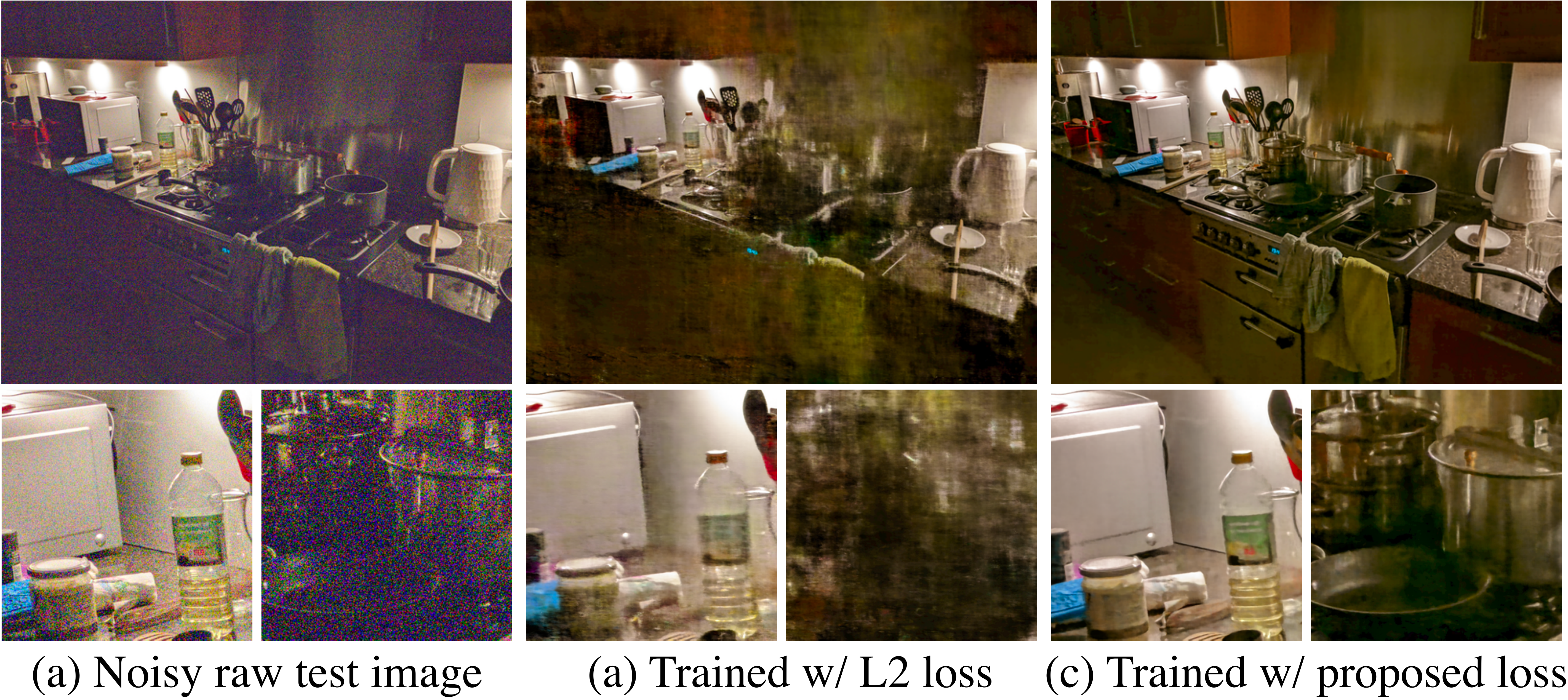}
    \caption{
    This challenging scene (a) has a $7000\times$ ratio between its $90^{\textrm{th}}$ and $10^{\textrm{th}}$ raw color percentiles.
    (b) When faced with such high-contrast inputs, the standard L2 loss from NeRF manages to recover the bright parts of the scene but produces poor results in darker regions, which becomes particularly apparent after LDR tonemapping. (c) Our proposed loss (\ref{eq:loss}), reweighted according to the gradient of a log tonemap curve, successfully reconstructs all parts of the scene. 
    Both rendered images are tonemapped using HDR+~\cite{hdrplus} for visualization.
    }
    \label{fig:l2loss}
\end{figure}

Since the color distribution in an HDR image can span many orders of magnitude, a standard L2 loss applied in HDR space will be completely dominated by error in bright areas and produce an image that has muddy dark regions with low contrast when tonemapped (see  Figure~\ref{fig:l2loss}).
Instead, we apply a loss that more strongly penalizes errors in dark regions to align with how human perception compresses dynamic range. One way to achieve this is by passing both the rendered estimate $\hat y$ and noisy observed intensity $y$ through a tonemapping curve $\psi$ before the loss is applied:
\begin{align}
    L_\psi(\hat y, y) = \sum_i (\psi(\hat y_i) - \psi(y_i))^2\, .
    \label{eq:tonemaploss}
\end{align}
However, in low-light raw images the observed signal $y$ is heavily corrupted by zero-mean noise, and a nonlinear tonemap will introduce bias that changes the noisy signal's expected value ($E[\psi(y)] \neq \psi(E[y])$). In order 
for the network to converge to an unbiased result~\cite{lehtinen2018}, 
we instead use a weighted L2 loss of the form
\begin{align}
    L(\hat y, y) = \sum_i w_i (\hat y_i - y_i)^2\,  .
\end{align}
We can approximate the tonemapped loss (\ref{eq:tonemaploss}) in this form by using a linearization of the tone curve $\psi$ around each $\hat y_i$:
\begin{align}
   \tilde L_{\psi}(\hat y, y) = &\sum_i \left[\psi'(\sg(\hat y_i))(\hat y_i - y_i) \right]^2 \, ,
\end{align}
where $\sg(\cdot )$ indicates a stop-gradient that treats its argument as an constant with zero derivative, preventing it from influencing the loss gradient during backpropagation.
We find that a ``gradient supervision'' tone curve $\psi(z) = \log(y + \epsilon)$ with $\epsilon=10^{-3}$ produces perceptually high quality results with minimal artifacts, implying a loss weighting term of $\psi'(\sg(\hat y_i)) = (\sg(\hat y_i) + \epsilon )^{-1}$ and final loss
\begin{align}
\label{eq:loss}
    \tilde L_{\psi}(\hat y, y) = &\sum_i \left(\frac{\hat y_i - y_i}{\sg(\hat y_i) + \epsilon} \right)^2 .
\end{align}
This corresponds exactly to the relative MSE loss used to achieve unbiased results when training on noisy HDR pathtracing data in Noise2Noise~\cite{lehtinen2018}. The curve $\psi$ is proportional to the $\mu$-law function used for range compression in audio processing, and has previously been applied as a tonemapping function when supervising a network to map from a burst of LDR images to an HDR output~\cite{kalantari2017hdr}.

\subsection{Variable exposure training}

\begin{figure}[]
    \centering
    \includegraphics[width=\columnwidth]{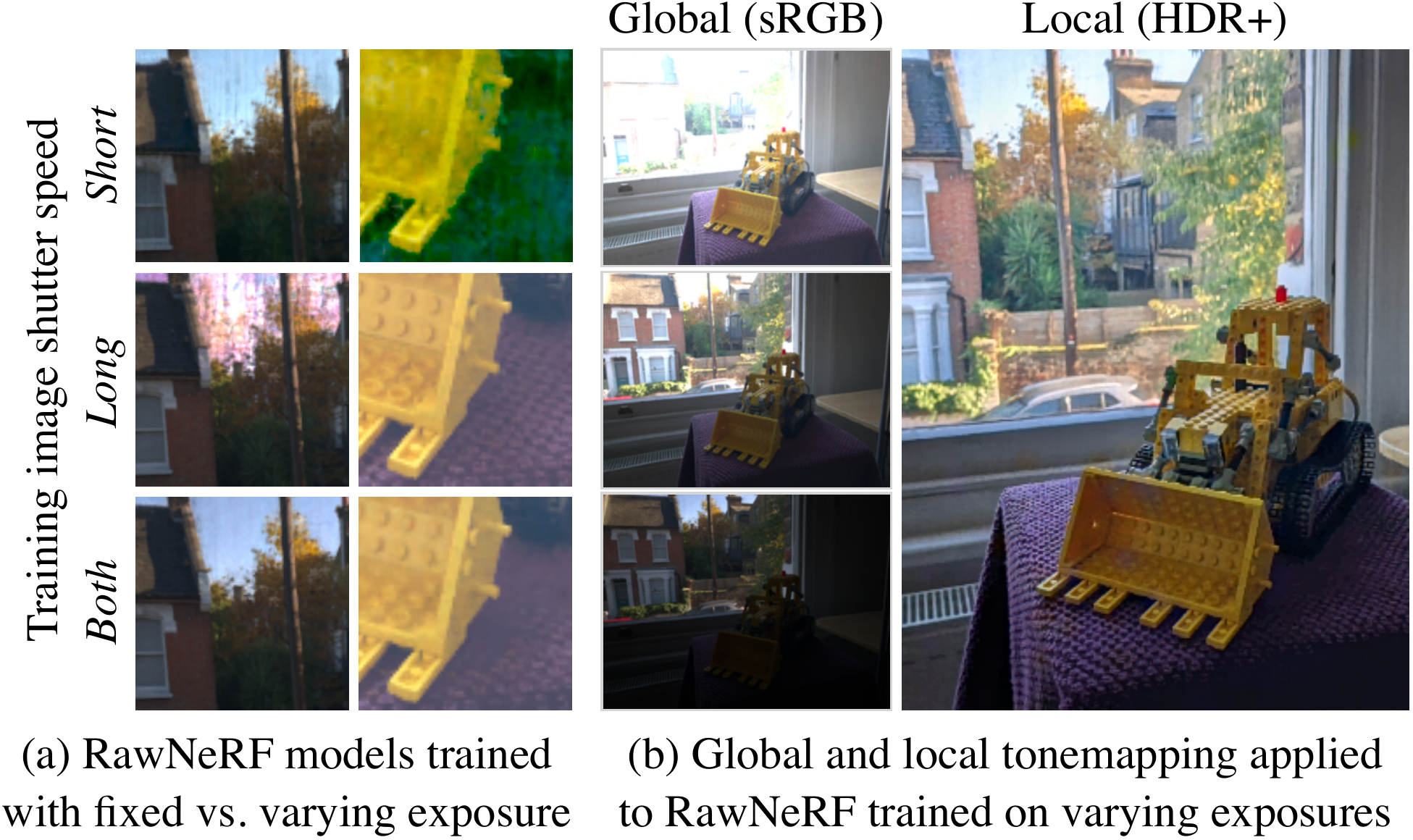}  
    \caption{A fixed shutter speed is not sufficient for capturing the full dynamic range in scenes with extreme brightness variation. (a) For example, this scene requires variable exposure capture to avoid either poor quality in dark indoor regions or blown-out sky highlights. Only a RawNeRF model optimized using both short and long exposures recovers the full dynamic range. (b) This brightness variation is too high to visualize in a single image using a simple global sRGB gamma curve, requiring a more sophisticated local tonemapping algorithm (\eg, HDR+ postprocessing~\cite{hdrplus}).
    }
    \label{fig:shortlong}
\end{figure}

\newcommand{\realinsetwidth}{.13\linewidth}
\newcommand{\realfullwidth}{.17333\linewidth}
\newcommand{\raisedrule}[2][0pt]{\leaders\hbox{\rule[#1]{1pt}{#2}}\hfill}
\begin{figure*}[]
    \centering
\resizebox{\linewidth}{!}{
    \begin{tabular}{c@{\,}c@{\,}c@{\,}c@{\,}c@{\,}c@{\,}c@{\,}c@{\,}c@{\,}c@{\,}}
    
\includegraphics[width=\realinsetwidth]{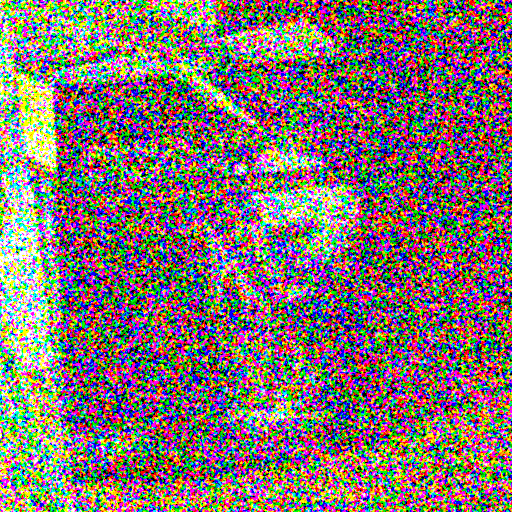} & 
\includegraphics[width=\realinsetwidth]{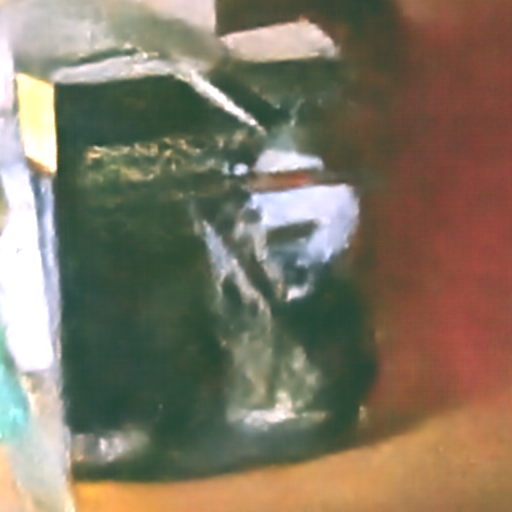} & 
\includegraphics[width=\realinsetwidth]{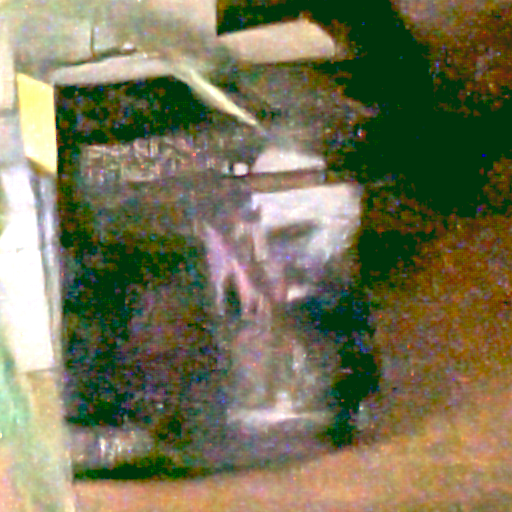} & 
\includegraphics[width=\realinsetwidth]{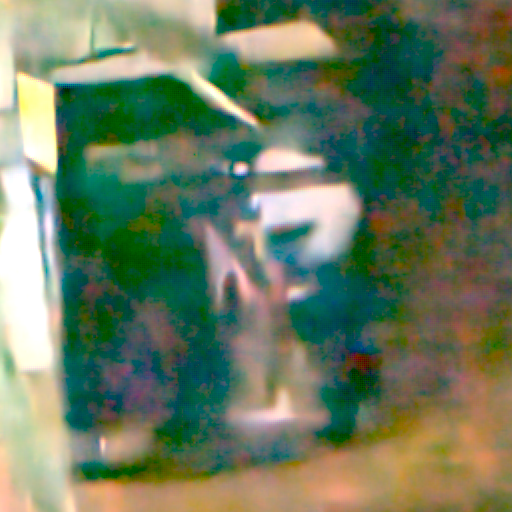} & 
\includegraphics[width=\realinsetwidth]{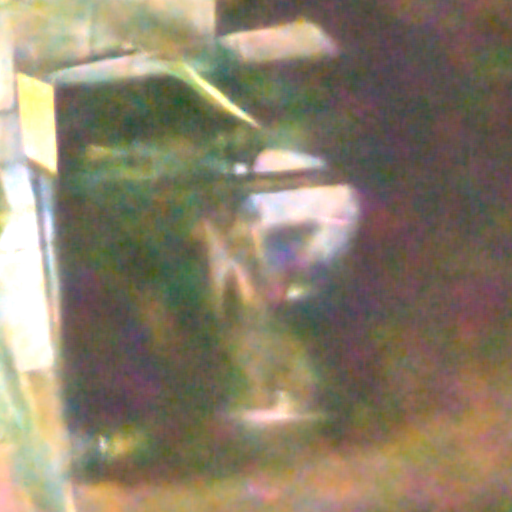} & 
\includegraphics[width=\realinsetwidth]{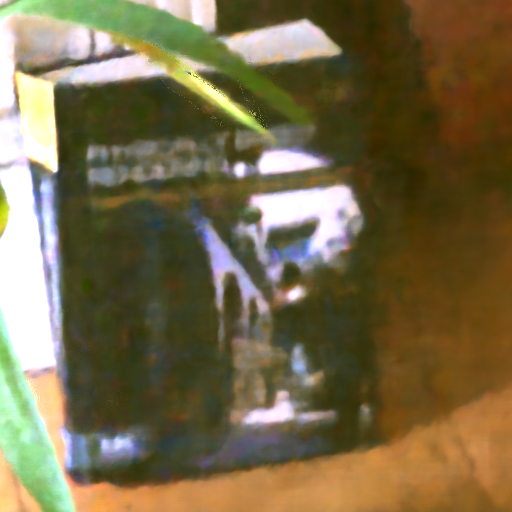} & 
\includegraphics[width=\realinsetwidth]{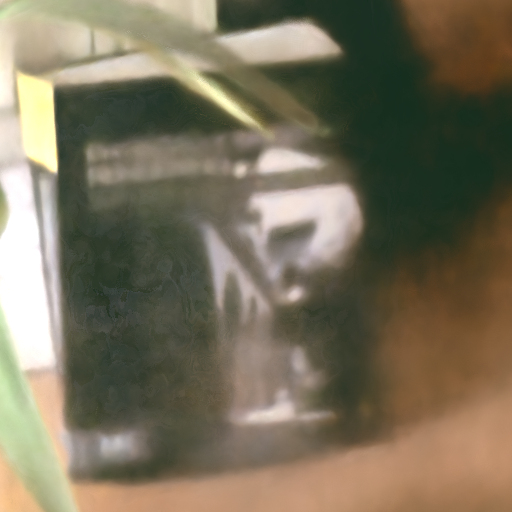} & 
\includegraphics[width=\realinsetwidth]{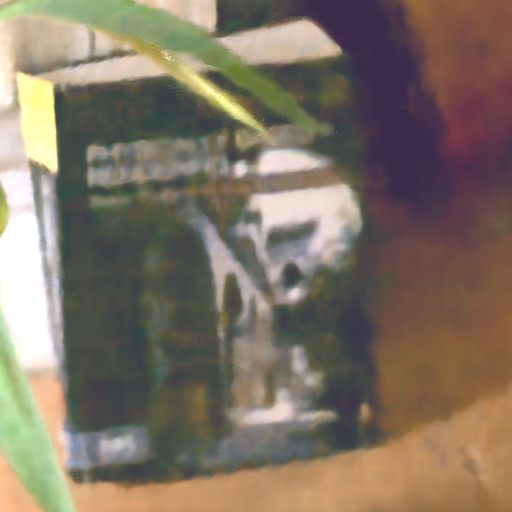} & 
\includegraphics[width=\realinsetwidth]{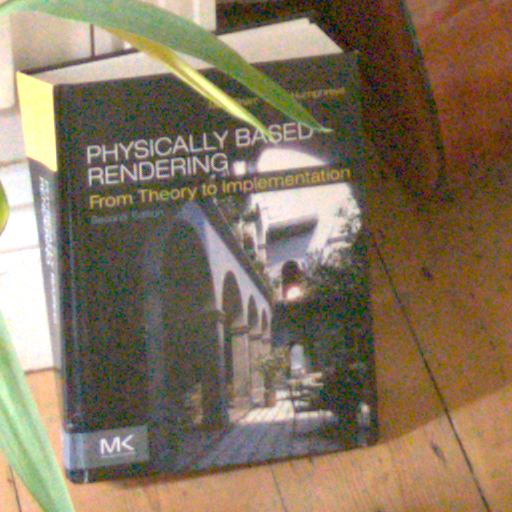} &
\includegraphics[width=\realfullwidth]{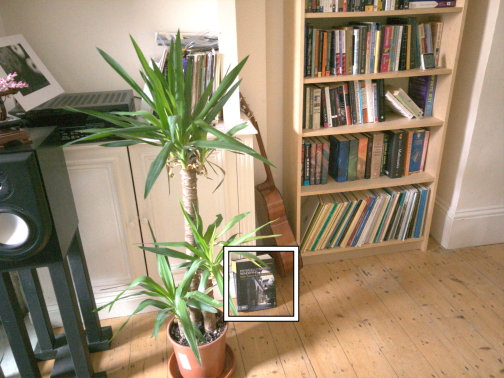} \\

\includegraphics[width=\realinsetwidth]{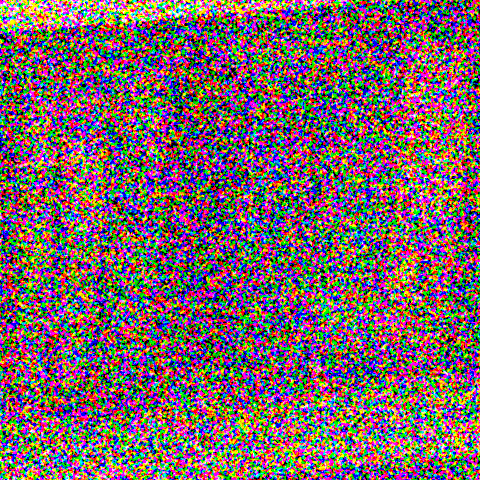} & 
\includegraphics[width=\realinsetwidth]{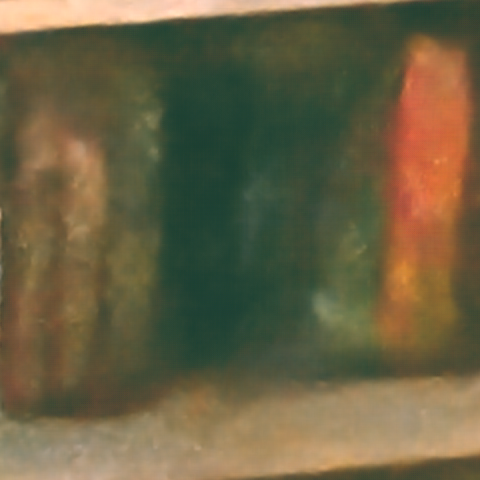} & 
\includegraphics[width=\realinsetwidth]{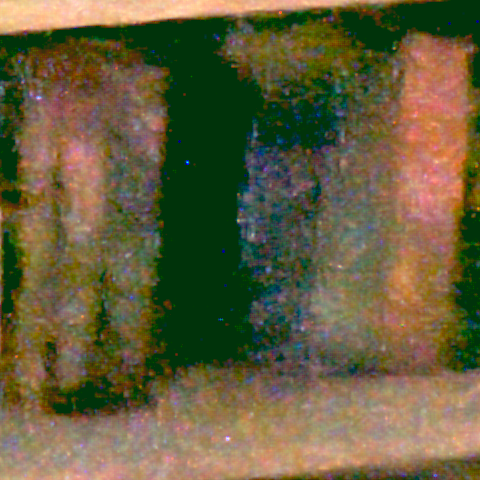} & 
\includegraphics[width=\realinsetwidth]{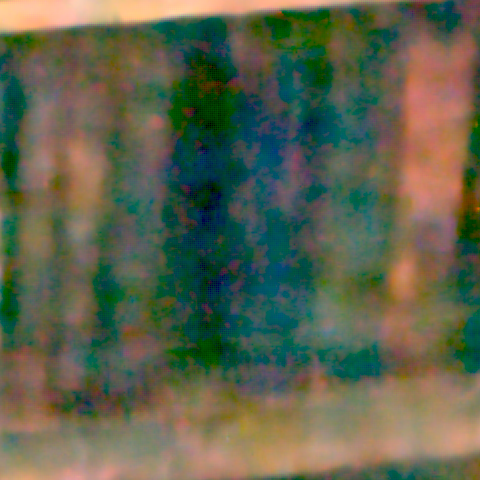} & 
\includegraphics[width=\realinsetwidth]{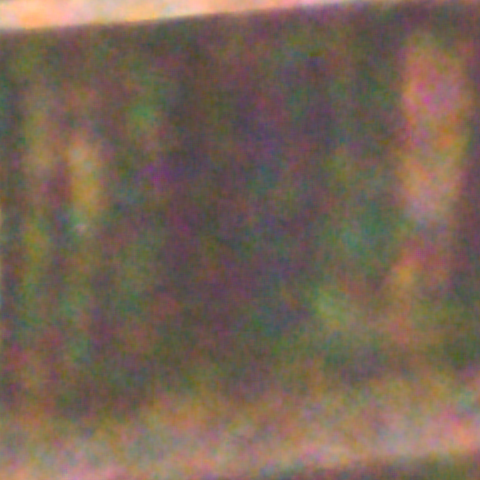} & 
\includegraphics[width=\realinsetwidth]{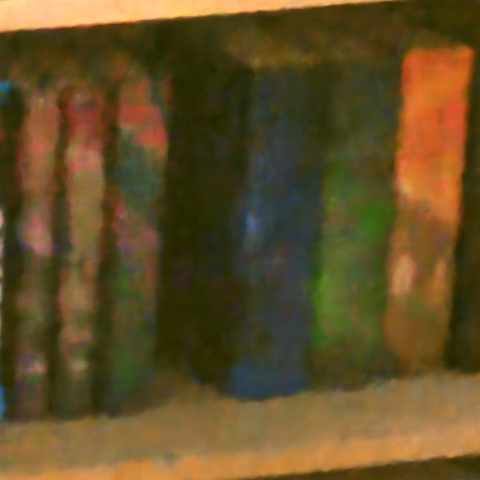} & 
\includegraphics[width=\realinsetwidth]{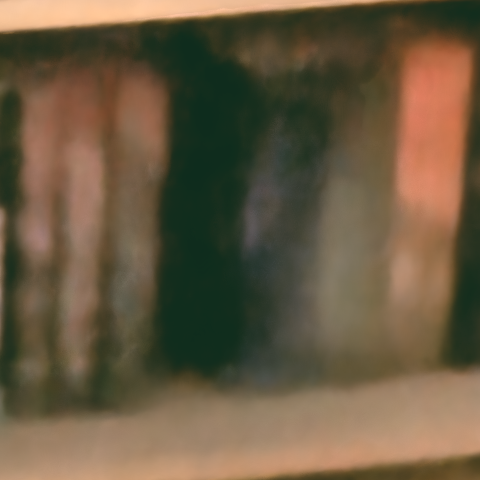} & 
\includegraphics[width=\realinsetwidth]{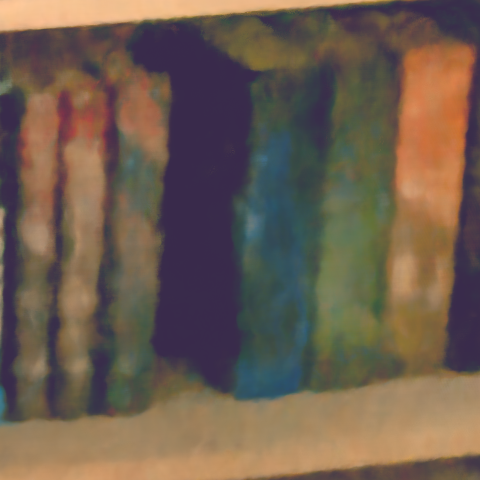} & 
\includegraphics[width=\realinsetwidth]{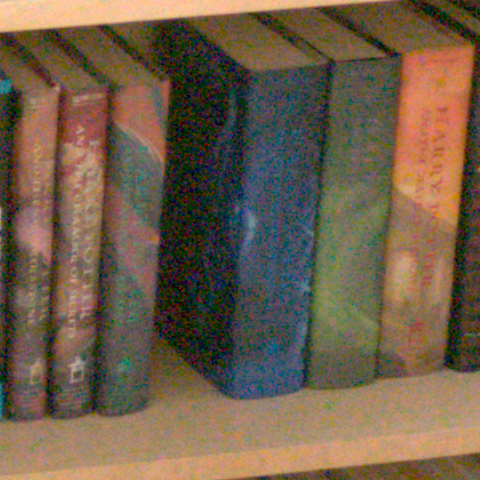} &
\includegraphics[width=\realfullwidth]{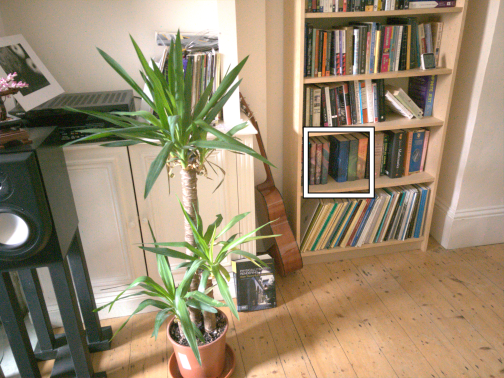} \\

\includegraphics[width=\realinsetwidth]{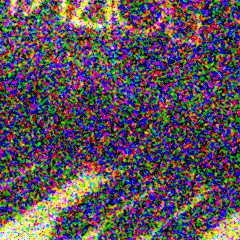} & 
\includegraphics[width=\realinsetwidth]{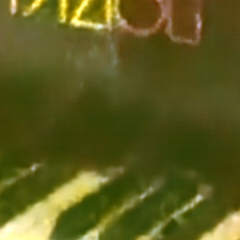} & 
\includegraphics[width=\realinsetwidth]{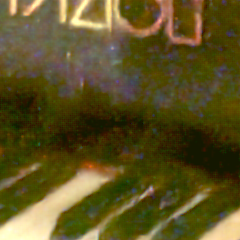} & 
\includegraphics[width=\realinsetwidth]{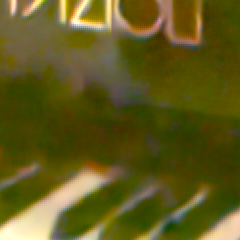} & 
\includegraphics[width=\realinsetwidth]{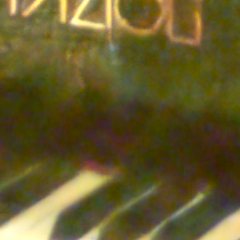} & 
\includegraphics[width=\realinsetwidth]{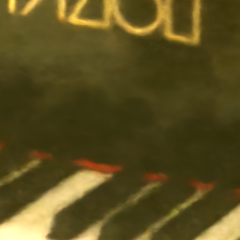} & 
\includegraphics[width=\realinsetwidth]{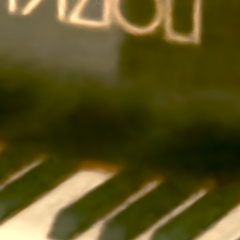} & 
\includegraphics[width=\realinsetwidth]{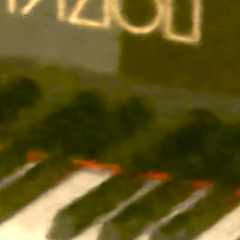} & 
\includegraphics[width=\realinsetwidth]{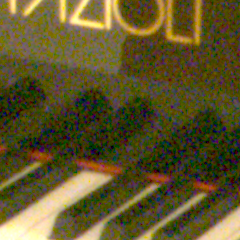} &
\includegraphics[width=\realfullwidth]{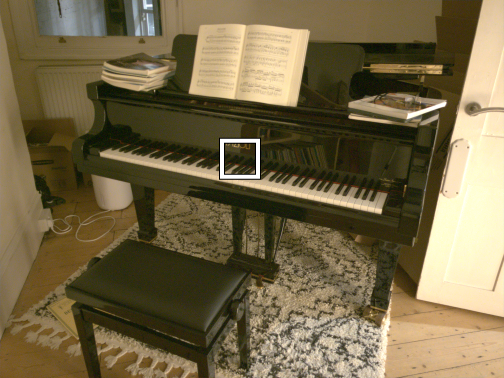} \\

\includegraphics[width=\realinsetwidth]{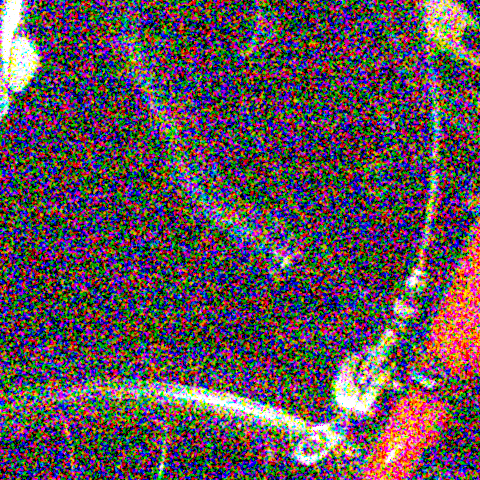} & 
\includegraphics[width=\realinsetwidth]{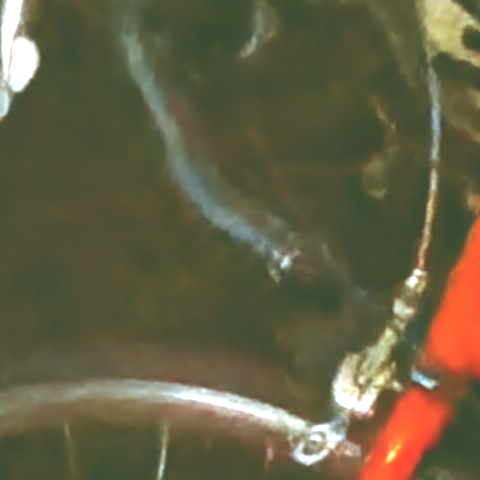} & 
\includegraphics[width=\realinsetwidth]{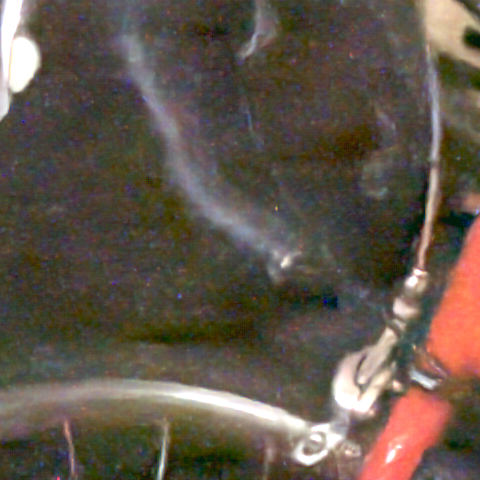} & 
\includegraphics[width=\realinsetwidth]{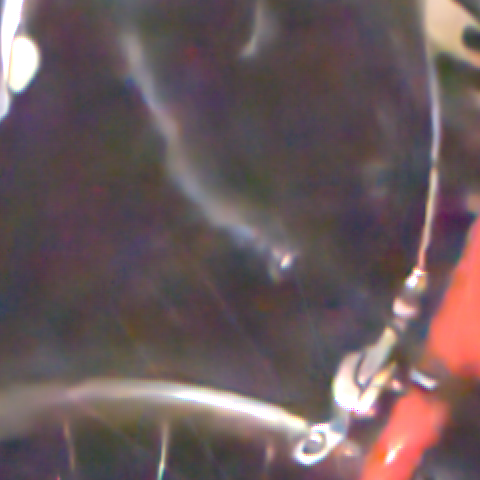} & 
\includegraphics[width=\realinsetwidth]{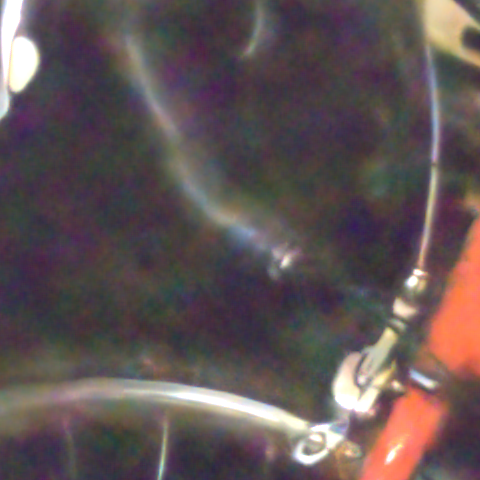} & 
\includegraphics[width=\realinsetwidth]{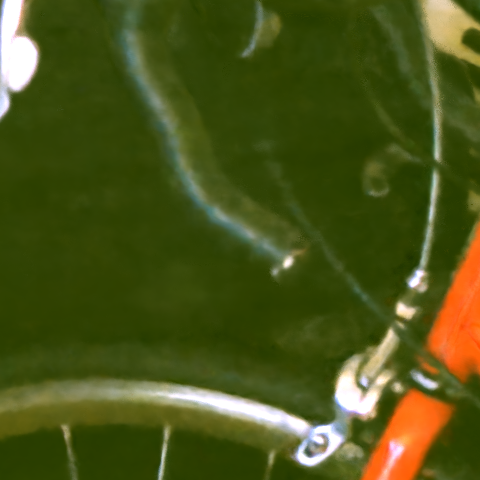} & 
\includegraphics[width=\realinsetwidth]{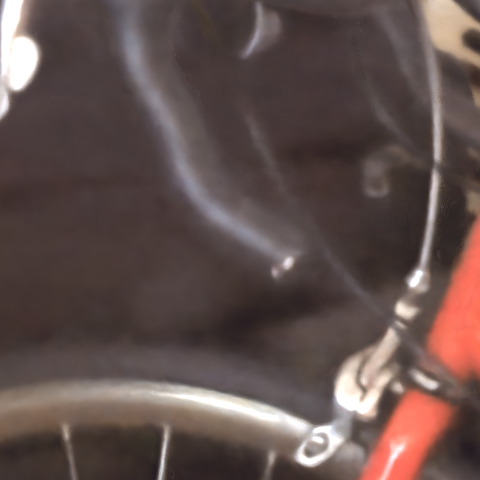} & 
\includegraphics[width=\realinsetwidth]{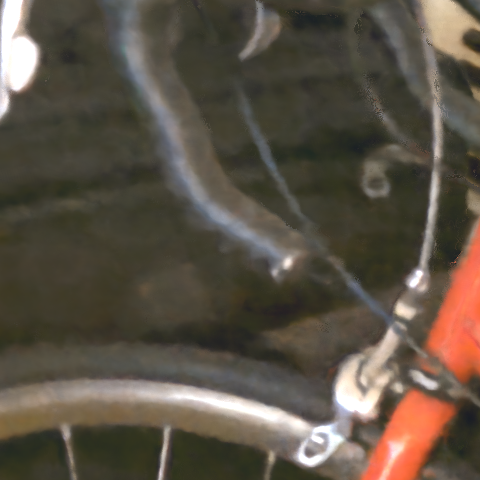} & 
\includegraphics[width=\realinsetwidth]{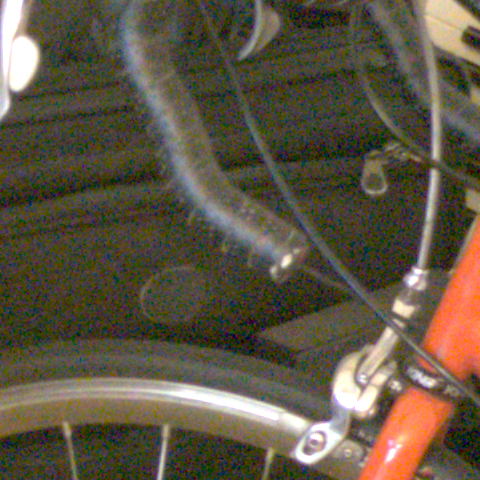} &
\includegraphics[width=\realfullwidth]{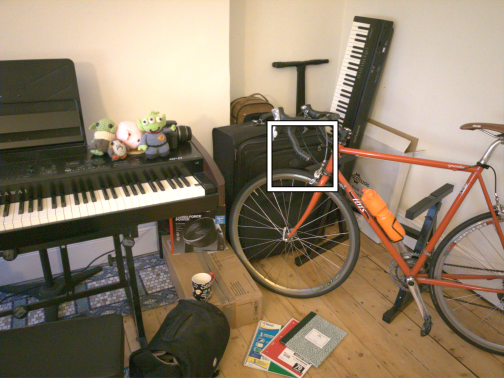} \\

Noisy image & SID~\cite{chen2018cvpr} & Unprocess~\cite{brooks2019cvpr} & RViDeNet~\cite{rvidenet} & UDVD~\cite{udvd} & LDR NeRF~\cite{barron2021} & Un+RawNeRF & RawNeRF & GT crop & Ground truth \\[-1ex]
& \multicolumn{2}{@{}c@{}}{\upbracefill} & \upbracefill & \upbracefill &  
\multicolumn{3}{@{}c@{}}{\upbracefill}
& \\[-0ex]
& \multicolumn{2}{@{}c@{}}{1 input} & 3 inputs & 5 inputs &  
\multicolumn{3}{@{}c@{}}{100 inputs, excluding test image}
& &

    \end{tabular}
    }
    \caption{
    Example postprocessed and color-aligned patches from our real denoising dataset. RawNeRF produces the most detailed output in each case. All deep denoising methods (columns 2-5) receive the noisy test image as input, whereas NeRF variants (columns 6-8) perform both novel view synthesis and denoising. 
    }
    \label{fig:realdenoising}
\end{figure*}

\newcommand{\cnnfigstreet}{%
  \parbox[b]{\fma}{\centering\vspace{0mm}%
    \includegraphics[width=\fma]{instant3d/figures/monodepth/street/example_color.jpg}\vspace{-1mm}\\%
  	\footnotesize Example input\vspace{1mm}\\
    \includegraphics[width=\fma]{instant3d/figures/monodepth/street/example_depth.png}\vspace{-1mm}\\%
	 \footnotesize CNN depth}%
  \hfill%
	$\underbracket[1pt][2.0mm]{%
	  \parbox[b]{3.1\fmb}{%
      \cnnres{street}{global}{\footnotesize\cnnglobal}{0mm}{0mm}{-0.5mm}%
      \hfill%
      \cnnres{street}{colmap}{\footnotesize\cnncolmap}{-1.5mm}{-1.5mm}{-1.5mm}%
      \hfill%
      \cnnres{street}{mono}{\footnotesize\cnnours}{0mm}{0mm}{-0.5mm}%
		}%
	}_{\substack{\vspace{-3.0mm}\\\colorbox{white}{\footnotesize~~CNN depth~~}}}$%
  \hfill%
	$\underbracket[1pt][2.0mm]{%
	  \parbox[t]{\fmb}{%
      \cnnres{street}{dual}{\footnotesize\final{\cnnours}}{0mm}{0mm}{-0.5mm}%
		}%
	}_{\substack{\vspace{-3.5mm}\\\colorbox{white}{\scriptsize~~\final{Dual camera depth}~~}}}$%
	\\}%

In scenes with very high dynamic range, even a 10-14 bit raw image may not be sufficient for capturing both bright and dark regions in a single exposure. This is addressed by the ``bracketing'' mode included in many digital cameras, where multiple images with varying shutter speeds are captured in a burst, then merged to take advantage of the bright highlights preserved in the shorter exposures and the darker regions captured with more detail in the faster exposures.

We can similarly take advantage of variable exposures in RawNeRF (Figure~\ref{fig:shortlong}). Given a sequence of images $I_i$ with exposure times $t_i$ (and all other capture parameters held constant), we can
``expose'' RawNeRF's linear space color output to match the brightness in image $I_i$ by scaling it by the recorded shutter speed $t_i$.
In practice, we find that varying exposures cannot be precisely aligned using shutter speed alone due to sensor miscalibration (see supplement). 
To correct for this, we add a learned per-color-channel scaling factor for each unique shutter speed present in the set of captured images, which we jointly optimize along with the NeRF network. The final RawNeRF ``exposure'' given a output color $\hat y_i$ from the network is then
$
    \min(\hat y_i^c \cdot t_i \cdot \alpha_{t_i}^c, 1),
$
where $c$ indexes color channels, and $\alpha_{t_i}^c$ is the learned scaling factor for shutter speed $t_i$ and channel $c$ (we constrain $\alpha_{t_\textrm{max}}^c=1$ for the longest exposure).
We clip from above at 1 to account for the fact that pixels saturate in overexposed regions. 
This scaled and clipped value is passed to the previously described loss (Equation~\ref{eq:loss}).

\begin{table}[]
\centering
\resizebox{\linewidth}{!}{
\begin{tabular}{@{}l|c|c|ccc@{}}
& Num. & Raw & \multicolumn{3}{c}{Affine-aligned sRGB} \\
Method  & inputs   & PSNR$\uparrow$  & PSNR$\uparrow$ & SSIM$\uparrow$ & LPIPS$\downarrow$ \\ \hline
Noisy input & -                          &  54.38  &  10.24  &  0.035  &  0.733   \\
SID~\cite{chen2018cvpr} & $1$            &    -    &  21.62  & \cellcolor{yellow}0.525  &  0.547   \\
Unprocess~\cite{brooks2019cvpr} & $1$    & \cellcolor{tabred}70.80  & \cellcolor{yellow}23.02  &  0.491  & \cellcolor{tabred}0.489   \\
RViDeNet~\cite{rvidenet} & $3$           & \cellcolor{yellow}68.29  &  22.20  &  0.516  &  0.545   \\
UDVD~\cite{udvd} & $5$                   & \cellcolor{orange}70.68  &  22.75  &  0.514  & \cellcolor{yellow}0.507   \\
LDR NeRF~\cite{barron2021} & $N-1$   &    -    &  19.43  &  0.518  &  0.544   \\
Un+RawNeRF & $N-1$                   &  67.99  & \cellcolor{orange}23.35  & \cellcolor{orange}0.531  &  0.507   \\
RawNeRF & $N-1$                          &  67.20  & \cellcolor{tabred}23.53  & \cellcolor{tabred}0.536  & \cellcolor{orange}0.501
\end{tabular}
}
\caption{We compare RawNeRF's denoising performance to various single and multi-image denoisers and NeRF ablations. Despite only being optimized on a single scene and never having seen even a noisy version of the test view, RawNeRF achieves results competitive with deep denoising methods trained on large image datasets. RawNeRF also outperforms NeRF trained on LDR sRGB images (LDR NeRF) and an ablation where RawNeRF's inputs have been denoised using ``Unprocess'' (Un+RawNeRF).
}
\label{table:realdenoising}
\end{table}

\subsection{Implementation details}

Our implementation is based on the mip-NeRF~\cite{barron2021} codebase, which improves upon the positional encoding used in the original NeRF method. Please see that paper for further details on the MLP scene representation and volumetric rendering algorithm. 
Our only network architecture change is to modify the activation function for the MLP's output color from a sigmoid to an exponential function to better parameterize linear radiance values. We use the Adam optimizer~\cite{adam} with batches of $16$k random rays sampled across all training images and a learning rate decaying from $10^{-3}$ to $10^{-5}$ over $500$k steps of optimization.

We find that extremely noisy scenes benefit from a regularization loss on volume density to prevent partially transparent ``floater'' artifacts. We apply a loss on the variance of the weight distribution used to accumulate color values along the ray during volume rendering; please see the supplement for details.

As our raw input data is mosaicked, it only contains one color value per pixel. We  
only apply the loss to the active color channel for each pixel,
such that optimizing NeRF effectively demosaics the input images. 
Since any resampling steps will effect the raw noise distribution, we do not undistort or downsample the inputs, and instead train using the full resolution mosaicked images (usually 12MP for our scenes).
To achieve this, we use camera intrinsics to account for radial distortion when generating rays.
We use full resolution postprocessed JPEG images to calculate camera poses as COLMAP~\cite{colmap} does not support raw images.

\section{Results}
\label{sec:results}

We present results exploring two consequences of supervising NeRF with raw HDR data. First, we show that RawNeRF is surprisingly robust to high levels of noise, to the extent that it can act as a competitive multi-image denoiser when applied to wide-baseline images of a static scene. Second, we demonstrate the HDR view synthesis applications enabled by recovering a scene representation that preserves high dynamic range color values.

\subsection{Denoising}

Recent years have seen an increasing focus on developing deep learning methods for denoising images directly in the raw linear domain~\cite{brooks2019cvpr,chen2018cvpr}. This effort has expanded to include multi-image denoisers that can be applied to burst images or video frames~\cite{chen2019iccv,rvidenet,udvd}. These multi-image denoisers typically assume that there is a relatively small amount of motion between frames, but that there may be large amounts of object motion within the scene. When nearby frames can be well aligned, these methods merge information from similar image patches (typically across 2-8 neighboring images) to outperform single image denoisers.

By comparison, NeRF (and by extension, RawNeRF) optimizes for a single scene reconstruction that is consistent with \emph{all} input images. By specializing to wide-baseline static scenes and taking advantage of 3D multiview information, RawNeRF can aggregate observations from much more widely spaced input images than a typical multi-image denoising method.

\paragraph{Real dataset} We collect a real world denoising dataset with 3 different scenes, each consisting of 101 noisy images and a clean reference image merged from stabilized long exposures. The first 100 images are taken handheld across a wide baseline (a standard forward-facing NeRF capture), using a fast shutter speed to accentuate noise. We then capture a stabilized burst of 50-100 longer exposures on a tripod and robustly merge them using HDR+~\cite{hdrplus} to create a clean ground truth frame. One additional tripod image taken at the original fast shutter speed serves as a noisy input ``base frame'' for the deep denoising methods. All images are taken with an iPhone X at 12MP resolution using the wide-angle lens and saved as 12-bit raw DNG files.

\paragraph{Comparisons} In Table~\ref{table:realdenoising} and Figure~\ref{fig:realdenoising}, we compare RawNeRF's joint view synthesis and denoising performance to several recent deep single and multi-image denoising methods. Note that all denoisers require the noisy version of the test image as input, whereas RawNeRF and its ablations only require its camera pose.

We focus our comparison on methods explicitly designed to handle raw input images. Chen \etal~\cite{chen2018cvpr} (SID) present a single image denoiser that maps from raw inputs to postprocessed LDR images and is trained on a large dataset of noisy raw and clean postprocessed image pairs collected by the authors. Brooks \etal~\cite{brooks2019cvpr} (Unprocess) is a method for training a raw single image denoiser on simulated raw data created from internet image datasets that transfers well to real raw images. RViDeNet~\cite{rvidenet} trains a raw video denoiser on a combination of Unprocessing-style synthetic data and a new real raw video dataset. Sheth \etal~\cite{udvd} (UDVD) present a ``self-supervised'' method for training a video denoiser only using noisy data, building on ideas from Noise2Noise~\cite{lehtinen2018} and blind-spot networks~\cite{laine2019blindspot}. UDVD provides network weights specifically trained on the raw video dataset from RViDeNet. For all methods, we use publicly available code and pretrained model weights.

We also compare to two ablations of our method. LDR NeRF represents mip-NeRF~\cite{barron2021} trained (as usual) in LDR sRGB space on images postprocessed by a minimal sRGB tonemapping pipeline. ``Un+RawNeRF'' preprocesses the training images using the single image raw denoiser from Brooks \etal~\cite{brooks2019cvpr} (``Unprocess'') before training RawNeRF.

\newcommand{\imwidth}{.25\columnwidth}

\newcommand{\textimage}[2]{
	\begin{overpic}[width=\imwidth]{figures/lego_noise_v2/#1}
	\put (67,2) {\footnotesize \sethlcolor{white}\hl{$#2$}}
    \end{overpic}
}

\begin{figure}[]
    \centering
\resizebox{\linewidth}{!}{
    
    \begin{tabular}{@{}c@{\,}c@{\,}c@{\,}c@{\,}c@{}}
    & \multicolumn{4}{c}{Simulated shutter speed (seconds)} \\
    & $\infty$ & $1/15$ & $1/60$ & $1/240$ 
    \\
    \rotatebox[origin=l]{90}{\,\,\,Test image}
    &
    {
	\begin{overpic}[width=\imwidth]{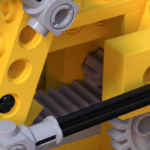}
	\put (67,2) { \phantom{\footnotesize $19.55$}}
    \end{overpic}
    }  & 
    \textimage{0_2.png}{19.55}  & 
    \textimage{0_4.png}{12.38}  & 
    \textimage{0_6.png}{\phantom{3}7.09}  \\ 
    
    \rotatebox[origin=l]{90}{\,\,\,LDR NeRF}
         & 
    {
	\begin{overpic}[width=\imwidth]{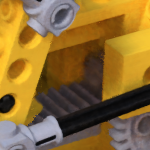}
	\put (67,2) {\footnotesize \sethlcolor{white}\hl{$31.71$}}
    \end{overpic}
} &
    \textimage{1_2.png}{28.76}  & 
    \textimage{1_4.png}{21.73}  & 
    \textimage{1_6.png}{14.51}  \\ 
    
    \rotatebox[origin=l]{90}{\,\,\,RawNeRF}
         & 
    {
	\begin{overpic}[width=\imwidth]{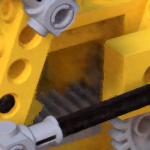}
	\put (67,2) {\footnotesize \sethlcolor{white}\hl{$30.40$}}
    \end{overpic}
} &
    \textimage{2_2.png}{29.76}  & 
    \textimage{2_4.png}{28.73}  & 
    \textimage{2_6.png}{24.64}  \\ 
    \end{tabular}
    
    }
    \caption{Example patches from the synthetic scene used in Table~\ref{tab:synthlego}, annotated with sRGB PSNR for each inset. 
    With perfectly clean inputs, training on LDR images is superior, but with any nonzero amount of noise, it is more beneficial to optimize NeRF in raw space, where the noise distribution remains unbiased.
    }
    \label{fig:synthlego}
\end{figure}

\begin{table}[]
    \centering

\resizebox{\linewidth}{!}{
\begin{tabular}{@{}l|ccccccc@{}}
\multicolumn{1}{c|}{} & \multicolumn{7}{c}{Simulated shutter speed (seconds)} \\
Method & $\infty$ &  $1/7   $ & $1/15  $ & $1/30  $ & $1/60  $ & $1/120 $ & $1/240 $ \\ \hline
Noisy input  &   -   & 23.33 & 19.65 & 16.03 & 12.51 &  9.40 &  7.18 \\
LDR NeRF     & \textbf{33.16} & 31.25 & 29.14 & 26.10 & 22.31 & 18.27 & 14.87 \\
RawNeRF      & 32.15 & \textbf{32.11} & \textbf{31.94} & \textbf{31.59} & \textbf{30.94} & \textbf{29.69} & \textbf{27.73} \\
    \end{tabular}
    }
    
    \caption{
    We perform an ablation study on a synthetically rendered raw dataset with 120 training images, simulating shot and read noise for 8 different shutter speeds.
    Here we report PSNR values in LDR sRGB space. 
    }
    \label{tab:synthlego}
\end{table}

\begin{figure*}
    \centering
    \includegraphics[width=\textwidth]{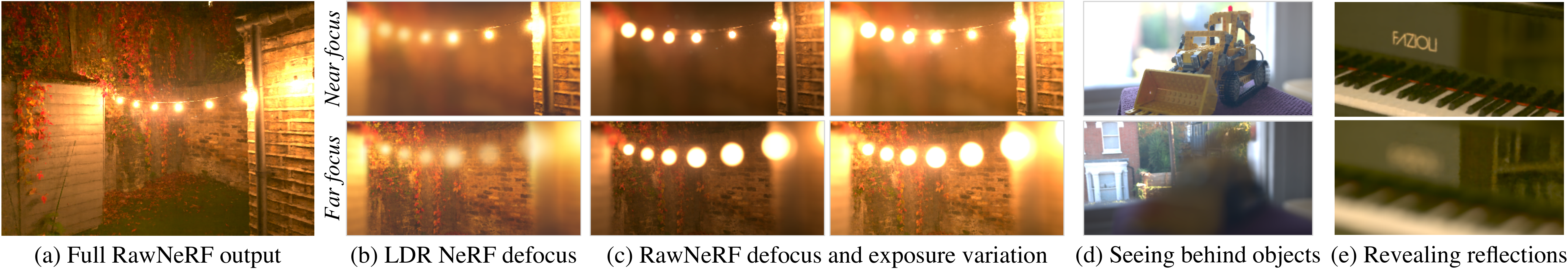}
    \caption{Synthetic defocus examples. In this nighttime garden scene (a), LDR NeRF cannot accurately render defocused bright highlights since it is trained on images that have already been tonemapped and clipped (b). RawNeRF recovers the linear intensity of the light sources such that applying defocus blur produces correctly oversaturated ``bokeh balls'' (c). 
    Since RawNeRF is optimized for view synthesis from wide-baseline inputs, it can achieve 3D defocus effects not possible with a single image and depth map, such as revealing occluded parts of the background by focusing behind the foreground bulldozer (d) or focusing on the bookshelves reflected above the piano keys (e).
    }
    \label{fig:bokeh_banner}
\end{figure*}

All compared methods take mosaicked raw images as input. 
Every deep denoiser~\cite{chen2018cvpr,brooks2019cvpr,rvidenet,udvd} uses the noisy ``base frame'' as input, and the two multi-image denoising networks~\cite{rvidenet,udvd} also receive the nearest images from the wide-baseline capture (based on camera position).
We convert the 12-bit raw input to floating point by normalizing with the white and black levels. 
Since each method was trained on raw data from a different source, they impart different color tints to the output. So this not affect metrics, we calculate a per-color-channel affine transform that best matches each method's raw output to the ground truth raw image. (The exceptions are SID and LDR NeRF, whose sRGB output we match to the postprocessed sRGB ground truth.) Our basic postprocessing pipeline for visualization and computing sRGB metrics is to apply a bilinear demosaic (when necessary), perform white balance/color correction, rescale white level, clip to $[0, 1]$, and apply the sRGB gamma curve. Please see the supplement for details.

\paragraph{Analysis} Despite simultaneously performing denoising and novel view synthesis, our method is competitive with all compared deep denoisers (Table~\ref{table:realdenoising}, Figure~\ref{fig:realdenoising}). We suspect that the multi-image denoisers struggle to make use of the additional frames provided from the wide-baseline capture, as the camera movement is larger than in a typical sub-second burst or video clip. By comparison, RawNeRF, despite lacking any explicitly learned image priors, clean training data, or even a ``base frame'' input image, produces high quality outputs by combining information from across all input images in its reconstruction. 
Despite the fact that LDR NeRF is directly trained to minimize mean-squared error in sRGB space, RawNeRF achieves significantly better sRGB metrics.
We also find that applying a single image denoiser to the inputs before training RawNeRF results in oversmoothed renderings (Un+RawNeRF).

\paragraph{Synthetic noise ablation} In Table~\ref{tab:synthlego} and Figure~\ref{fig:synthlego}, we demonstrate the impact of noise level on RawNeRF image quality. For training, we render 120 linear HDR images using the \emph{Lego} scene from NeRF~\cite{mildenhall2020nerf}, borrowing color correction, white balance, and noise parameters from our iPhone captures' EXIF metadata to ``unprocess'' this data into raw space~\cite{brooks2019cvpr}. Since the renderings have a large amount of empty space, we report sRGB PSNR on the object only, by using the provided alpha masks (otherwise error from the background pixels heavily penalizes LDR NeRF). Even in this synthetic setting free from camera miscalibration issues, we can clearly observe the color bias and loss of detail caused by training LDR NeRF on postprocessed noisy data.

\subsection{HDR view synthesis applications}

\paragraph{Modifying exposure and tonemapping} Figures~\ref{fig:teaser}, \ref{fig:raw_vs_srgb},
\ref{fig:l2loss}, \ref{fig:shortlong}, and \ref{fig:bokeh_banner} include examples of varying the exposure level and tonemapping algorithm for images output by RawNeRF, which exist in linear HDR space and can thus be postprocessed like a raw photo from a digital camera. 
Please see our supplement and video for many more examples.

\paragraph{Synthetic defocus} 
Given a full 3D model of a scene, physically-based renderers accurately simulate camera lens defocus effects by tracing rays refracted through each lens element~\cite{lenstracing}, but this process is extremely computationally expensive. 
A reasonably convincing and much cheaper solution is to apply a varying blur kernel to different depth layers of the scene and composite them together~\cite{barron2015stereo,wadhwa2018defocus}. In Figure~\ref{fig:bokeh_banner}, we apply this synthetic defocus rendering model to sets of RGBA depth layers precomputed from trained RawNeRF models (similar to a multiplane image~\cite{zhou18stereomag}). 
As shown by Zhang \etal~\cite{zhang2019defocus}, recovering linear HDR color is critical for achieving the characteristic oversaturated ``bokeh balls'' around defocused bright light sources.

\section{Discussion}

We have demonstrated the benefits of training NeRF directly on linear raw camera images. However, this modification is not without tradeoffs. Most digital cameras can only save raw images at full resolution with minimal compression, 
resulting in huge storage requirements when capturing tens or hundreds of images per scene.
Our method is also dependent on COLMAP's~\cite{colmap} robustness for computing camera poses, preventing us from capturing scenes below a certain light level. This could potentially be addressed by jointly optimizing RawNeRF and the input camera poses~\cite{wang2021nerfmm,lin2021barf}. Finally, despite its robustness to noise, RawNeRF cannot be considered a general purpose denoiser as it cannot handle scene motion and requires orders of magnitude more computation than a feed-forward network. 

Despite these shortcomings, we believe that RawNeRF represents a step toward robust, high quality capture of real world environments. Training on raw images with variable exposure allows us to capture scenes with a much wider dynamic range, and robustness to noise makes reconstructing dark nighttime captures possible. Lifting these constraints greatly increases the fraction of the world that can be reconstructed and explored with photorealistic view synthesis.

{\small
\bibliographystyle{ieee_fullname}
\bibliography{egbib}

\begin{thebibliography}{10}\itemsep=-1pt

\bibitem{attal2021torf}
Benjamin Attal, Eliot Laidlaw, Aaron Gokaslan, Changil Kim, Christian Richardt,
  James Tompkin, and Matthew O'Toole.
\newblock T\"orf: Time-of-flight radiance fields for dynamic scene view
  synthesis, 2021.

\bibitem{barron2015stereo}
Jonathan~T. Barron, Andrew Adams, YiChang Shih, and Carlos Hern\'andez.
\newblock Fast bilateral-space stereo for synthetic defocus.
\newblock {\em CVPR}, 2015.

\bibitem{barron2021}
Jonathan~T. Barron, Ben Mildenhall, Matthew Tancik, Peter Hedman, Ricardo
  Martin-Brualla, and Pratul~P. Srinivasan.
\newblock {Mip-NeRF}: A multiscale representation for anti-aliasing neural
  radiance fields.
\newblock {\em ICCV}, 2021.

\bibitem{batson2019noise2self}
Joshua Batson and Loic Royer.
\newblock Noise2self: Blind denoising by self-supervision.
\newblock {\em ICML}, 2019.

\bibitem{brooks2019cvpr}
Tim Brooks, Ben Mildenhall, Tianfan Xue, Jiawen Chen, Dillon Sharlet, and
  Jonathan~T. Barron.
\newblock Unprocessing images for learned raw denoising.
\newblock {\em CVPR}, 2019.

\bibitem{buehler01unstructlumigraph}
Chris Buehler, Michael Bosse, Leonard McMillan, Steven Gortler, and Michael
  Cohen.
\newblock Unstructured lumigraph rendering.
\newblock {\em SIGGRAPH}, 2001.

\bibitem{mit5k}
Vladimir Bychkovsky, Sylvain Paris, Eric Chan, and Fr{\'e}do Durand.
\newblock Learning photographic global tonal adjustment with a database of
  input / output image pairs.
\newblock {\em CVPR}, 2011.

\bibitem{chen2019iccv}
Chen Chen, Qifeng Chen, Minh Do, and Vladlen Koltun.
\newblock Seeing motion in the dark.
\newblock {\em ICCV}, 2019.

\bibitem{chen2018cvpr}
Chen Chen, Qifeng Chen, Jia Xu, and Vladlen Koltun.
\newblock Learning to see in the dark.
\newblock {\em CVPR}, 2018.

\bibitem{cook1984distributed}
Robert~L. Cook, Thomas Porter, and Loren Carpenter.
\newblock Distributed ray tracing.
\newblock {\em SIGGRAPH Comput. Graph.}, 1984.

\bibitem{bm3d}
Kostadin Dabov, Alessandro Foi, Vladimir Katkovnik, and Karen Egiazarian.
\newblock Image denoising by sparse 3-d transform-domain collaborative
  filtering.
\newblock {\em TIP}, 2007.

\bibitem{debevec1997}
Paul~E. Debevec and Jitendra Malik.
\newblock Recovering high dynamic range radiance maps from photographs.
\newblock {\em SIGGRAPH}, 1997.

\bibitem{ehret2019f2f}
Thibaud Ehret, Axel Davy, Jean-Michel Morel, Gabriele Facciolo, and Pablo
  Arias.
\newblock Model-blind video denoising via frame-to-frame training.
\newblock {\em CVPR}, 2019.

\bibitem{ldr2hdrcnn}
Gabriel Eilertsen, Joel Kronander, Gyorgy Denes, Rafał Mantiuk, and Jonas
  Unger.
\newblock Hdr image reconstruction from a single exposure using deep cnns.
\newblock {\em ACM Transactions on Graphics (TOG)}, 2017.

\bibitem{flynn2016deepstereo}
John Flynn, Ivan Neulander, James Philbin, and Noah Snavely.
\newblock Deepstereo: Learning to predict new views from the world's imagery.
\newblock {\em CVPR}, 2016.

\bibitem{foi2008noisemodel}
Alessandro Foi, Mejdi Trimeche, Vladimir Katkovnik, and Karen~O. Egiazarian.
\newblock Practical poissonian-gaussian noise modeling and fitting for
  single-image raw-data.
\newblock {\em IEEE Transactions on Image Processing}, 2008.

\bibitem{hdrnet}
Micha{\"e}l Gharbi, Jiawen Chen, Jonathan~T Barron, Samuel~W Hasinoff, and
  Fr{\'e}do Durand.
\newblock Deep bilateral learning for real-time image enhancement.
\newblock {\em ACM Transactions on Graphics (TOG)}, 2017.

\bibitem{godard2018burst}
Clement Godard, Kevin Matzen, and Matt Uyttendaele.
\newblock Deep burst denoising.
\newblock {\em ECCV}, 2018.

\bibitem{cohen96lumigraph}
Steven~J Gortler, Radek Grzeszczuk, Richard Szeliski, and Michael~F Cohen.
\newblock The lumigraph.
\newblock {\em SIGGRAPH}, pages 43--54, 1996.

\bibitem{hdrplus}
Samuel~W. Hasinoff, Dillon Sharlet, Ryan Geiss, Andrew Adams, Jonathan~T..
  Barron, Florian Kainz, Jiawen Chen, and Marc Levoy.
\newblock Burst photography for high dynamic range and low-light imaging on
  mobile cameras.
\newblock {\em SIGGRAPH Asia}, 2016.

\bibitem{hedman2018deepblending}
Peter Hedman, Julien Philip, True Price, Jan-Michael Frahm, George Drettakis,
  and Gabriel Brostow.
\newblock Deep blending for free-viewpoint image-based rendering.
\newblock {\em SIGGRAPH Asia}, 2018.

\bibitem{hu2018whitebox}
Yuanming Hu, Hao He, Chenxi Xu, Baoyuan Wang, and Stephen Lin.
\newblock Exposure: A white-box photo post-processing framework.
\newblock {\em ACM Transactions on Graphics (TOG)}, 2018.

\bibitem{kalantari2017hdr}
Nima~Khademi Kalantari and Ravi Ramamoorthi.
\newblock Deep high dynamic range imaging of dynamic scenes.
\newblock {\em ACM Transactions on Graphics (Proceedings of SIGGRAPH 2017)},
  36(4), 2017.

\bibitem{kangle2021dsnerf}
Jun-Yan~Zhu Kangle~Deng, Andrew~Liu and Deva Ramanan.
\newblock Depth-supervised nerf: Fewer views and faster training for free.
\newblock {\em arXiv:2107.02791}, 2021.

\bibitem{adam}
Diederik~P. Kingma and Jimmy Ba.
\newblock Adam: A method for stochastic optimization.
\newblock {\em ICLR}, 2015.

\bibitem{lenstracing}
Craig Kolb, Don Mitchell, and Pat Hanrahan.
\newblock A realistic camera model for computer graphics.
\newblock {\em SIGGRAPH}, 1995.

\bibitem{kopanas2021pbr}
Georgios Kopanas, Julien Philip, Thomas Leimk{\"u}hler, and George Drettakis.
\newblock Point-based neural rendering with per-view optimization.
\newblock {\em Eurographics}, 2021.

\bibitem{krull2019noise2void}
Alexander Krull, Tim-Oliver Buchholz, and Florian Jug.
\newblock Noise2void-learning denoising from single noisy images.
\newblock {\em CVPR}, 2019.

\bibitem{laine2019blindspot}
Samuli Laine, Tero Karras, Jaakko Lehtinen, and Timo Aila.
\newblock High-quality self-supervised deep image denoising.
\newblock {\em NeurIPS}, 2019.

\bibitem{lehtinen2018}
Jaakko Lehtinen, Jacob Munkberg, Jon Hasselgren, Samuli Laine, Tero Karras,
  Miika Aittala, and Timo Aila.
\newblock Noise2noise: Learning image restoration without clean data.
\newblock {\em ICML}, 2018.

\bibitem{levoy96lightfields}
Marc Levoy and Pat Hanrahan.
\newblock Light field rendering.
\newblock {\em SIGGRAPH}, 1996.

\bibitem{li2008demosaic}
Xin Li, Bahadir Gunturk, and Lei Zhang.
\newblock Image demosaicing: A systematic survey.
\newblock {\em Visual Communications and Image Processing 2008}, 2008.

\bibitem{lin2021barf}
Chen-Hsuan Lin, Wei-Chiu Ma, Antonio Torralba, and Simon Lucey.
\newblock {BARF}: Bundle-adjusting neural radiance fields.
\newblock {\em ICCV}, 2021.

\bibitem{lombardi2019neuralvolumes}
Stephen Lombardi, Tomas Simon, Jason Saragih, Gabriel Schwartz, Andreas
  Lehrmann, and Yaser Sheikh.
\newblock Neural volumes: Learning dynamic renderable volumes from images.
\newblock {\em SIGGRAPH}, 2019.

\bibitem{martinbrualla2020nerfw}
Ricardo Martin-Brualla, Noha Radwan, Mehdi S.~M. Sajjadi, Jonathan~T. Barron,
  Alexey Dosovitskiy, and Daniel Duckworth.
\newblock {NeRF in the Wild: Neural Radiance Fields for Unconstrained Photo
  Collections}.
\newblock {\em CVPR}, 2021.

\bibitem{kpn}
Ben Mildenhall, Jonathan~T. Barron, Jiawen Chen, Dillon Sharlet, Ren Ng, and
  Robert Carroll.
\newblock Burst denoising with kernel prediction networks.
\newblock {\em CVPR}, 2018.

\bibitem{mildenhall2020nerf}
Ben Mildenhall, Pratul~P. Srinivasan, Matthew Tancik, Jonathan~T. Barron, Ravi
  Ramamoorthi, and Ren Ng.
\newblock {NeRF}: Representing scenes as neural radiance fields for view
  synthesis.
\newblock {\em ECCV}, 2020.

\bibitem{plotz2017cvpr}
Tobias Pl\"otz and Stefan Roth.
\newblock Benchmarking denoising algorithms with real photographs.
\newblock {\em CVPR}, 2017.

\bibitem{riegler2020fvs}
Gernot Riegler and Vladlen Koltun.
\newblock Free view synthesis.
\newblock {\em ECCV}, 2020.

\bibitem{riegler2021svs}
Gernot Riegler and Vladlen Koltun.
\newblock Stable view synthesis.
\newblock {\em CVPR}, 2021.

\bibitem{ruckert2021adop}
Darius Rückert, Linus Franke, and Marc Stamminger.
\newblock Adop: Approximate differentiable one-pixel point rendering, 2021.

\bibitem{colmap}
Johannes~Lutz Sch\"{o}nberger and Jan-Michael Frahm.
\newblock Structure-from-motion revisited.
\newblock {\em CVPR}, 2016.

\bibitem{udvd}
Dev~Yashpal Sheth, Sreyas Mohan, Joshua Vincent, Ramon Manzorro, Peter~A.
  Crozier, Mitesh~M. Khapra, Eero~P. Simoncelli, and Carlos Fernandez-Granda.
\newblock Unsupervised deep video denoising.
\newblock {\em ICCV}, 2021.

\bibitem{srinivasan2018aperture}
Pratul~P. Srinivasan, Rahul Garg, Neal Wadhwa, Ren Ng, and Jonathan~T. Barron.
\newblock Aperture supervision for monocular depth estimation.
\newblock {\em CVPR}, 2018.

\bibitem{wadhwa2018defocus}
Neal Wadhwa, Rahul Garg, David~E. Jacobs, Bryan~E. Feldman, Nori Kanazawa,
  Robert Carroll, Yair Movshovitz-Attias, Jonathan~T. Barron, Yael Pritch, and
  Marc Levoy.
\newblock Synthetic depth-of-field with a single-camera mobile phone.
\newblock {\em ACM Trans. Graph.}, 2018.

\bibitem{wang2021nerfmm}
Zirui Wang, Shangzhe Wu, Weidi Xie, Min Chen, and Victor~Adrian Prisacariu.
\newblock Ne{RF}$--$: Neural radiance fields without known camera parameters.
\newblock {\em arXiv:2102.07064}, 2021.

\bibitem{wei2021nerfingmvs}
Yi Wei, Shaohui Liu, Yongming Rao, Wang Zhao, Jiwen Lu, and Jie Zhou.
\newblock Nerfingmvs: Guided optimization of neural radiance fields for indoor
  multi-view stereo.
\newblock {\em ICCV}, 2021.

\bibitem{jeong2021scnerf}
Christopehr Choy Animashree Anandkumar Minsu~Cho Yoonwoo~Jeong, Seokjun~Ahn and
  Jaesik Park.
\newblock Self-calibrating neural radiance fields.
\newblock {\em ICCV}, 2021.

\bibitem{rvidenet}
Huanjing Yue, Cong Cao, Lei Liao, Ronghe Chu, and Jingyu Yang.
\newblock Supervised raw video denoising with a benchmark dataset on dynamic
  scenes.
\newblock {\em CVPR}, 2020.

\bibitem{zhang2017beyond}
Kai Zhang, Wangmeng Zuo, Yunjin Chen, Deyu Meng, and Lei Zhang.
\newblock Beyond a {Gaussian} denoiser: Residual learning of deep {CNN} for
  image denoising.
\newblock {\em IEEE Transactions on Image Processing}, 2017.

\bibitem{zhang2019defocus}
Xuaner Zhang, Kevin Matzen, Vivien Nguyen, Dillon Yao, You Zhang, and Ren Ng.
\newblock Synthetic defocus and look-ahead autofocus for casual videography.
\newblock {\em SIGGRAPH}, 2019.

\bibitem{zhi2021semanticnerf}
Shuaifeng Zhi, Tristan Laidlow, Stefan Leutenegger, and Andrew Davison.
\newblock In-place scene labelling and understanding with implicit scene
  representation.
\newblock {\em ICCV}, 2021.

\bibitem{zhou18stereomag}
Tinghui Zhou, Richard Tucker, John Flynn, Graham Fyffe, and Noah Snavely.
\newblock Stereo magnification: Learning view synthesis using multiplane
  images.
\newblock {\em SIGGRAPH}, 2018.

\end{thebibliography}
}

\appendix

\section{Potential negative impact}

Training any NeRF model for scene reconstruction has potential negative environmental impact, as current algorithms are very compute-intensive, requiring hours of training per scene even when run on specialized ML accelerators. This also creates an unfair advantage for research groups with access to more computational resources. Future work will likely address this issue, as it blocks the widespread practical adoption of these models.

Any image restoration model could potentially be applied for illicit surveillance purposes. Multi-image denoisers provide the additional capability of potentially revealing details that are not visible in any single image due to noise. ML-based algorithms further complicate this situation by potentially ``hallucinating'' details in ambiguous regions, either intentionally (as with generative methods) or unintentionally (in the form of reconstruction artifacts). RawNeRF has a minimal ability to hallucinate, as it largely works by simply averaging the input data, but it does occasionally produce high frequency grid-like patterns due to the bias induced by positional encoding.

\section{Additional qualitative results}

We include additional qualitative results for both dark (Figure~\ref{fig:tableau_dark}) and high contrast scenes (Figure~\ref{fig:tableau_hdr}). We urge the reader to view our supplemental video as the results are more compelling when animated.

\newcommand{\tableauheight}{4cm}
\newcommand{\tableauinheight}{1.25cm}

\begin{figure*}[]
    \vspace{1.5cm}
    \centering
\resizebox{.965\linewidth}{!}{
    \begin{tabular}{@{}c@{\,}c@{\quad}c@{\,}c@{\quad}c@{}}
    \includegraphics[height=\tableauheight]{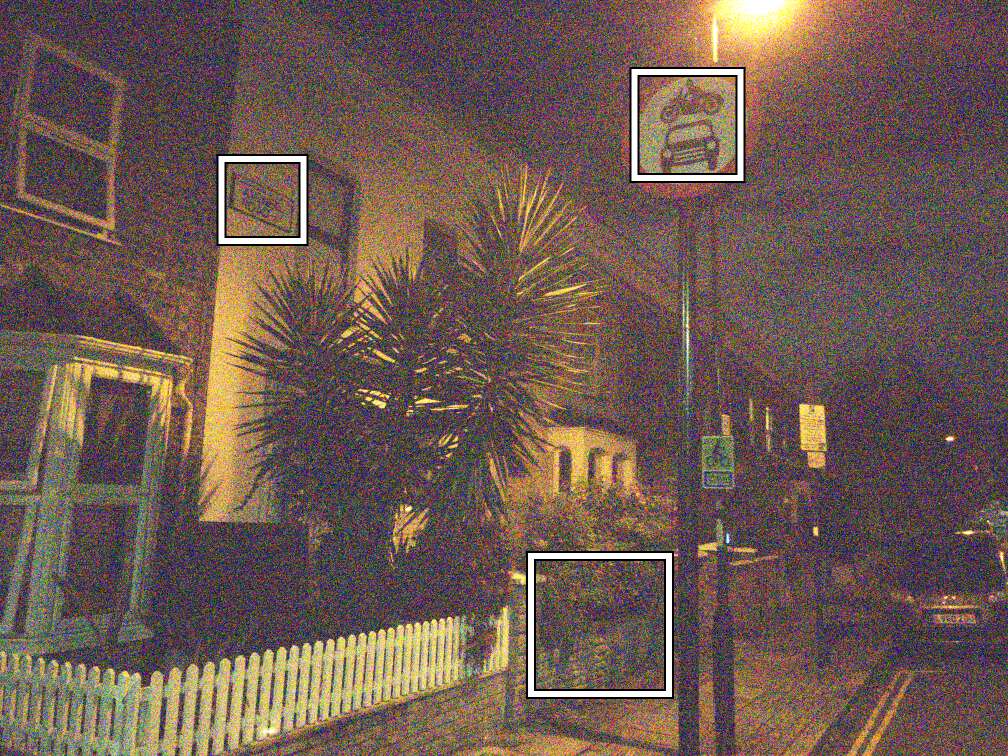} & 
    \begin{tabular}[b]{@{}c@{}}
    \includegraphics[height=\tableauinheight]{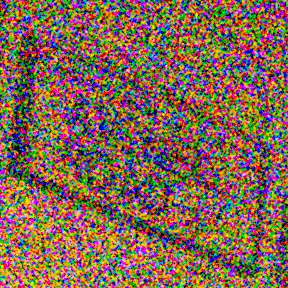}\\
    \includegraphics[height=\tableauinheight]{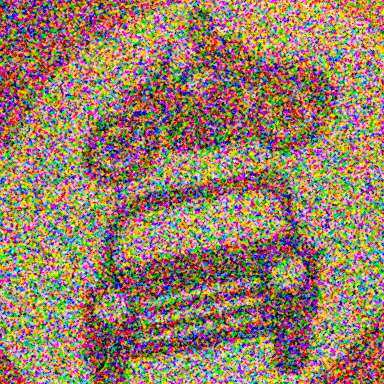}\\
    \includegraphics[height=\tableauinheight]{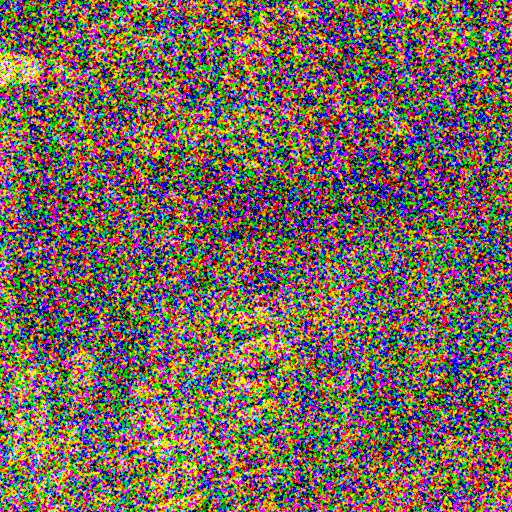}
    \end{tabular} &
    \includegraphics[height=\tableauheight]{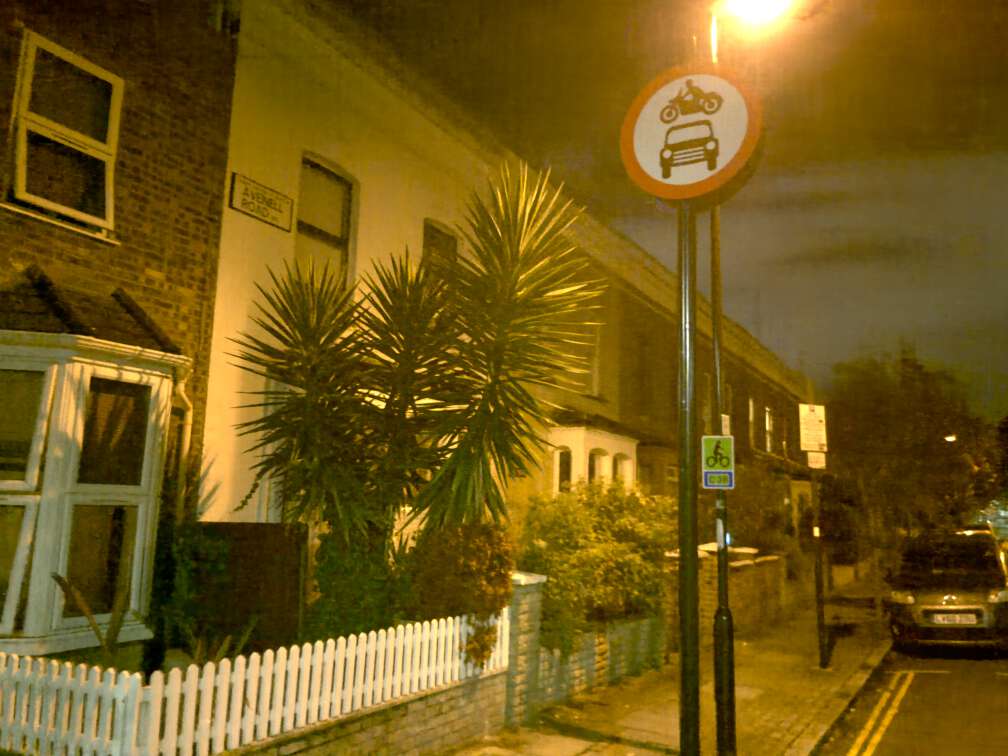} & 
    \begin{tabular}[b]{@{}c@{}}
    \includegraphics[height=\tableauinheight]{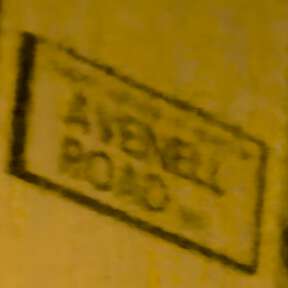}\\
    \includegraphics[height=\tableauinheight]{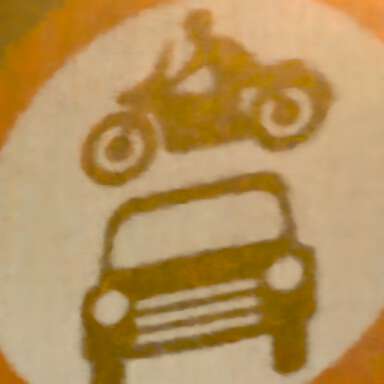}\\
    \includegraphics[height=\tableauinheight]{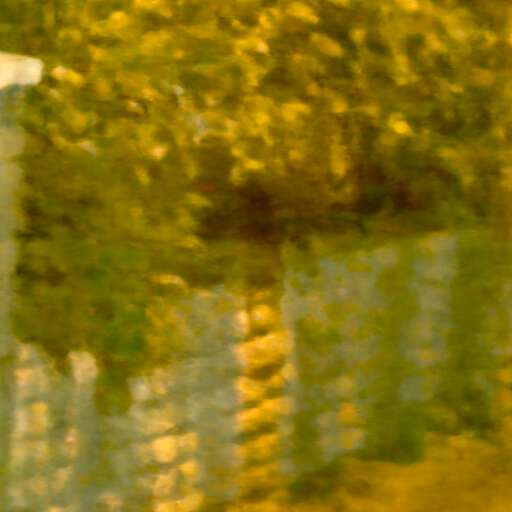}
    \end{tabular} &
    \includegraphics[height=\tableauheight]{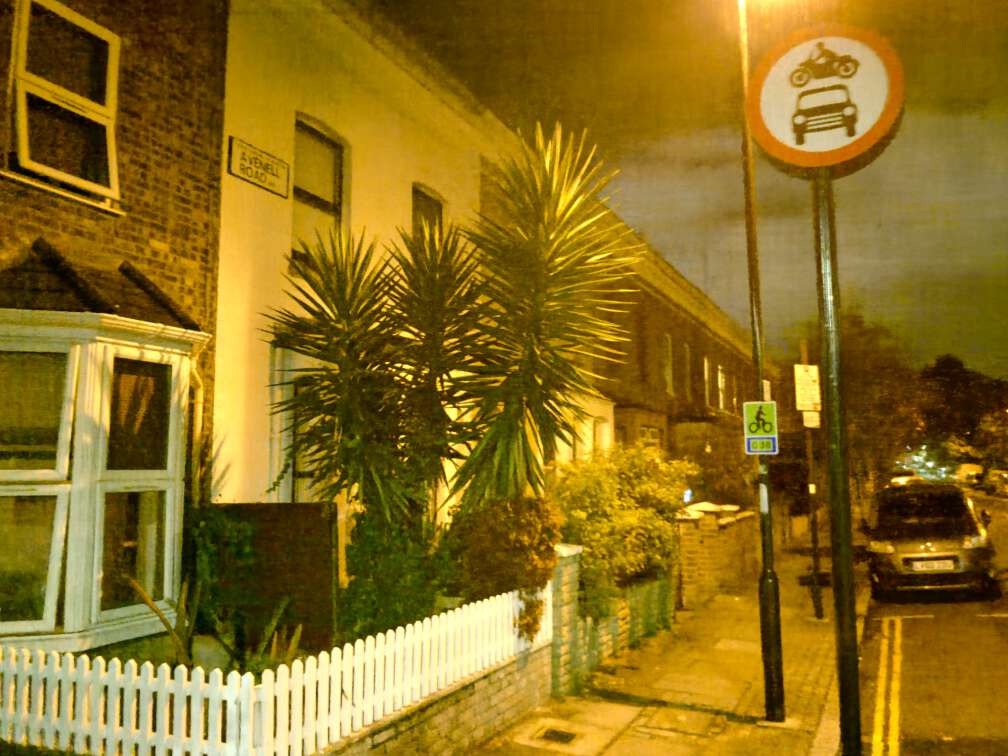} \\ 
    \includegraphics[height=\tableauheight]{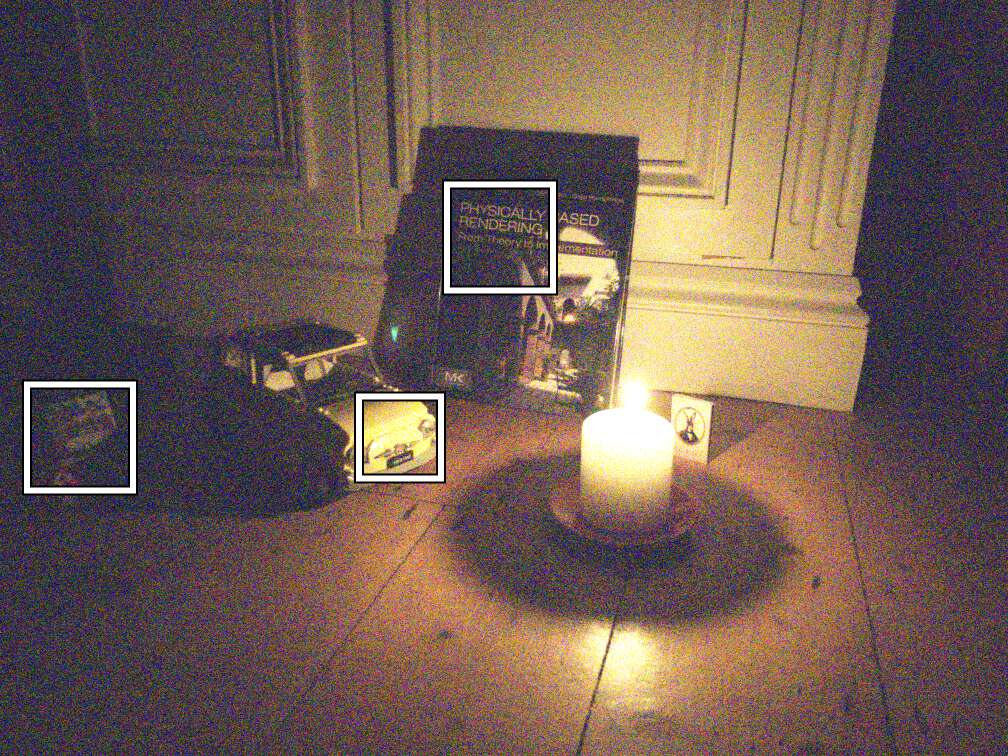} & 
    \begin{tabular}[b]{@{}c@{}}
    \includegraphics[height=\tableauinheight]{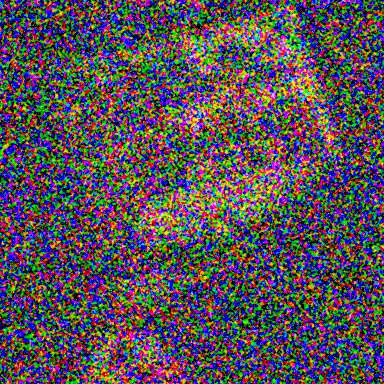}\\
    \includegraphics[height=\tableauinheight]{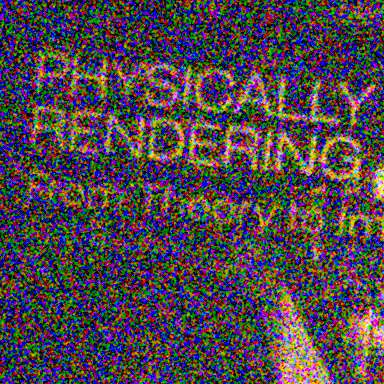}\\
    \includegraphics[height=\tableauinheight]{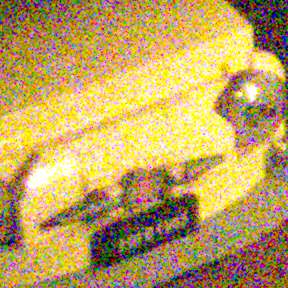}
    \end{tabular} &
    \includegraphics[height=\tableauheight]{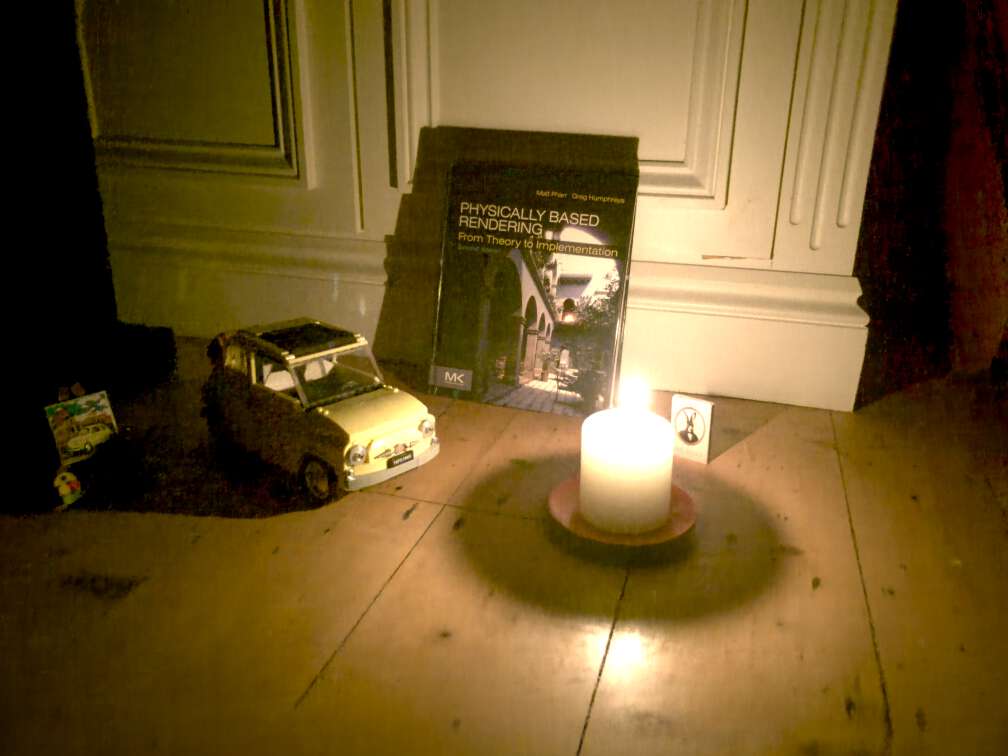} & 
    \begin{tabular}[b]{@{}c@{}}
    \includegraphics[height=\tableauinheight]{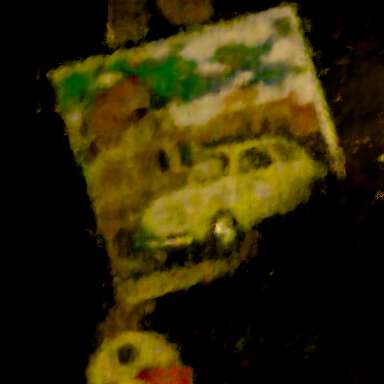}\\
    \includegraphics[height=\tableauinheight]{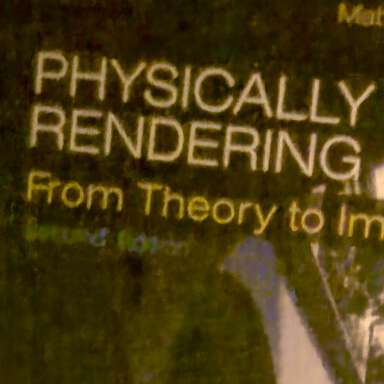}\\
    \includegraphics[height=\tableauinheight]{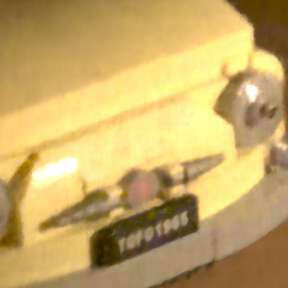}
    \end{tabular} &
    \includegraphics[height=\tableauheight]{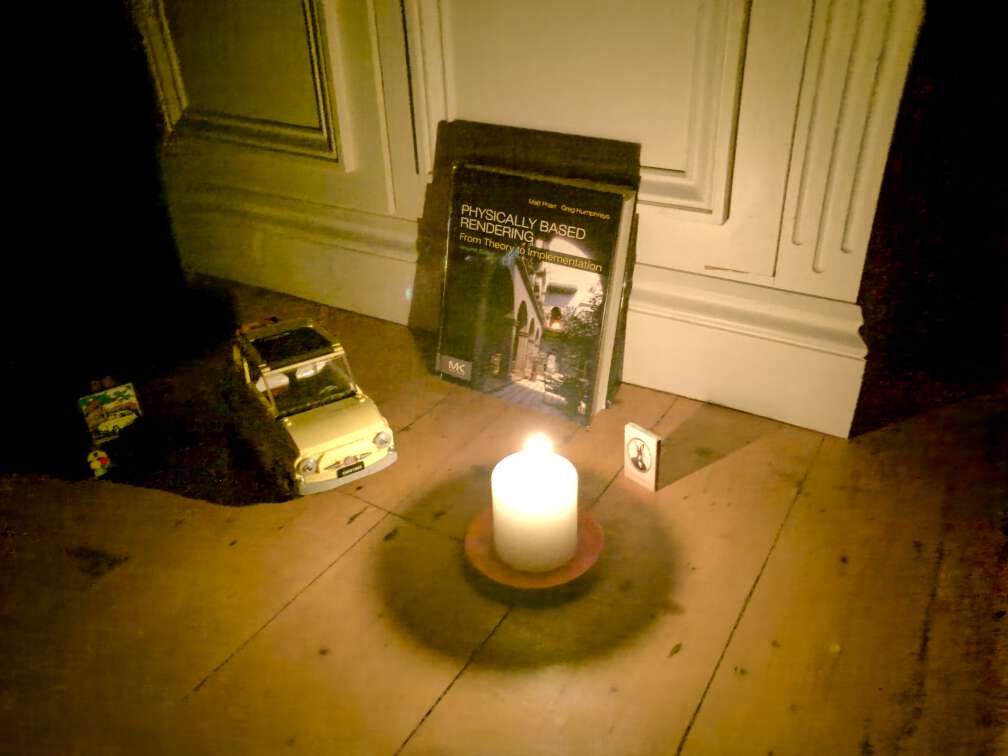} \\ 
    \includegraphics[height=\tableauheight]{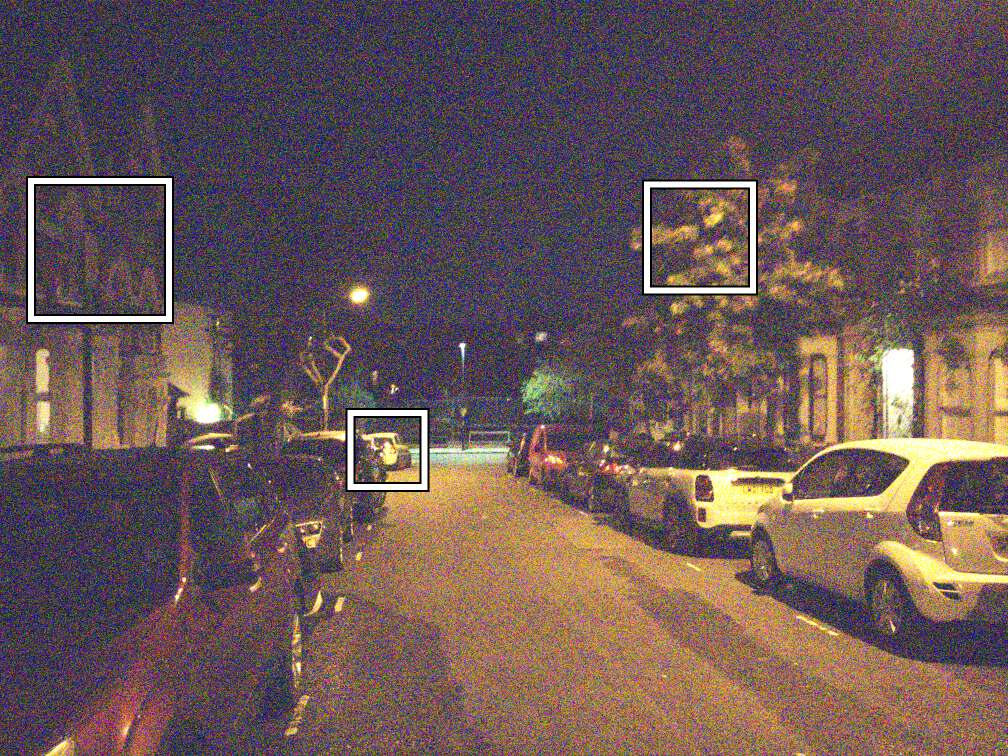} & 
    \begin{tabular}[b]{@{}c@{}}
    \includegraphics[height=\tableauinheight]{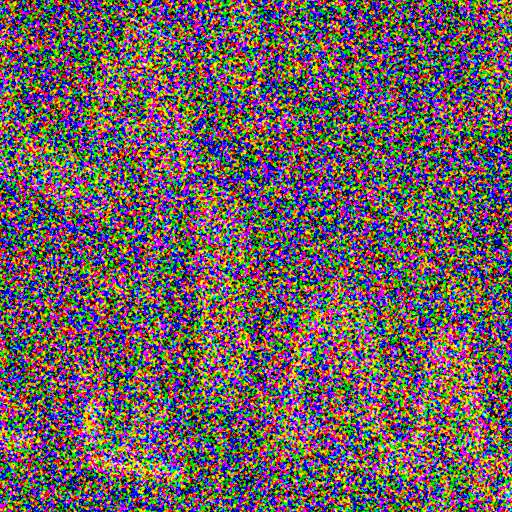}\\
    \includegraphics[height=\tableauinheight]{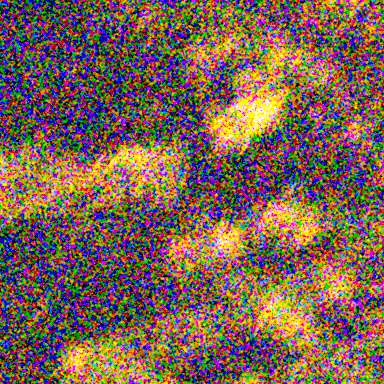}\\
    \includegraphics[height=\tableauinheight]{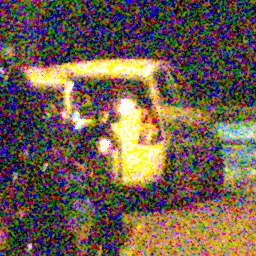}
    \end{tabular} &
    \includegraphics[height=\tableauheight]{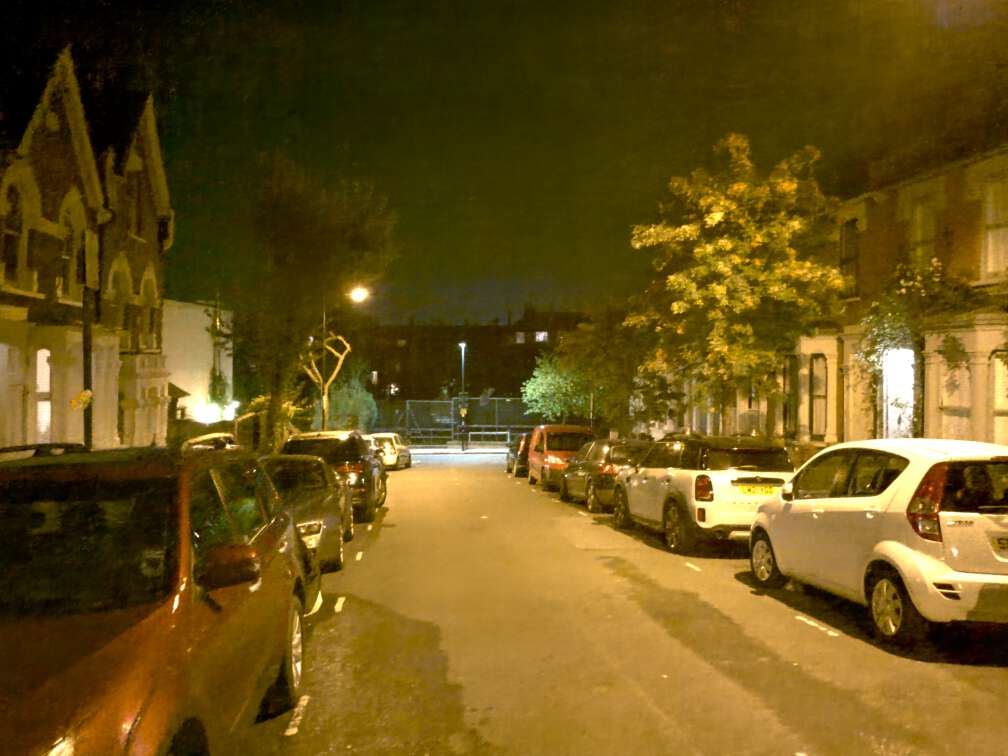} & 
    \begin{tabular}[b]{@{}c@{}}
    \includegraphics[height=\tableauinheight]{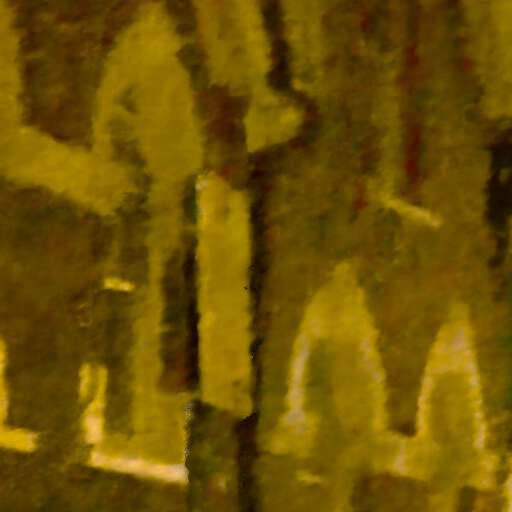}\\
    \includegraphics[height=\tableauinheight]{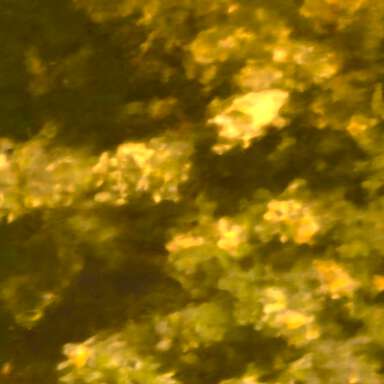}\\
    \includegraphics[height=\tableauinheight]{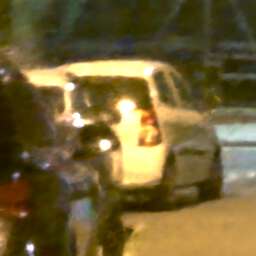}
    \end{tabular} &
    \includegraphics[height=\tableauheight]{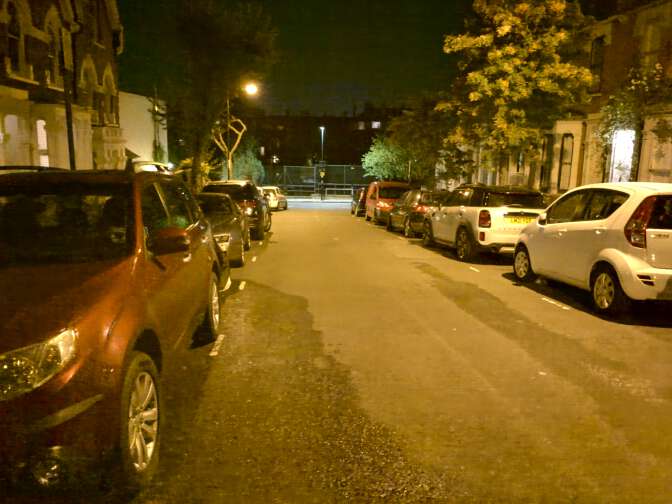} \\ 
    \includegraphics[height=\tableauheight]{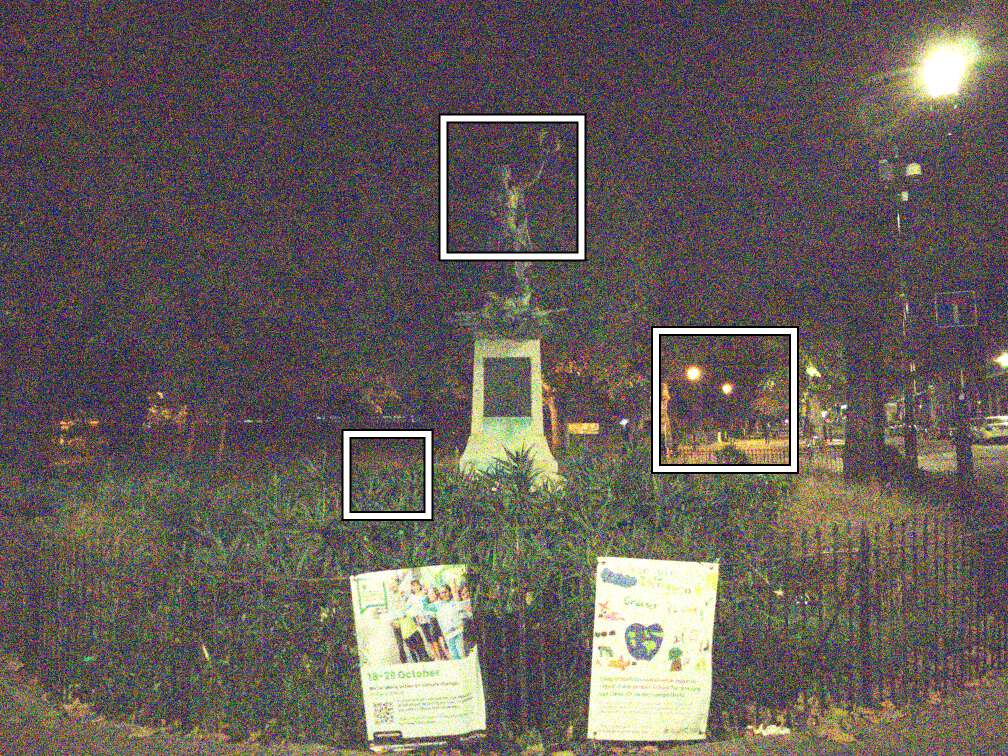} & 
    \begin{tabular}[b]{@{}c@{}}
    \includegraphics[height=\tableauinheight]{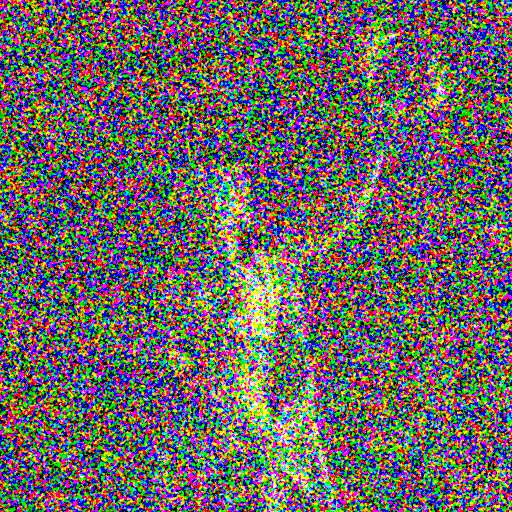}\\
    \includegraphics[height=\tableauinheight]{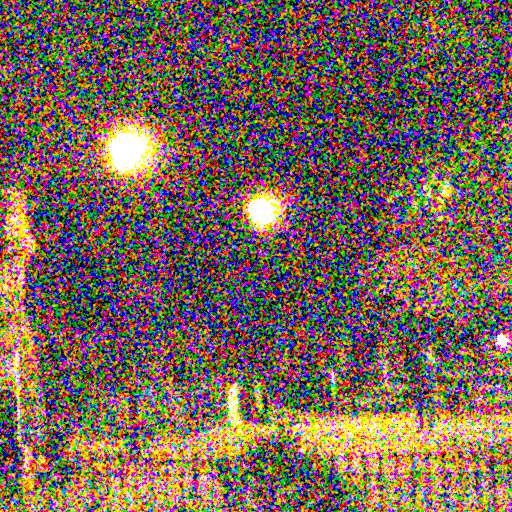}\\
    \includegraphics[height=\tableauinheight]{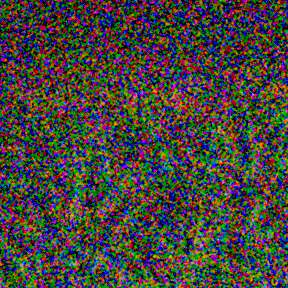}
    \end{tabular} &
    \includegraphics[height=\tableauheight]{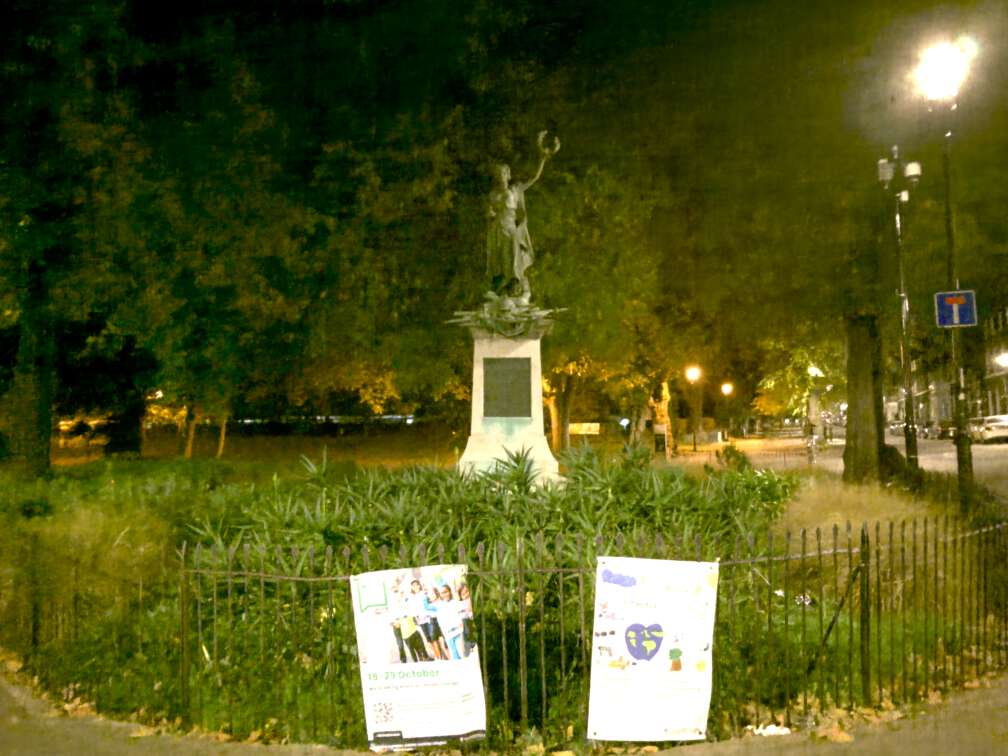} & 
    \begin{tabular}[b]{@{}c@{}}
    \includegraphics[height=\tableauinheight]{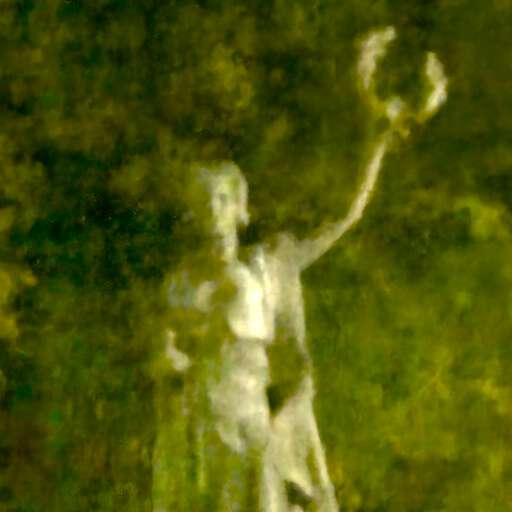}\\
    \includegraphics[height=\tableauinheight]{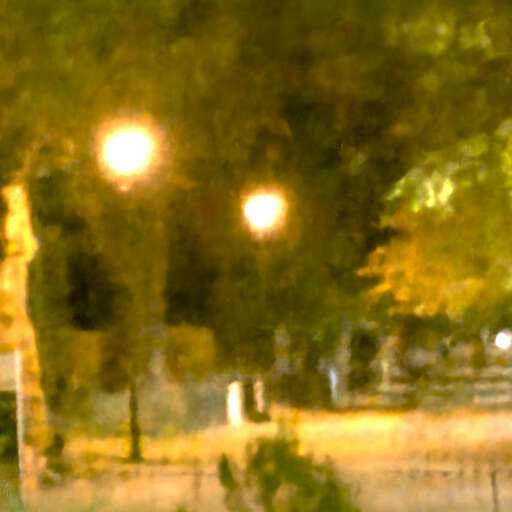}\\
    \includegraphics[height=\tableauinheight]{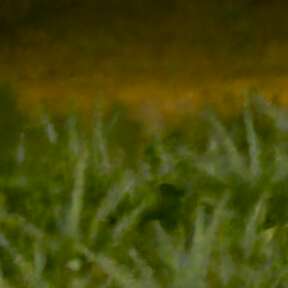}
    \end{tabular} &
    \includegraphics[height=\tableauheight]{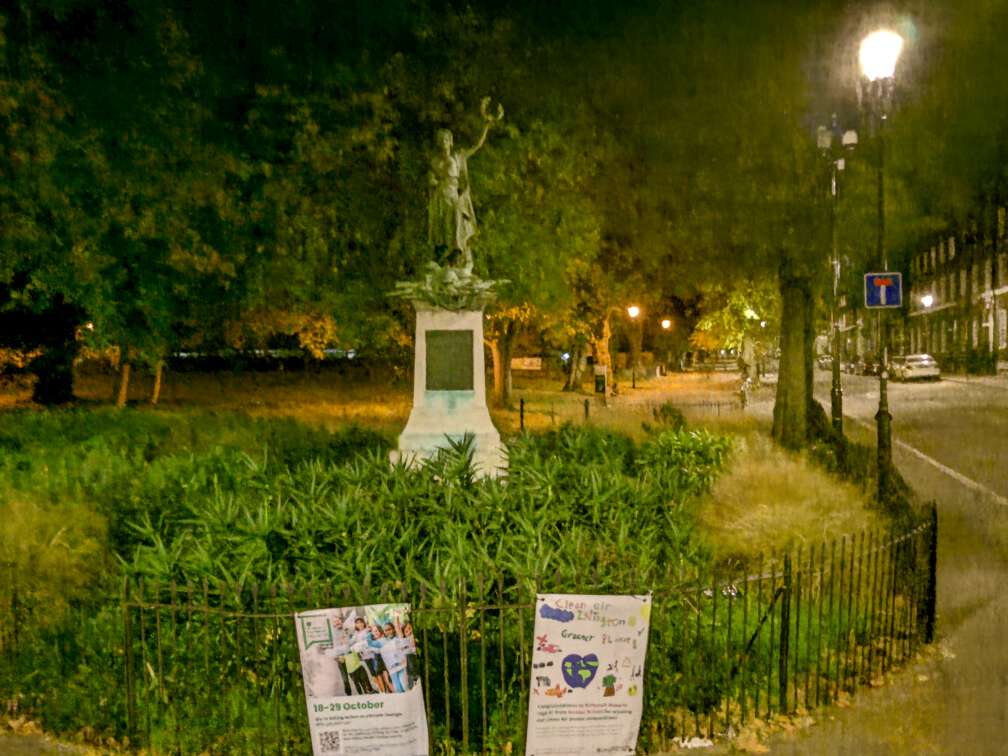} \\ 
    \includegraphics[height=\tableauheight]{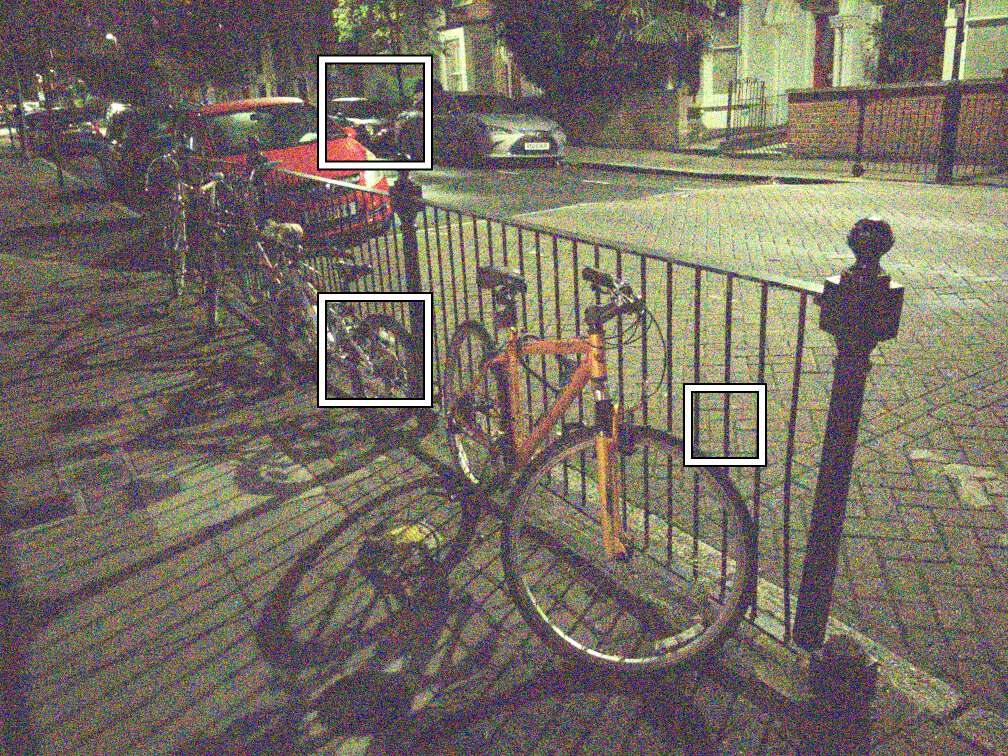} & 
    \begin{tabular}[b]{@{}c@{}}
    \includegraphics[height=\tableauinheight]{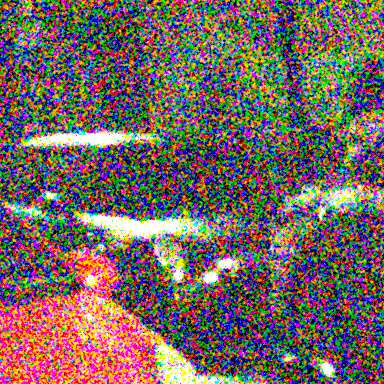}\\
    \includegraphics[height=\tableauinheight]{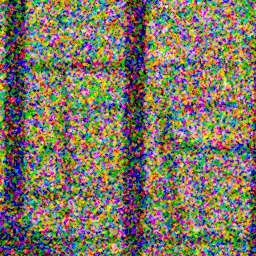}\\
    \includegraphics[height=\tableauinheight]{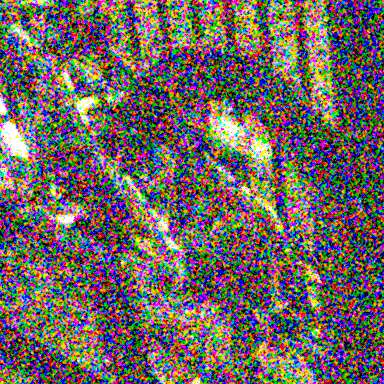}
    \end{tabular} &
    \includegraphics[height=\tableauheight]{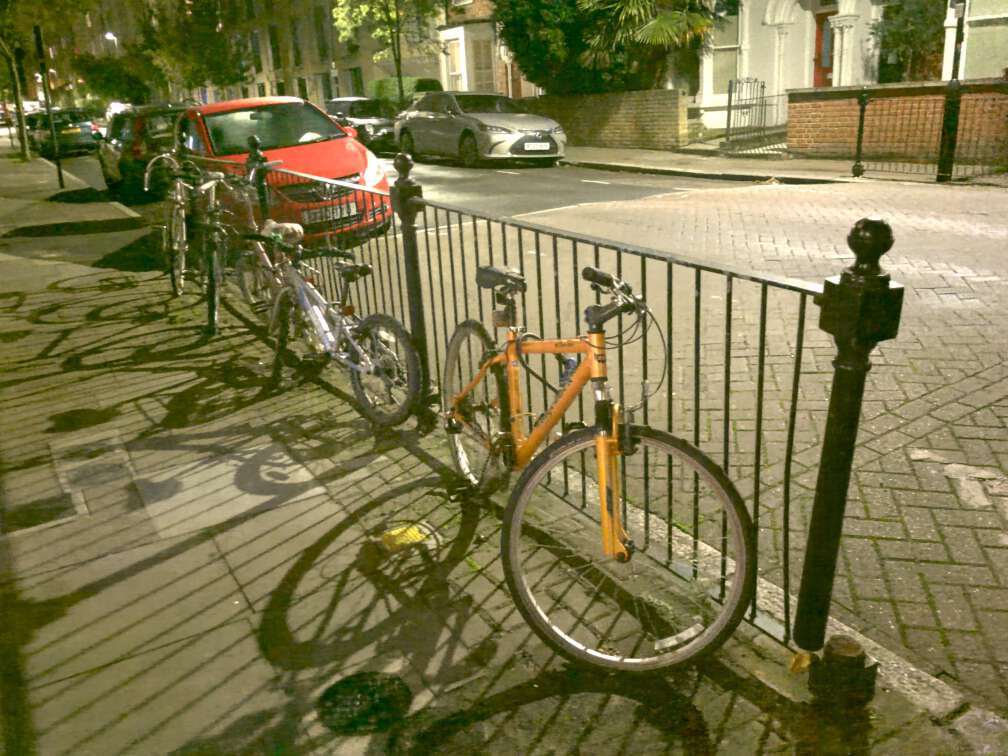} & 
    \begin{tabular}[b]{@{}c@{}}
    \includegraphics[height=\tableauinheight]{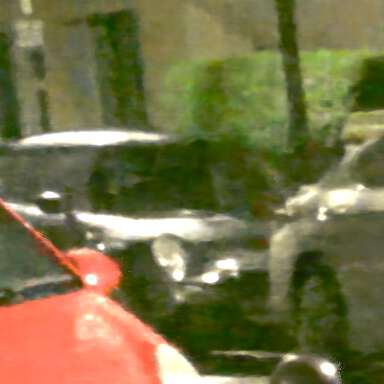}\\
    \includegraphics[height=\tableauinheight]{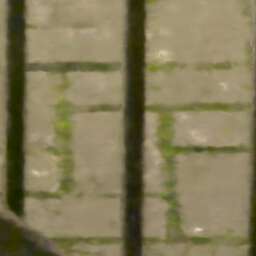}\\
    \includegraphics[height=\tableauinheight]{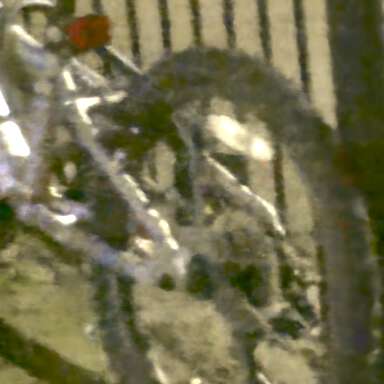}
    \end{tabular} &
    \includegraphics[height=\tableauheight]{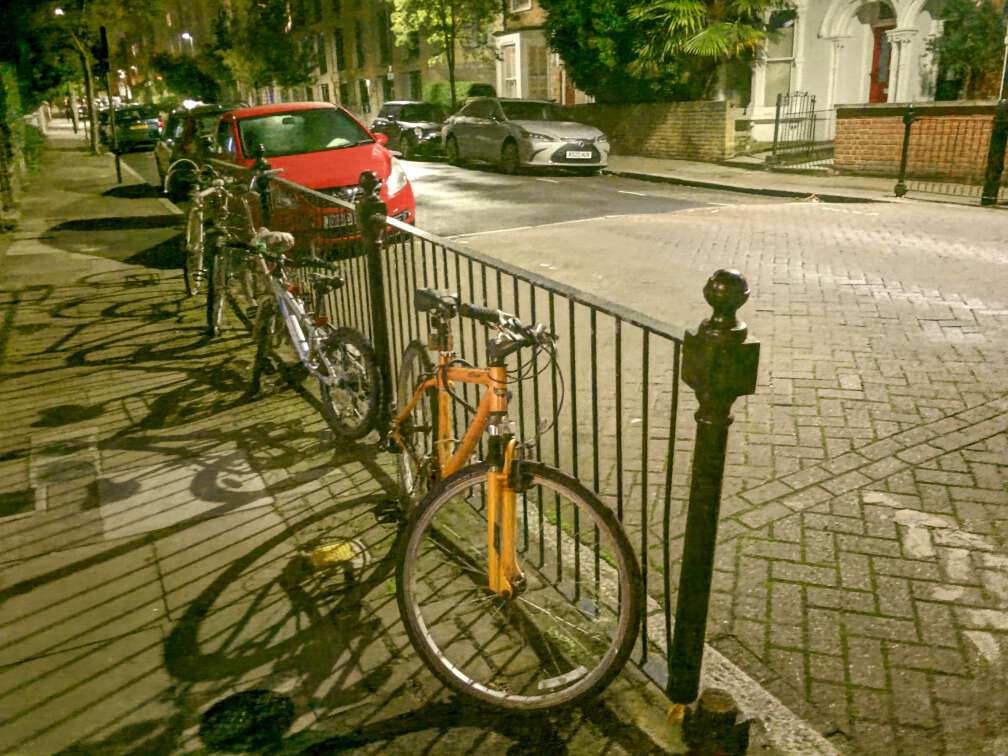} \\ 
    \multicolumn{2}{c}{Noisy test image} & 
    \multicolumn{2}{c}{RawNeRF rendering} &
    New viewpoint, HDR tonemapping
    \end{tabular}
    }
    \caption{RawNeRF in the dark.}
    \label{fig:tableau_dark}
    \vspace{1.5cm}
\end{figure*}

\begin{figure*}[]
    \centering
\resizebox{.965\linewidth}{!}{
    \begin{tabular}{@{}c@{\,}c@{\quad}c@{\,}c@{\quad}c@{}}
    \includegraphics[height=\tableauheight]{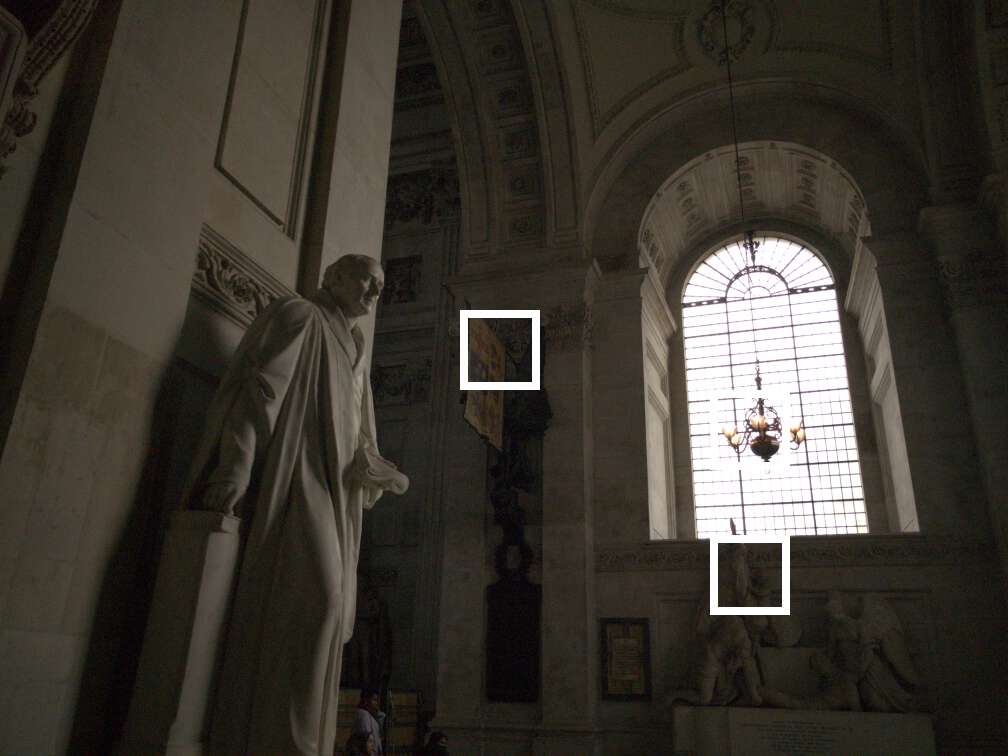} & 
    \begin{tabular}[b]{@{}c@{}}
    \includegraphics[height=\tableauinheight]{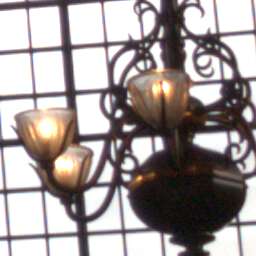}\\
    \includegraphics[height=\tableauinheight]{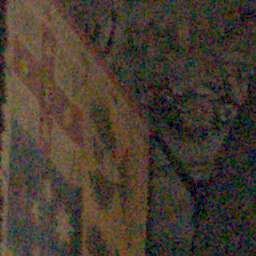}\\
    \includegraphics[height=\tableauinheight]{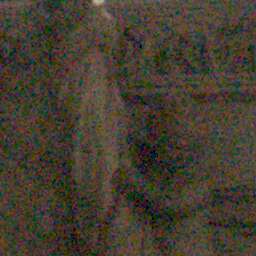}
    \end{tabular} &
    \includegraphics[height=\tableauheight]{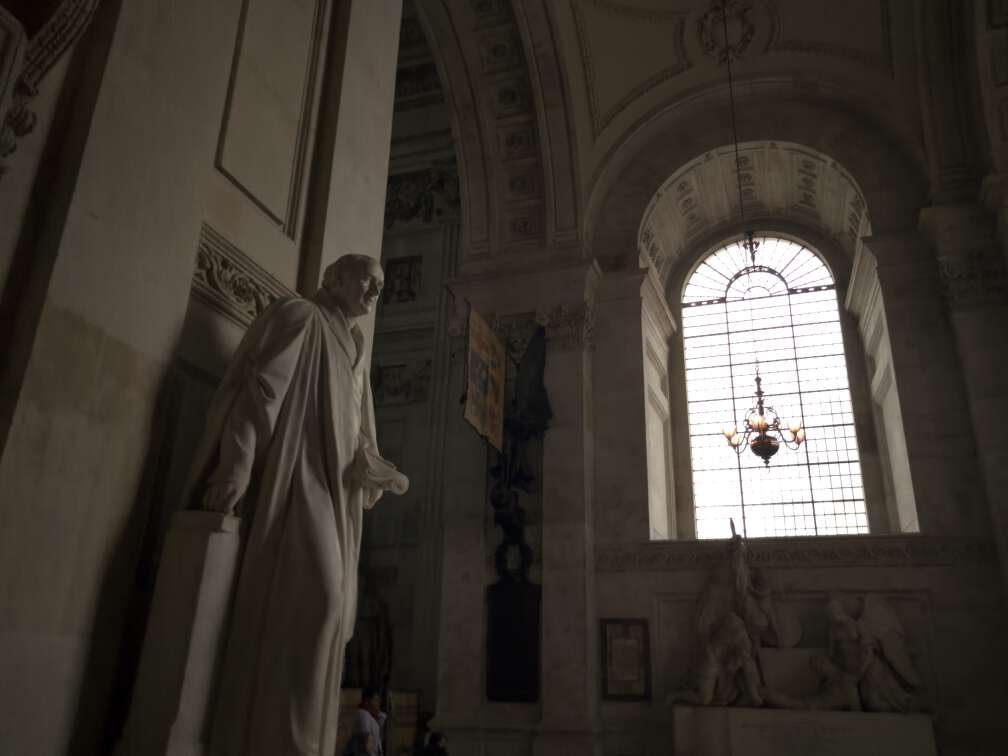} & 
    \begin{tabular}[b]{@{}c@{}}
    \includegraphics[height=\tableauinheight]{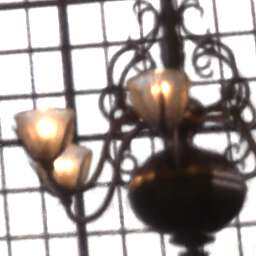}\\
    \includegraphics[height=\tableauinheight]{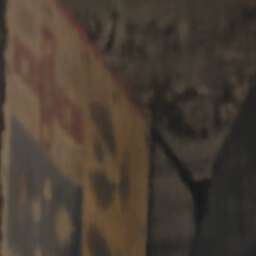}\\
    \includegraphics[height=\tableauinheight]{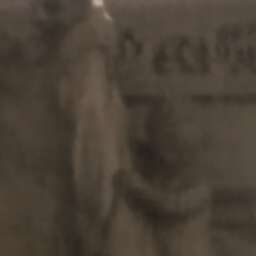}
    \end{tabular} &
    \includegraphics[height=\tableauheight]{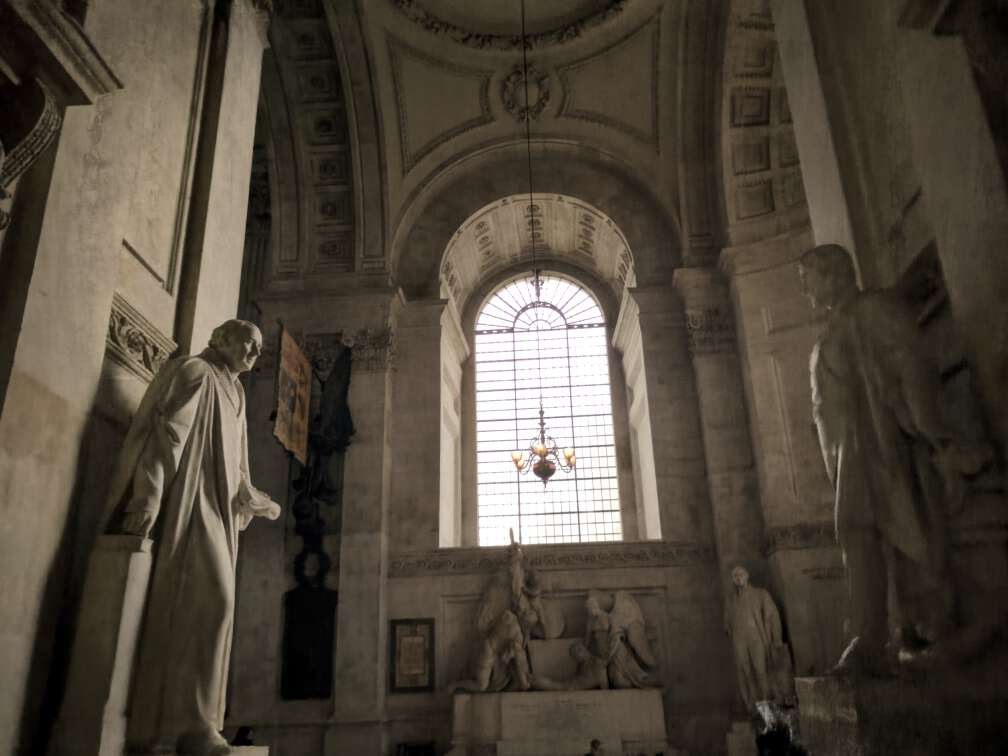} \\ 
    \includegraphics[height=\tableauheight]{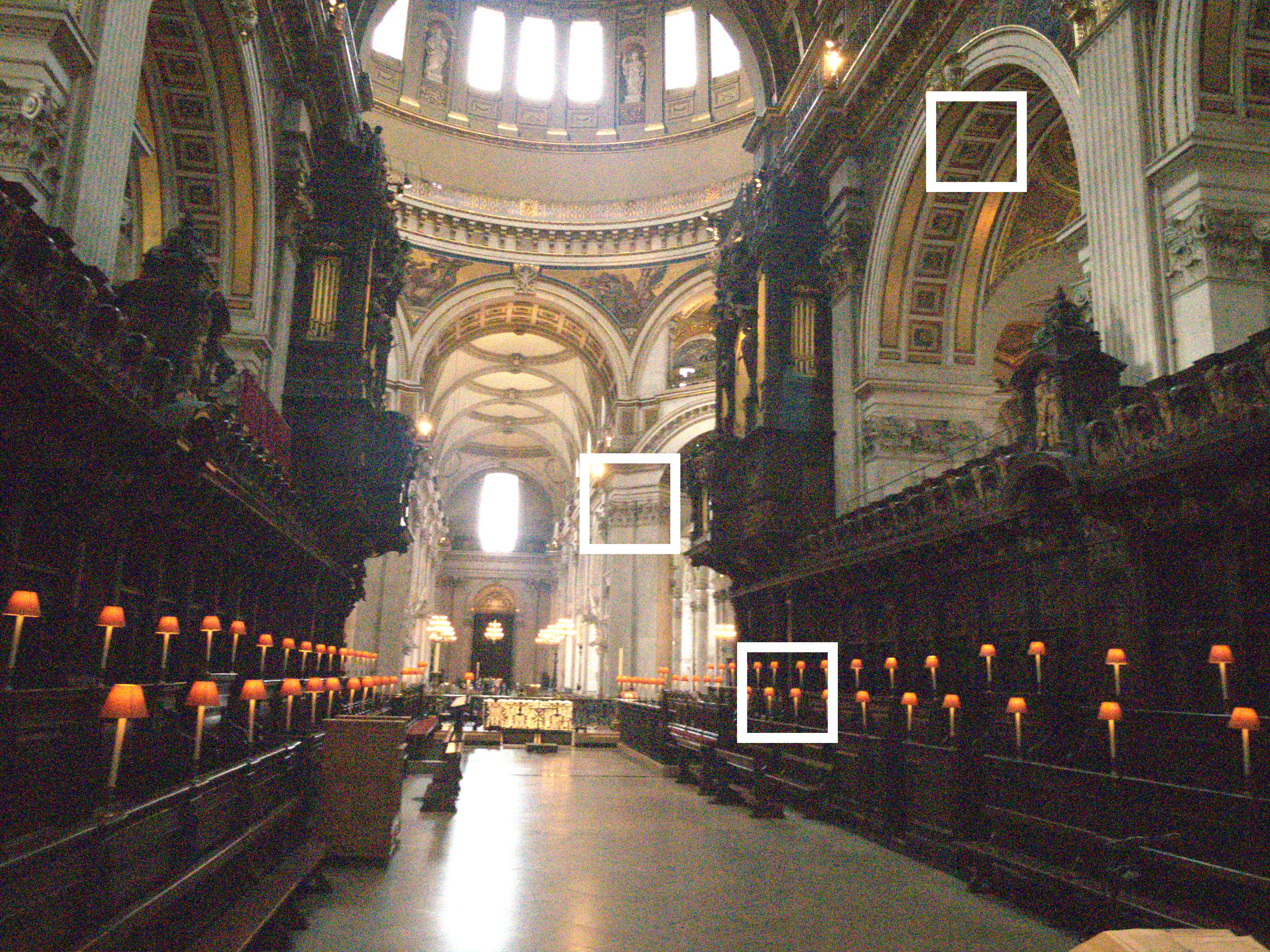} & 
    \begin{tabular}[b]{@{}c@{}}
    \includegraphics[height=\tableauinheight]{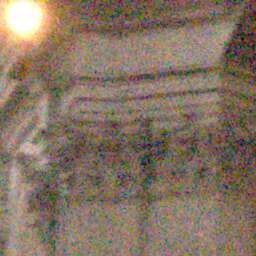}\\
    \includegraphics[height=\tableauinheight]{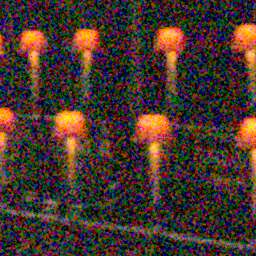}\\
    \includegraphics[height=\tableauinheight]{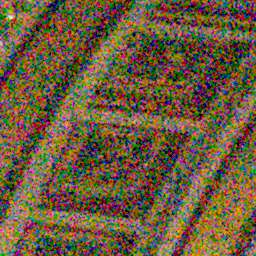}
    \end{tabular} &
    \includegraphics[height=\tableauheight]{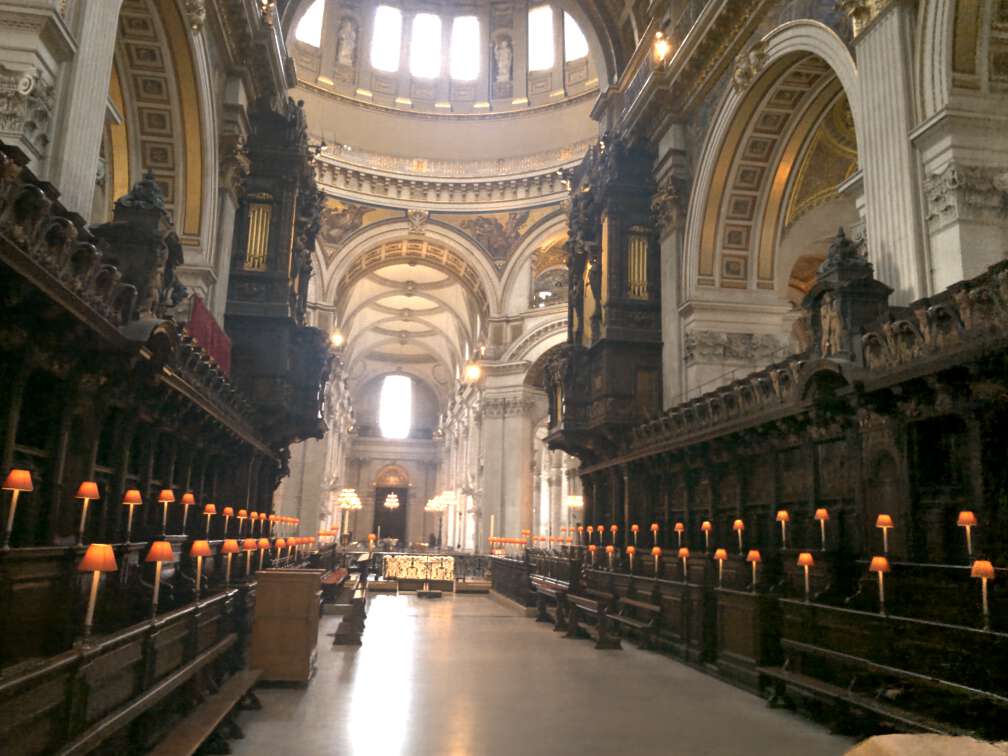} & 
    \begin{tabular}[b]{@{}c@{}}
    \includegraphics[height=\tableauinheight]{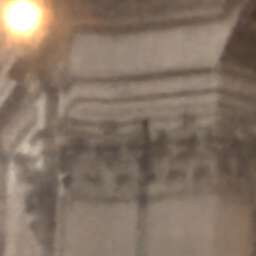}\\
    \includegraphics[height=\tableauinheight]{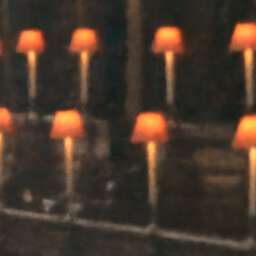}\\
    \includegraphics[height=\tableauinheight]{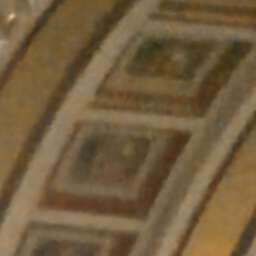}
    \end{tabular} &
    \includegraphics[height=\tableauheight]{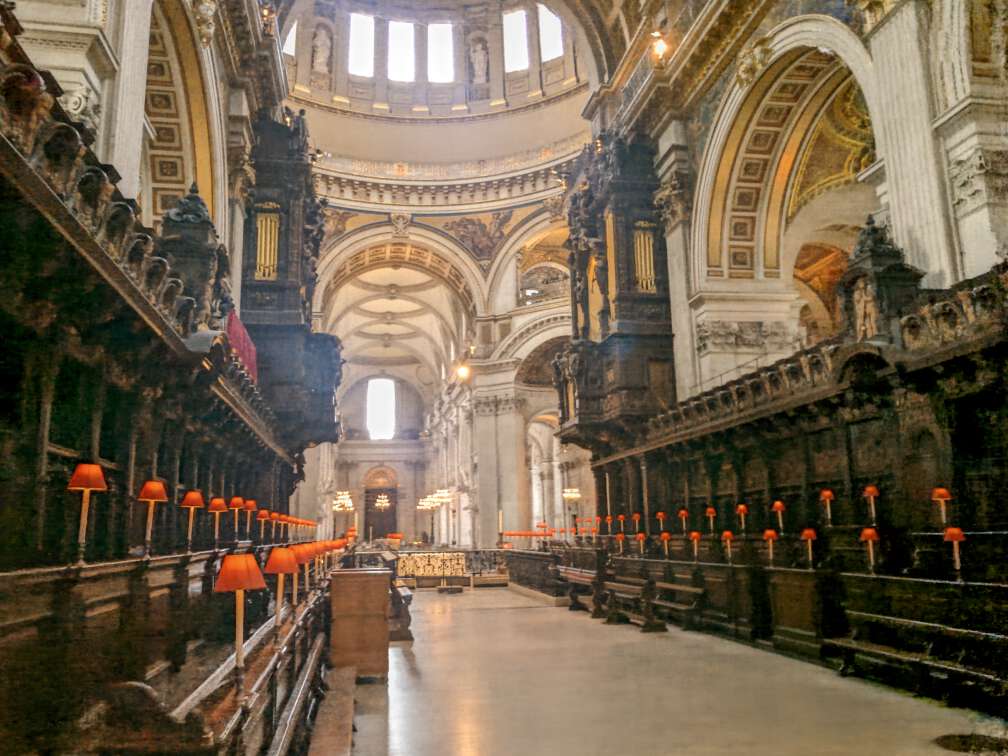} \\ 
    \includegraphics[height=\tableauheight]{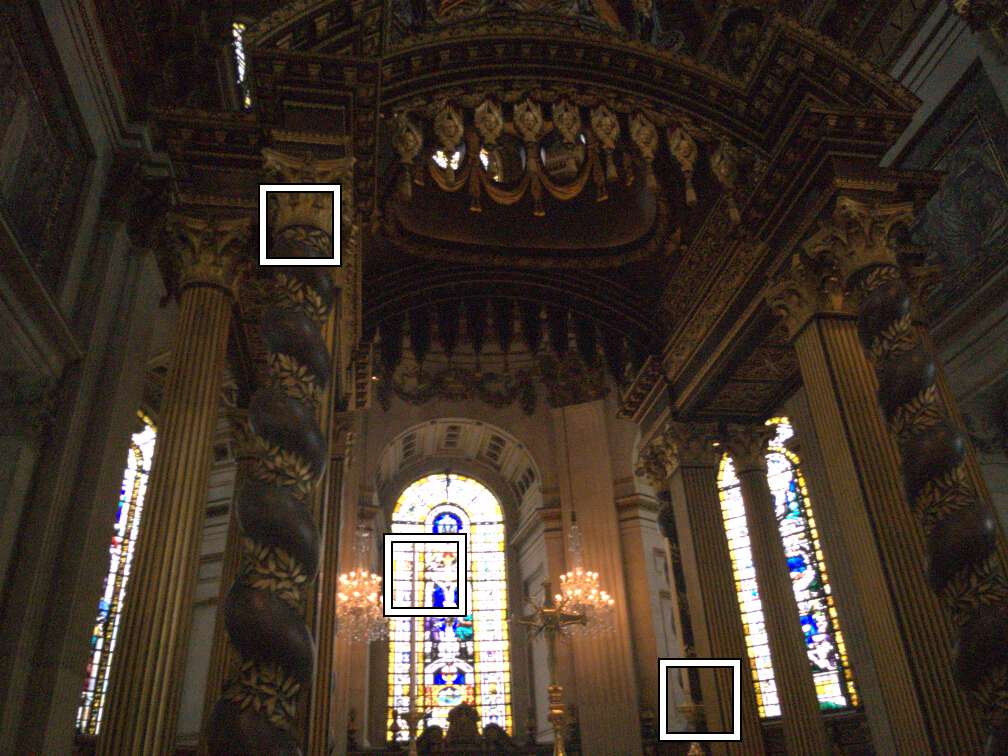} & 
    \begin{tabular}[b]{@{}c@{}}
    \includegraphics[height=\tableauinheight]{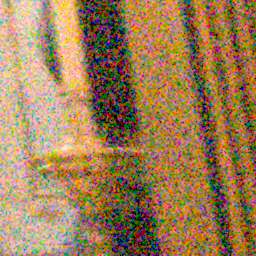}\\
    \includegraphics[height=\tableauinheight]{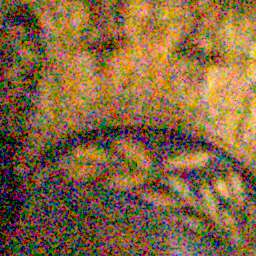}\\
    \includegraphics[height=\tableauinheight]{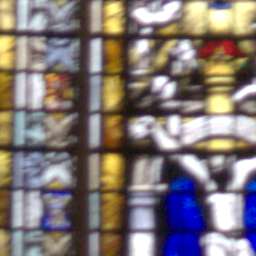}
    \end{tabular} &
    \includegraphics[height=\tableauheight]{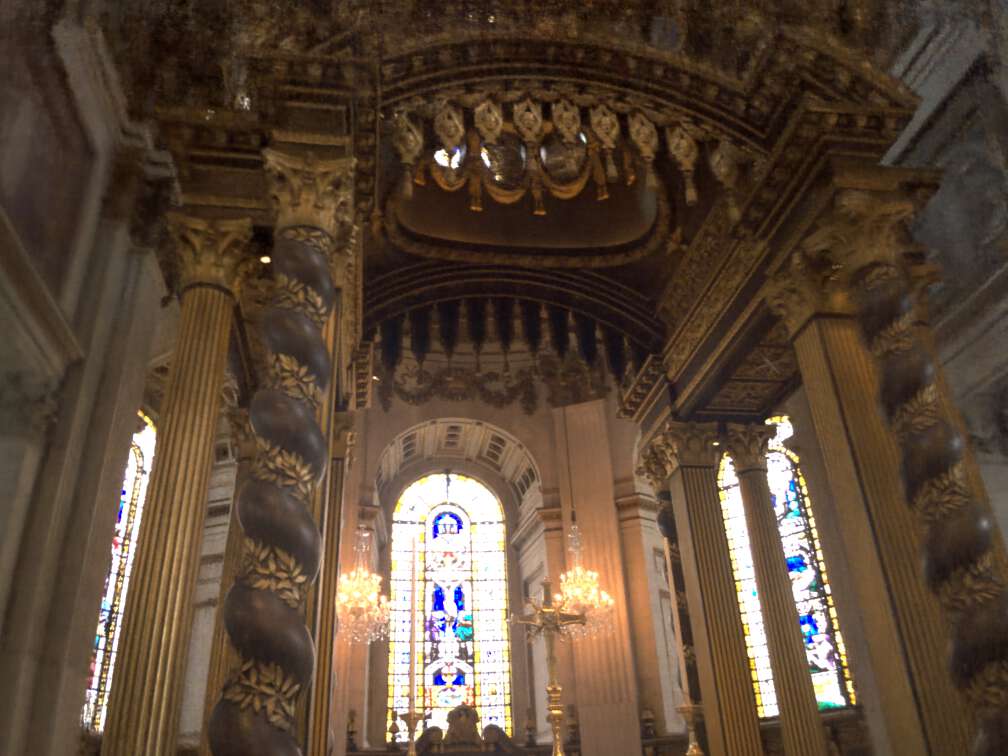} & 
    \begin{tabular}[b]{@{}c@{}}
    \includegraphics[height=\tableauinheight]{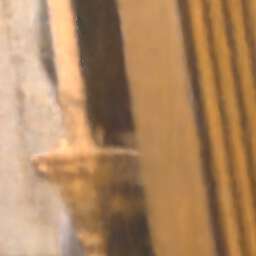}\\
    \includegraphics[height=\tableauinheight]{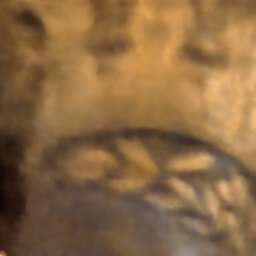}\\
    \includegraphics[height=\tableauinheight]{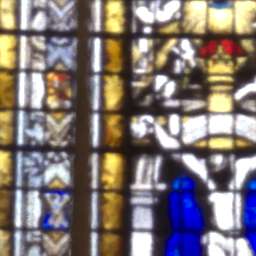}
    \end{tabular} &
    \includegraphics[height=\tableauheight]{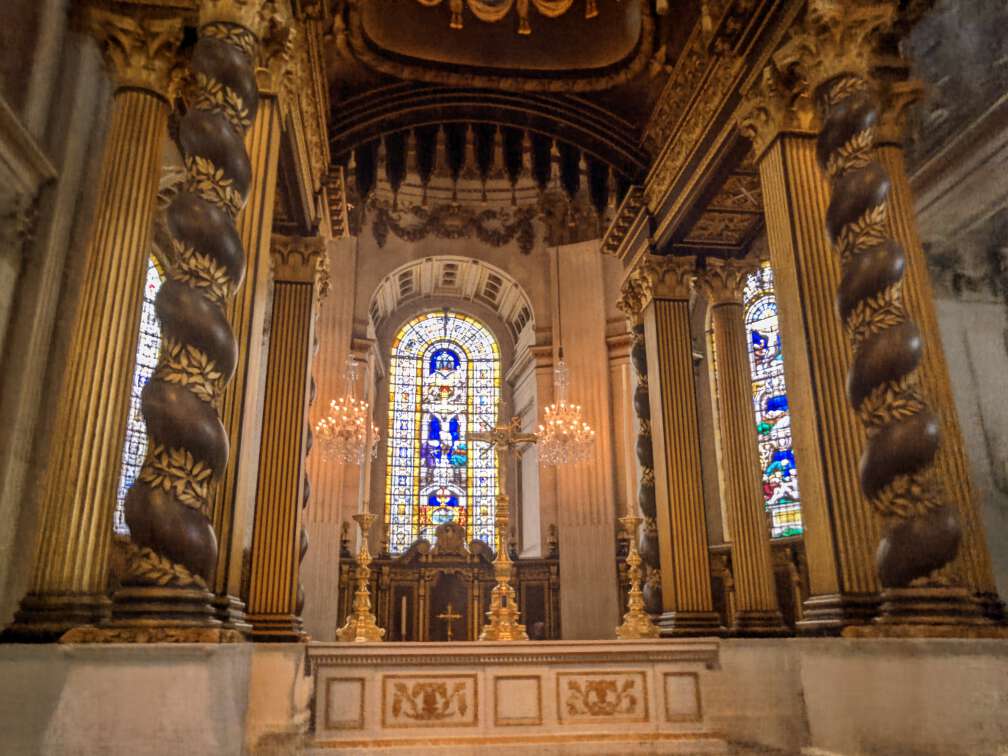} \\ 
    \includegraphics[height=\tableauheight]{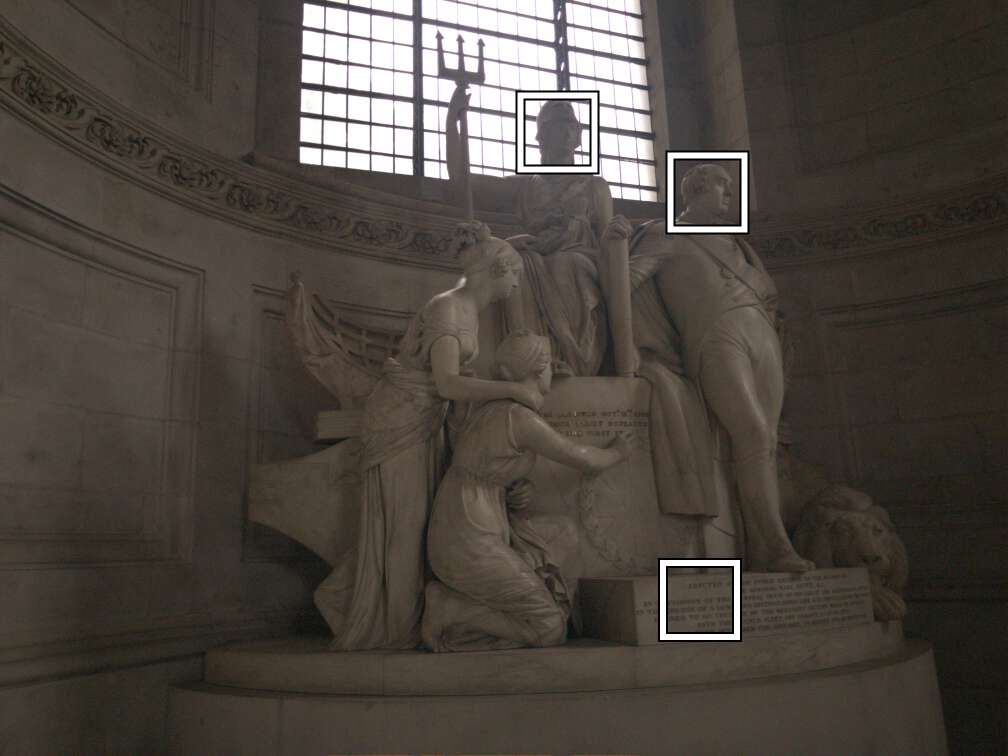} & 
    \begin{tabular}[b]{@{}c@{}}
    \includegraphics[height=\tableauinheight]{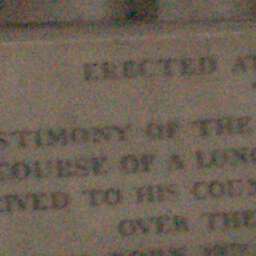}\\
    \includegraphics[height=\tableauinheight]{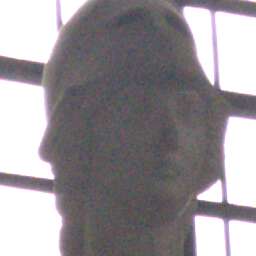}\\
    \includegraphics[height=\tableauinheight]{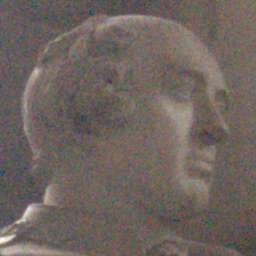}
    \end{tabular} &
    \includegraphics[height=\tableauheight]{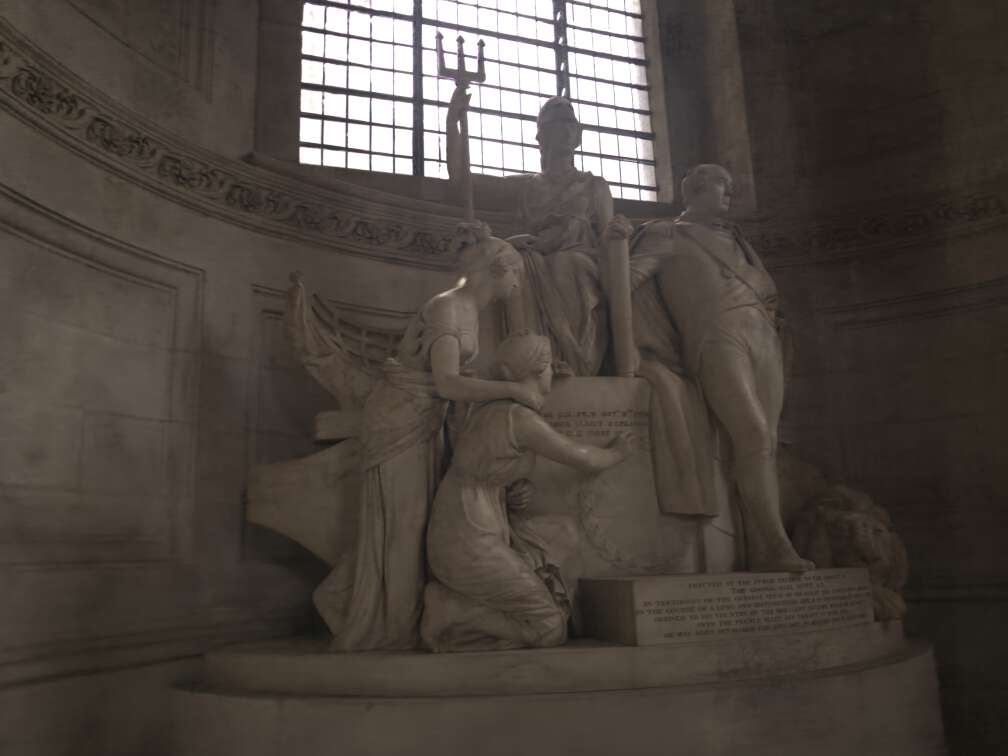} & 
    \begin{tabular}[b]{@{}c@{}}
    \includegraphics[height=\tableauinheight]{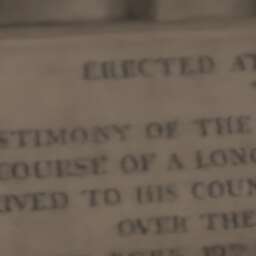}\\
    \includegraphics[height=\tableauinheight]{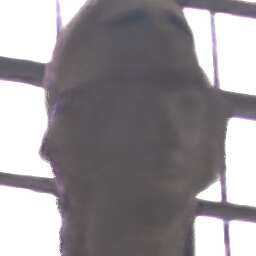}\\
    \includegraphics[height=\tableauinheight]{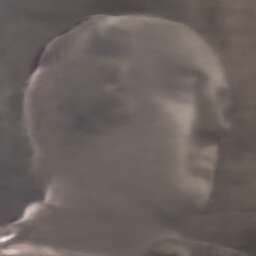}
    \end{tabular} &
    \includegraphics[height=\tableauheight]{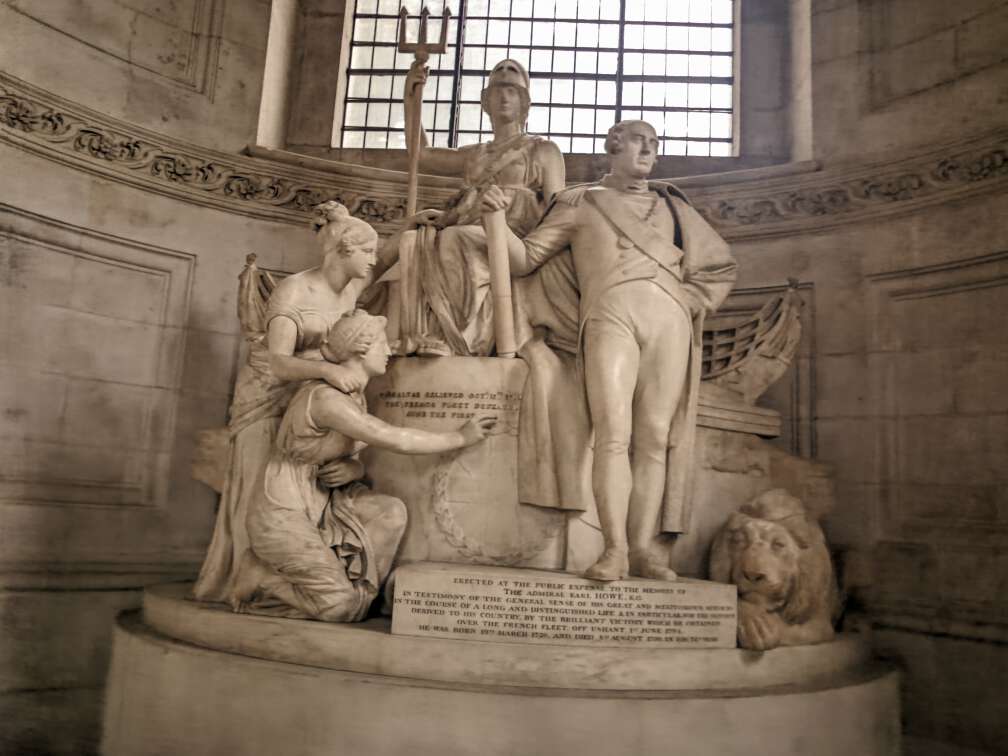} \\ 
    \includegraphics[height=\tableauheight]{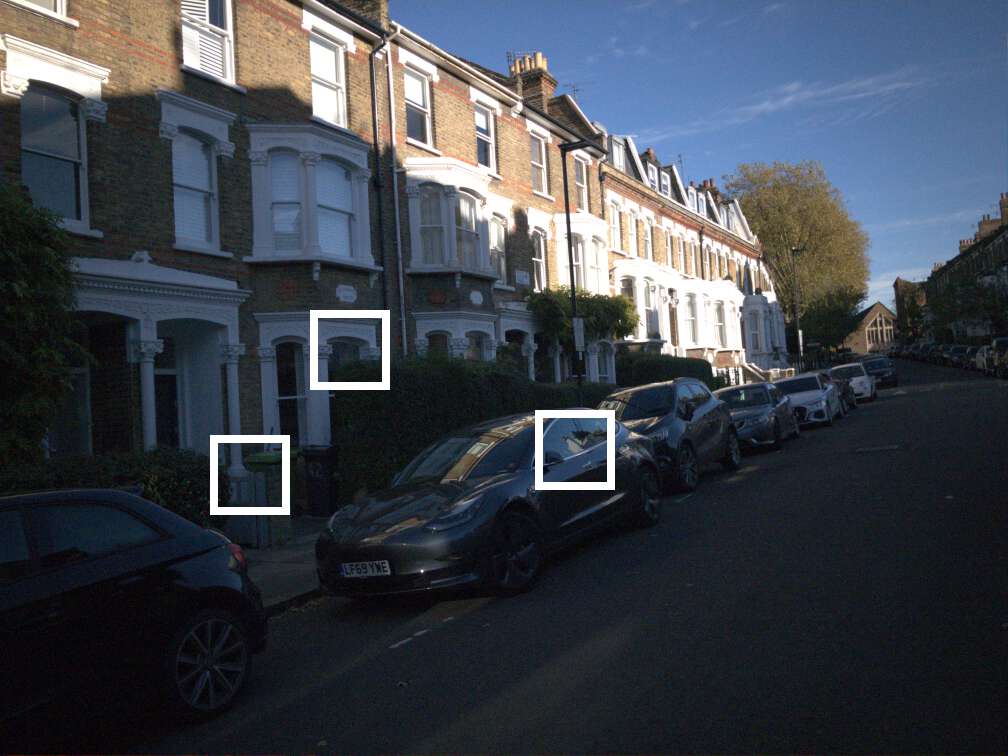} & 
    \begin{tabular}[b]{@{}c@{}}
    \includegraphics[height=\tableauinheight]{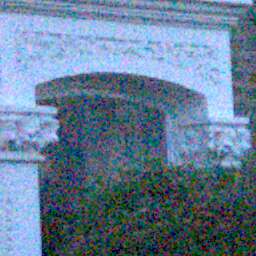}\\
    \includegraphics[height=\tableauinheight]{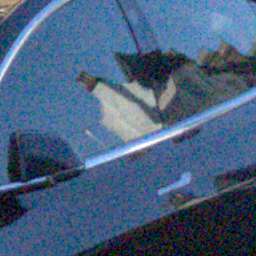}\\
    \includegraphics[height=\tableauinheight]{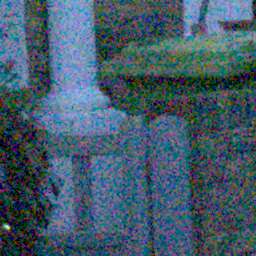}
    \end{tabular} &
    \includegraphics[height=\tableauheight]{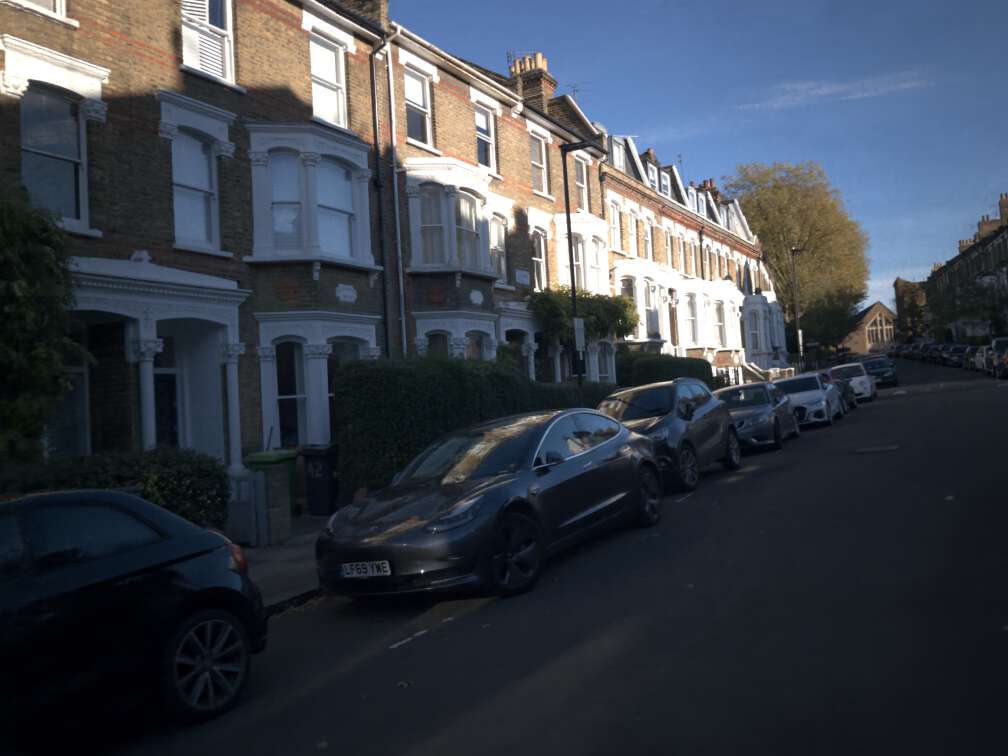} & 
    \begin{tabular}[b]{@{}c@{}}
    \includegraphics[height=\tableauinheight]{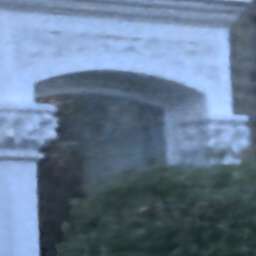}\\
    \includegraphics[height=\tableauinheight]{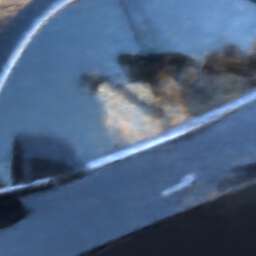}\\
    \includegraphics[height=\tableauinheight]{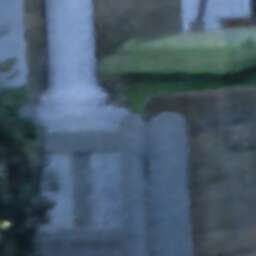}
    \end{tabular} &
    \includegraphics[height=\tableauheight]{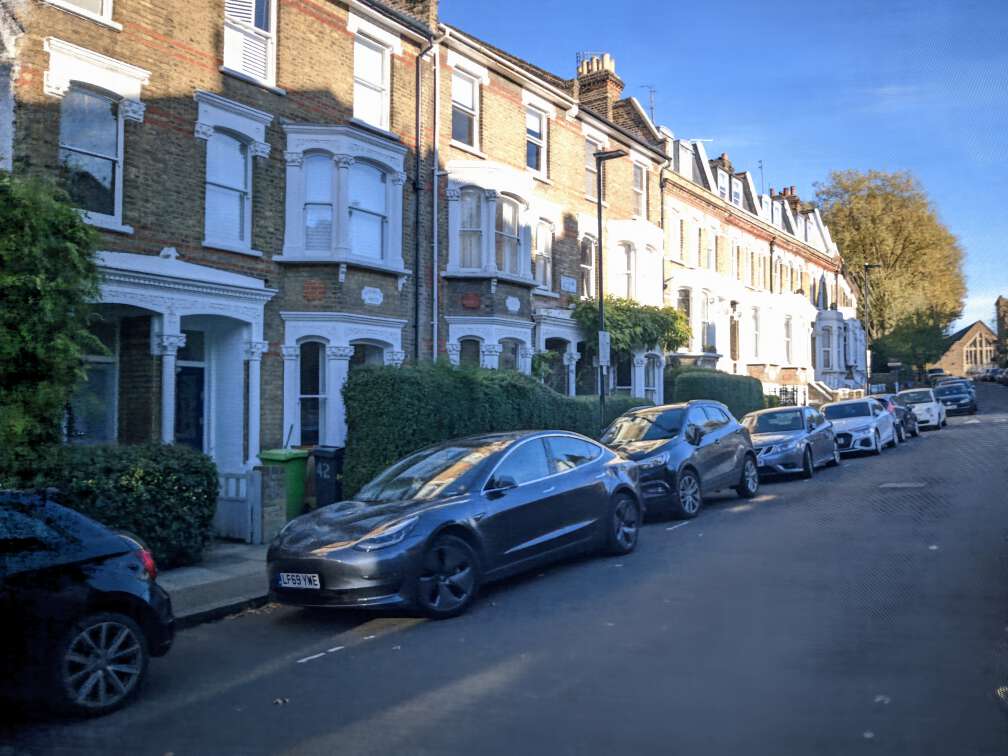} \\ 
    \includegraphics[height=\tableauheight]{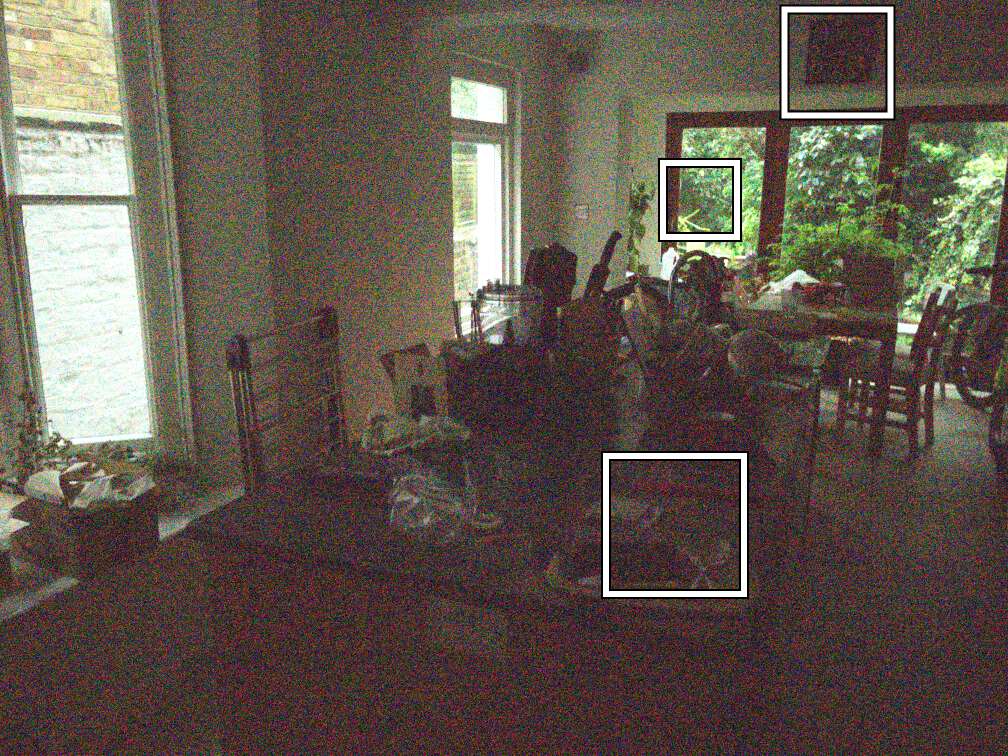} & 
    \begin{tabular}[b]{@{}c@{}}
    \includegraphics[height=\tableauinheight]{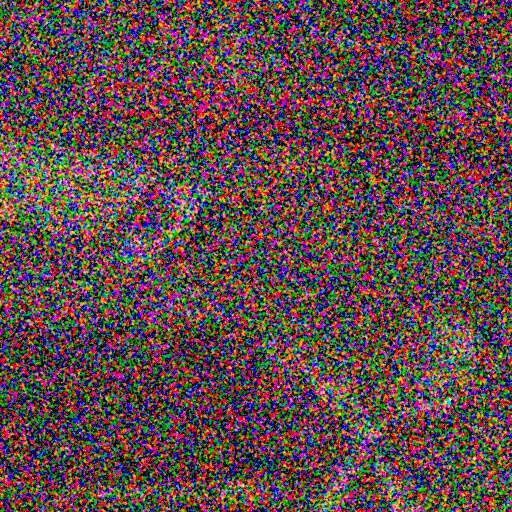}\\
    \includegraphics[height=\tableauinheight]{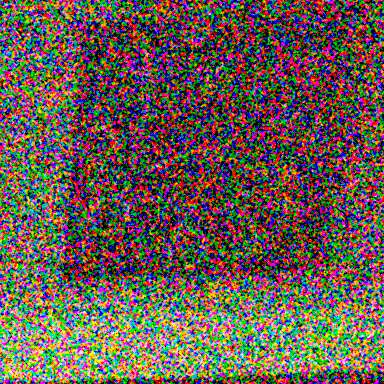}\\
    \includegraphics[height=\tableauinheight]{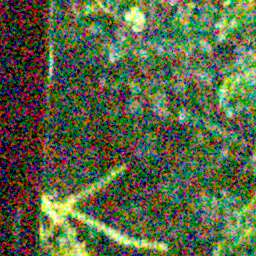}
    \end{tabular} &
    \includegraphics[height=\tableauheight]{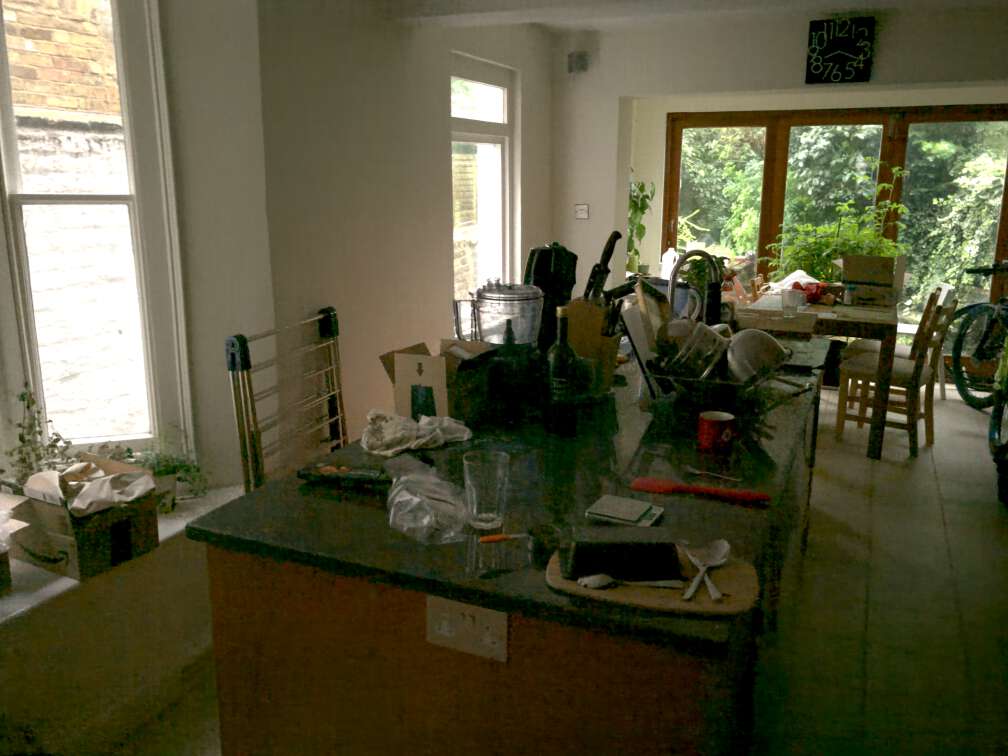} & 
    \begin{tabular}[b]{@{}c@{}}
    \includegraphics[height=\tableauinheight]{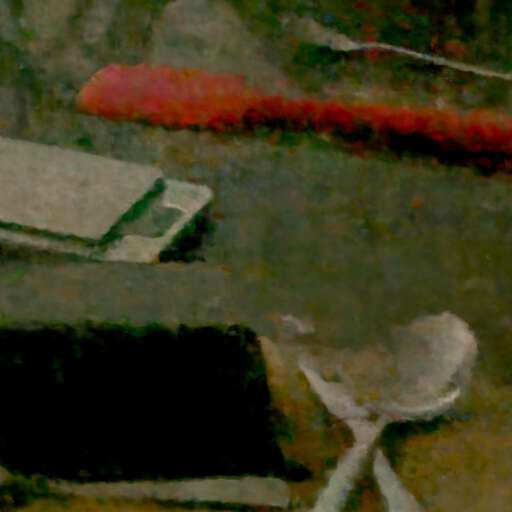}\\
    \includegraphics[height=\tableauinheight]{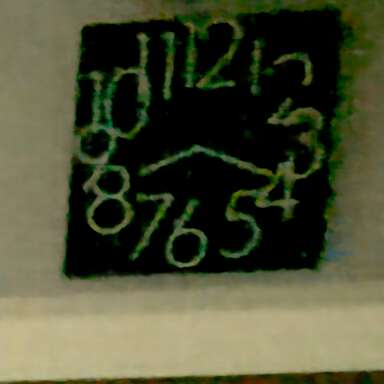}\\
    \includegraphics[height=\tableauinheight]{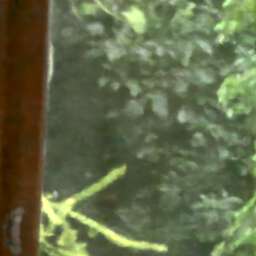}
    \end{tabular} &
    \includegraphics[height=\tableauheight]{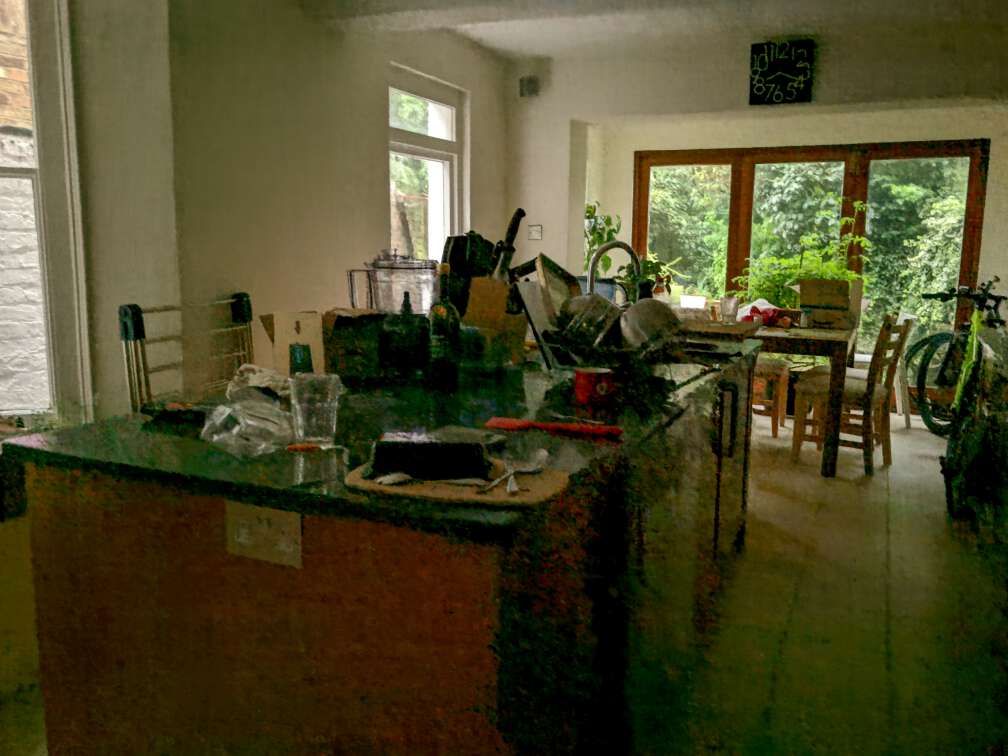} \\ 
    \multicolumn{2}{c}{Noisy test image} & 
    \multicolumn{2}{c}{RawNeRF rendering} &
    New viewpoint, HDR tonemapping
    \end{tabular}
    }
    \caption{Examples of scenes with very high dynamic range.}
    \label{fig:tableau_hdr}
\end{figure*}

\section{Training details}

\subsection{Full derivation of gradient-weighted loss}

We wish to approximate the effect of training with the following loss
\begin{align}
    L_\psi(\hat y, y) = \sum_i (\psi(\hat y_i) - \psi(y_i))^2\, 
    \label{eq:supptonemaploss}
\end{align}
while converging to an unbiased result. This can be accomplished by using a locally valid linear approximation for the error term:
\begin{align}
    \psi(\hat y_i) - \psi(y_i) &\approx \psi(\hat y_i) - (\psi(\hat y_i) + \psi'(\hat y_i) (y_i - \hat y_i) \nonumber \\
    &= \psi'(\hat y_i) (\hat y_i - y_i) \, .
    \label{eq:linearization}
\end{align}
Note that we choose to linearize around $\hat y_i$ because, unlike the noisy observation $y_i$, $\hat y_i$ tends towards the true signal value $x_i = \operatorname{E}[y_i]$ over the course of training.

If we use a weighted L2 loss, then as we train the network we will have $\hat y_i \to \operatorname{E}[y_i] = x_i$ in expectation (where $x_i$ is the true signal value). This means that the terms summed in our gradient-weighted loss
\begin{align}
   \tilde L_{\psi}(\hat y, y) = &\sum_i \left[\psi'(\sg(\hat y_i))(\hat y_i - y_i) \right]^2 
   \label{eq:suppourloss}
\end{align}
will tend towards  $\psi'(x_i)(\hat y_i - y_i)$ over the course of training. Additionally, we note that the gradient of our reweighted loss~\ref{eq:suppourloss} is a linear approximation of the gradient of the tonemapped loss~\ref{eq:supptonemaploss}:
\begin{align}
    \nabla_\theta L_\psi(\hat y, y) 
    &= \sum_i \nabla_\theta (\psi(\hat y_i) - \psi(y_i))^2 \\
    &= \sum_i 2(\psi(\hat y_i) - \psi(y_i))\psi'(\hat y_i) \nabla_\theta y_i \\
    &\approx \sum_i 2(\psi'(\hat y_i) (\hat y_i - y_i))\psi'(\hat y_i) \nabla_\theta y_i \label{eq:sublinear}\\
    &= \sum_i 2(\psi'(\sg(\hat y_i)) (\hat y_i - y_i)) \psi'(\sg(\hat y_i)) \nabla_\theta y_i \label{eq:subsg} \\
    &= \nabla_\theta \tilde L_{\psi}(\hat y, y) \, .
\end{align}
In line~\ref{eq:sublinear} we substitute the linearization from~\ref{eq:linearization}, and in line~\ref{eq:subsg} we exploit the fact that a stop-gradient has no effect for expressions that will not be further differentiated.

\subsection{Weight variance regularizer}

Our weight variance regularizer is a function of the compositing weights used to calculate the final color for each ray. Given MLP outputs $c_i, \sigma_i$ for respective ray segments $[t_{i-1}, t_i)$ with lengths $\Delta_i$ (see~\cite{barron2021}), these weights are 
\begin{align}
    w_i &= (1-\exp(-\Delta_i \sigma_i)) \exp\!\left(-\sum_{j < i} \Delta_j \sigma_j\right) \, .
\end{align}
If we define a piecewise-constant probability distribution $p_w$ over the ray segments using these weights, then our variance regularizer is equal to 
\begin{align}
    \mathcal L_w = \operatorname{Var}_{X \sim p_w}(X) = \operatorname{E}_{X\sim p_W}\left[(X - \operatorname{E}[X])^2 \right]
\end{align}
Calculating the mean (expected depth):
\begin{align}
    E_{X\sim p_W}[X] &= \sum_i \int_{t_{i-1}}^{t_i} \frac{w_i}{\Delta_i} t \, dt \\
    &= \sum_i \frac{w_i}{\Delta_i} \frac{t_i^2 - t_{i-1}^2}{2} \\
    &= \sum_i w_i \frac{t_i + t_{i-1}}{2} \, .
\end{align}
We will denote this value as $\overline{t}$. Calculating the regularizer:
\begin{align}
    \operatorname{Var}_{X \sim p_w}(X) &= \operatorname{E}_{X\sim p_W}\left[(X - \operatorname{E}[X])^2 \right] \\
    &= \sum_i \int_{t_{i-1}}^{t_i} \frac{w_i}{\Delta_i} \left(t-\overline t \right)^2 \, dt \\
    &= \sum_i \frac{w_i}{\Delta_i} \frac{\left(t_i-\overline t \right)^3 - \left( t_{i-1}-\overline t \right)^3}{3} \\
    = \sum_i  w_i& \frac{(t_i-\overline t)^2 + \left( t_i-\overline t \right) \left(t_{i-1}-\overline t \right) + \left( t_{i-1}-\overline t \right)^2}{3} 
\end{align}
We apply a weight between $1 \times 10^{-2}$ and $1 \times 10^{-1}$ to $\mathcal L_w$ (relative to the rendering loss), typically using higher weights in noisier or darker scenes that are more prone to ``floater'' artifacts. Applying this regularizer with a high weight can result in a minor loss of sharpness, which can be ameliorated by annealing its weight from 0 to 1 over the course of training.

\subsection{Findings with alternate loss functions}

In practice, we directly scale our loss by the derivative of the desired tone curve:
\begin{equation}
   \psi'(\sg(\hat y_i)) = \frac{1}{\sg(\hat y_i) + \epsilon}
\end{equation}
We performed a hyperparameter sweep over loss weightings of the form $(\sg(\hat y_i) + \epsilon)^{-p}$ for $\epsilon$ and $p$ and found that $\epsilon=1\times 10^{-3}$ and $p=1$ produced the best qualitative results. 

We also experimented with using a reweighted L1 loss or the negative log-likelihood function of the actual camera noise model (using shot/read noise parameters from the EXIF data) but found that this performed worse than reweighted L2. RawNeRF models supervised with a standard unweighted L2 or L1 loss tended to diverge early in training, particularly in very noisy scenes.

We tried using the unclipped sRGB gamma curve (extended as a linear function below zero and as an exponential function above 1) in our loss, but found that it caused many color artifacts in dark regions. Directly applying our log tone curve (rather than reweighting by its gradient) before the L2 loss caused training to diverge.

\subsection{Quality limitations}

As briefly mentioned in the main text, our method cannot scale to arbitrary amounts of noise in real world scenes. For our darkest nighttime scenes, we often must run COLMAP~\cite{colmap} multiple times (varying the random seed) or tune its parameters to obtain camera poses. Even when COLMAP reports a successful reconstruction, the results are sometimes poorly aligned at image corners, where the distortion model used for camera intrinsics may not fit well. 

RawNeRF itself is prone to reconstruction artifacts in very noisy scenes or scenes captured with few images (under 30), typically in the form of positional encoding grid-like artifacts. These artifacts are often more evident in videos than in still frames. In regions that are essentially pure noise and no signal, RawNeRF sometimes produces a foggy ``cloud'', since no multiview information exists to guide its recovery of geometry.

The near and far plane bounds calculated using the point cloud from COLMAP are sometimes wider than the true bounds of the scene. Using these bounds wastes many samples at the front of each ray, which reduces sharpness and can cause additional ``floater'' artifacts. We therefore sometimes retrain RawNeRF models using tighter depth bounds than those reported by COLMAP.

We found it necessary to use gradient clipping due to the high level of noise in the data we use for supervision. Certain losses (such as standard L2) are prone to producing NaN gradient values and require careful tuning of the clipping values. We found our reweighted loss to be more stable.

\section{Data capture and postprocessing details}

\subsection{Data capture}

We captured all images using a 2017 iPhone X with the Halide app\footnote{\url{https://halide.cam/}} and a 2020 iPhone SE with the Adobe Lightroom app. We used manual modes in both apps with focus and ISO level fixed for each capture, manually adjusting shutter speed to achieve an exposure with no clipped highlights (except in scenes with varying exposure) and minimal motion blur (at least $1/100$s when possible). At night, it was usually necessary to use the maximum ISO level (approximately 2000 on the iPhones) to achieve minimal motion blur. Each capture took around 10-200 seconds, except for the denoising test scenes. All raw images are stored as Adobe DNG\footnote{\url{https://www.adobe.com/content/dam/acom/en/products/photoshop/pdfs/dng_spec_1.4.0.0.pdf}} files.

We extract the following parameters from the EXIF metadata using \texttt{exiftool}:

\begin{table}[h]
    \centering
    \begin{tabular}{llc}
        Variable & EXIF field name & \# values \\ \hline
        $w$& \texttt{WhiteLevel} & $1$ \\
        $b$& \texttt{BlackLevel} & $1$ \\
        $g_\mathrm{wb}$& \texttt{AsShotNeutral}  & $3$ \\
        $C_\mathrm{ccm}$ & \texttt{ColorMatrix2} & $3\times 3$ \\
        $t$ &  \texttt{ShutterSpeed} & $1$  \\
    \end{tabular}
    \label{tab:exif}
\end{table}

The color correction matrix $C_\mathrm{ccm}$ is an XYZ-to-camera-RGB transform under the D65 illuminant, so we use the corresponding RGB-to-XYZ matrix\footnote{\url{http://www.brucelindbloom.com/index.html?Eqn_RGB_XYZ_Matrix.html}}:
\begin{equation}
    \!\!\!C_{\textrm{rgb-xyz}} = \left[ \begin{array}{ccc}
0.4124564  & 0.3575761  & 0.1804375\\
0.2126729  & 0.7151522  & 0.0721750\\
0.0193339  & 0.1191920  & 0.9503041 
    \end{array} \right]
\end{equation}
We use these to create a single color transform $C_\mathrm{all}$ mapping from camera RGB directly to standard linear RGB space:
\begin{equation}
    C_\mathrm{all} = \mathrm{rownorm}((C_{\textrm{rgb-xyz}}C_\mathrm{ccm})^{-1})
\end{equation}
where $\mathrm{rownorm}$ normalizes each to sum to 1.

We use the standard sRGB gamma curve as a basic tonemap for linear RGB space data:
\begin{equation}
    \gamma_\mathrm{sRGB}(z) = \begin{cases} 
      12.92z & z \leq 0.0031308 \\
      1.055z^{1/2.4}-0.055 & z > 0.0031308
   \end{cases}
\end{equation}

\subsection{Postprocessing pipeline}
\label{sec:post}

Our exact postprocessing pipeline for converting raw images to postprocessed sRGB space is detailed below.
\begin{enumerate}
    \item Load 12-bit raw data using \texttt{rawpy}.
    \item Cast to 32-bit floating point.
    \item Rescale so that the black level is 0 and the white level is 1, preserving values below zero. (The result here is used to train RawNeRF.)
    \begin{equation}
        z \leftarrow \frac{z - b}{w - b}
    \end{equation} 
    \item Apply bilinear demosaicking (when necessary).
    \item Apply elementwise white balance gains.
    \begin{equation}
        z \leftarrow \frac{z}{g_{\mathrm{wb}}}
    \end{equation}
    \item Apply a color correction matrix (from camera RGB to canonical XYZ) and XYZ-to-RGB matrix, combined into a $3\times 3$ transformation.
    \begin{equation}
        z \leftarrow C_{\mathrm{all}} z
    \end{equation}
    \item Adjust the exposure to set the white level to the $p$-th percentile ($p=97$ by default).
    \begin{equation}
        z \leftarrow \frac{z}{\mathrm{percentile}(z, p)}
    \end{equation}
    \item Clip to $[0,1]$.
    \begin{equation}
        z \leftarrow \mathrm{clip}(z,0,1)
    \end{equation}
    \item Apply the sRGB gamma curve to each color channel.
    \begin{align}
        z &\leftarrow \gamma_\mathrm{sRGB}(z)
    \end{align}
\end{enumerate}
When applying a different tonemapping algorithm, we take the color corrected output from step 6 and pass it through the alternate method, while tuning exposure and other tonemapping parameters manually per scene.

\subsection{Camera shutter speed miscalibration}
\label{sec:miscalibration}

In Section 4.2 of the main text, we discuss our implementation of a learned per-color-channel scaling to account for miscalibration when using variable exposure inputs. Here, we document this miscalibration effect for completeness. 

\begin{figure}
    \centering
    \includegraphics[height=2.45cm]{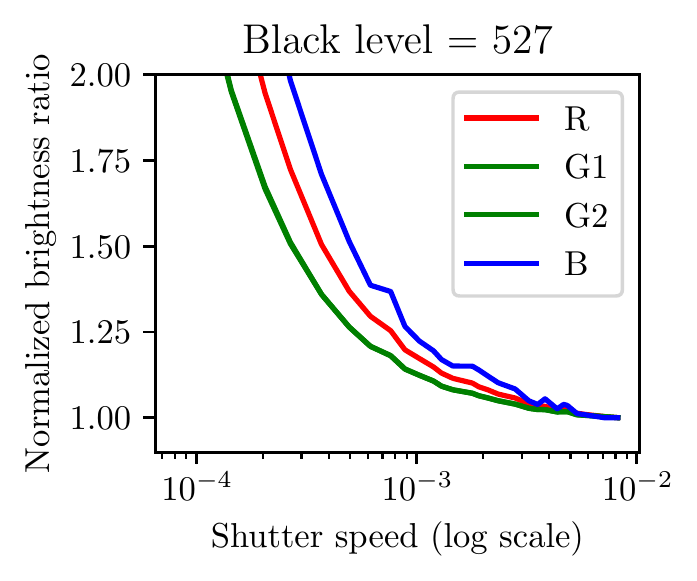}\hspace{-.2cm}
    \includegraphics[height=2.45cm]{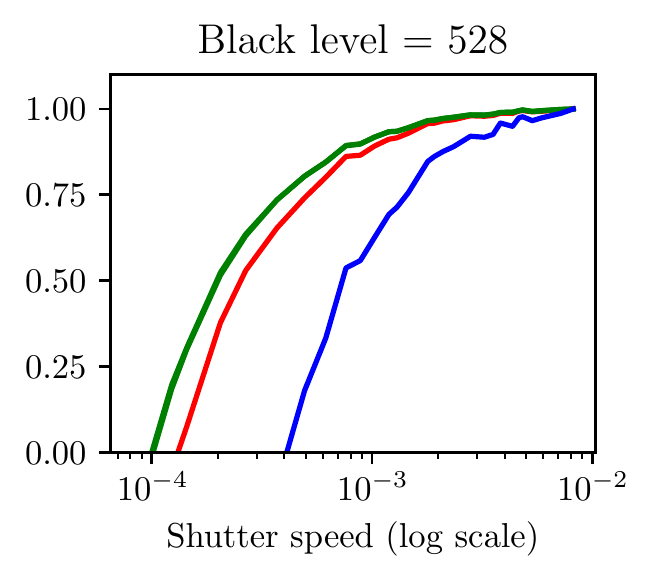}\hspace{-.2cm}
    \includegraphics[height=2.45cm]{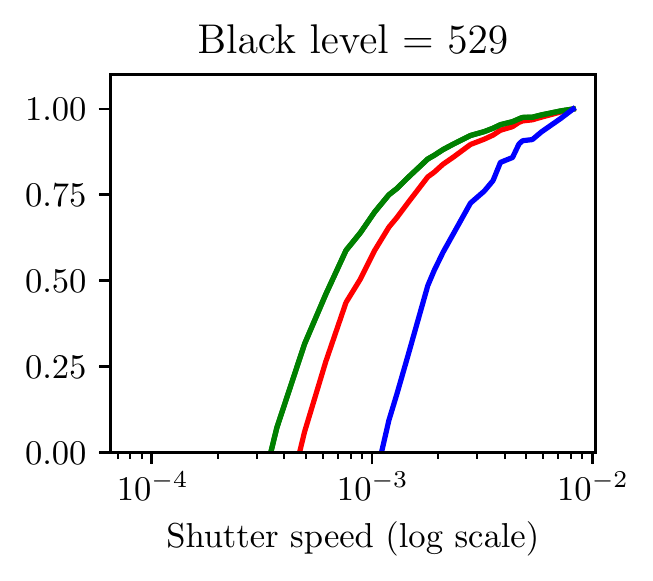}
    \caption{Camera shutter speed miscalibration. We plot normalized brightness for each Bayer color channel, relative to its value at the longest shutter speed. For a perfectly calibrated sensor, these lines would all be at a constant height of 1. We show plots using both the true black level (528) and surrounding values.}
    \label{fig:miscalibration}
\end{figure}

Figure~\ref{fig:miscalibration} plots data taken from a ``sweep'' over many shutter speeds. The 2017 iPhone X (used for most data capture in the paper) is held fixed on a tripod, all other parameters (focus, ISO, white balance, etc.) are held fixed, and shutter speeds are sampled roughly logarithmically from 1/100 to 1/10000 seconds. We ensure that no pixels are saturated. To minimize the effect of image noise, we study the average color value $y_{t_i}^c$ for each Bayer filter channel (R, G1, G2, B) over the entire 12MP sensor. Specifically, we plot:
\begin{equation}
    \frac{y_{t_i}^c}{t_i}  \cdot   \frac{t_{\mathrm{max}}}{y_{t_\mathrm{max}}^c} 
\end{equation}
which is the ratio of normalized brightness at speed $t_i$ to normalized brightness at the longest shutter speed $t_{\mathrm{max}}$. In the case of perfect calibration, this should be equal to 1 everywhere since dividing out by shutter speed should perfectly normalize the brightness value. However, from Figure~\ref{fig:miscalibration} we see that not only does this quantity decay for faster shutter speeds, it decays at \emph{different rates} per color channel. To preempt concerns that this problem is due to black level miscalibration, we include the plot based on the correct black level 528, as well as the surrounding values, which shows that this problem is only worsened by shifting the black level higher or lower. Note that black level is an integer on the scale of 0 to 4095 (since this is a 12-bit sensor).

\begin{figure}
    \centering
\resizebox{\linewidth}{!}{
    \begin{tabular}{@{}c@{\,}c@{\,}c@{}}
    \includegraphics[height=2.4cm]{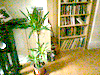} & 
    \includegraphics[height=2.4cm]{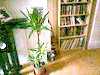} &
    \includegraphics[height=2.4cm]{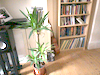} \\
    (a) Fast shutter & (b) Fast shutter, corrected & (c) Slow shutter
    \end{tabular}
    }
    \caption{(a) Fast and (c) slow captures of the \emph{testyucca} scene, with brightness normalized by shutter speed (heavily downscaled to minimize noise). These two images should match perfectly, but have a perceptible color difference due to the miscalibration documented in Section~\ref{sec:miscalibration} and Figure~\ref{fig:miscalibration}. (b) In the center, we show a version of (a) with per-channel rescaling in the raw domain to match the global color balance of (c).}
    \label{fig:yucca_wb}
\end{figure}

We show an example of the resulting qualitative color shift in Figure~\ref{fig:yucca_wb} using images from one of our three real test scenes. Here the two shutter speeds are 1/1104 and 1/181 seconds, and the relative color shift from the slow to the fast channel is calculated to be $(0.89, 0.93, 0.75)$ for red, green, and blue in the raw domain. The effect of undoing this shift before postprocessing is shown in Figure~\ref{fig:yucca_wb}b. This miscalibration is another reason for primarily reporting affine-aligned metrics on our real test set, since we cannot rely on perfect color alignment between the input noisy image and the clean ground truth frame.

We do not fully understand the cause of this issue. We speculate that it could be due to the sensor temperature changing over the course of capture, imprecise shutter speed timing for very fast exposures, or any number of other factors related to low level sensor hardware. Given that the effect exists and affects our captures in an unmeasurable manner, it must be accounted for. Using a DSLR or mirrorless camera with a better sensor may avoid this issue.

\section{Comparison and ablation details}

\subsection{Real test dataset details}

\paragraph{Affine alignment} As mentioned in the main text, we solve for an affine color alignment between each output and the ground truth clean image. For all methods but SID and LDR NeRF, this is done directly in raw Bayer space for each RGGB plane separately. For SID and LDR NeRF (which output images in tonemapped sRGB space), this is done for each RGB plane against the tonemapped sRGB clean image. If the ground truth channel is $x$ and the channel to be matched is $y$, we specifically compute 
\begin{align}
    a &= \frac{\overline{xy} - \overline x \overline y}{\overline{x^2} - \overline x^2} = \frac{\mathrm{Cov}(x, y)}{\mathrm{Var}(x)} \, , \\
    b &= \overline y - a \overline x
\end{align}
to get the least-squares fit of an affine transform $ax+b \approx y$ (here $\overline z$ indicates the mean over all elements of $z$). We then apply the inverse transform as $(y-b)/a$ to match the estimated $y$ to $x$. In the case where matching happens in the raw domain, we postprocess $(y-b)/a$ through our standard pipeline (Section~\ref{sec:post}) before calculating sRGB-space metrics.

\paragraph{Compared baselines} We provide an overview of each baseline and the pre- and post-processing pipelines used in the main text. Unprocessing~\cite{brooks2019cvpr} is the only method that is a ``non-blind'' denoiser, and therefore requires a per-pixel noise level as input. We calculate this by using the empirical per-pixel variance from our tripod-aligned fast and clean images to estimate shot and read noise parameters as a best-fit 1D affine transform mapping from clean signal values to empirical variances. Each method required its own relative input rescaling and clipping convention, which we set based on each authors' source code.

\subsection{Synthetic Lego dataset details}

In the synthetic Lego dataset, we did \emph{not} include the effects of remosaicking/demosaicking or quantization when unprocessing/reprocessing the data. We wanted the ``infinite'' shutter speed case to be perfectly clean, with no degradation resulting from unprocessing and reprocessing in the absence of noise, thus providing an upper bound on possible performance. This example does not particularly test the ability of RawNeRF to encode high dynamic range since the object is diffusely lit, resulting in fairly dim highlights and negligible clipping; instead, it focuses on robustness to noise.

We rendered new randomly sampled images of the scene using the Blender file\footnote{\url{https://drive.google.com/file/d/1RjwxZCUoPlUgEWIUiuCmMmG0AhuV8A2Q/view?usp=sharing}} provided by the NeRF authors~\cite{mildenhall2020nerf}, saving the resulting linear space color data in EXR format. There are 120 images in the training set and 40 images in the test set. Note that metric values on this data are not comparable to metrics on the original scene, since it uses images from different random poses generated using a different postprocessing pipeline.

For completeness, we report the unmasked PSNR values for this experiment in Table~\ref{tab:synthlegounmasked} (Table 2 in the main text reports masked PSNR), which is heavily skewed by the LDR NeRF's color bias in the black background regions.

\begin{table}[]
    \centering
\resizebox{\linewidth}{!}{
\begin{tabular}{@{}l|ccccccc@{}}
\multicolumn{1}{c|}{} & \multicolumn{7}{c}{Simulated shutter speed (seconds)} \\
Method & $\infty$ &  $1/7   $ & $1/15  $ & $1/30  $ & $1/60  $ & $1/120 $ & $1/240 $ \\ \hline
Noisy input  &   -   & 20.16 & 16.90 & 13.81 & 10.83 &  8.06 &  5.95 \\
LDR NeRF     & \textbf{38.06} & 24.66 & 21.39 & 18.27 & 15.31 & 12.47 & 10.13 \\
RawNeRF      & 36.85 & \textbf{36.82} & \textbf{36.65} & \textbf{36.27} & \textbf{35.62} & \textbf{34.33} & \textbf{32.37} \\
    \end{tabular}
    }
    
    \caption{
    Unmasked LDR sRGB PSNRs for the ablation study on our synthetic \emph{Lego} scene data.
    }
    \label{tab:synthlegounmasked}
\end{table}

\section{Further qualitative ablations}

\subsection{Training with iPhone JPEG inputs}

\begin{figure}
    \centering
\resizebox{\linewidth}{!}{
    \begin{tabular}{@{}c@{\,}c@{\,}c@{}}
    \includegraphics[height=2.4cm]{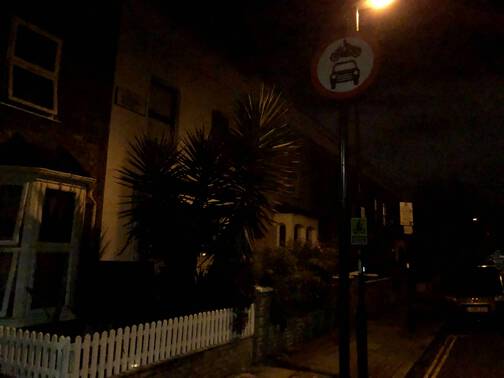} & 
    \includegraphics[height=2.4cm]{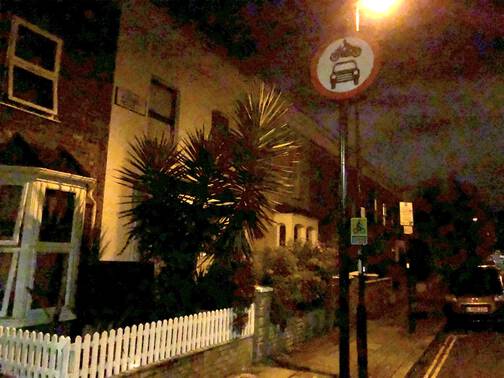} &
    \includegraphics[height=2.4cm]{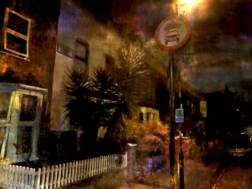} \\
    (a) iPhone JPEG & (b) Brightened (a) & (c) LDR NeRF on (a) \\
    \includegraphics[height=2.4cm]{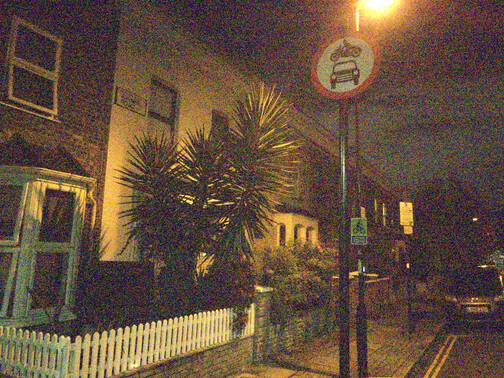} & 
    \includegraphics[height=2.4cm]{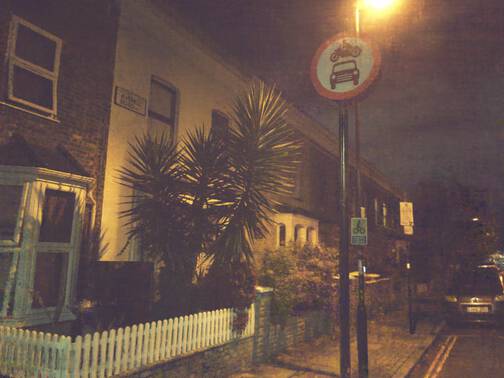} &
    \includegraphics[height=2.4cm]{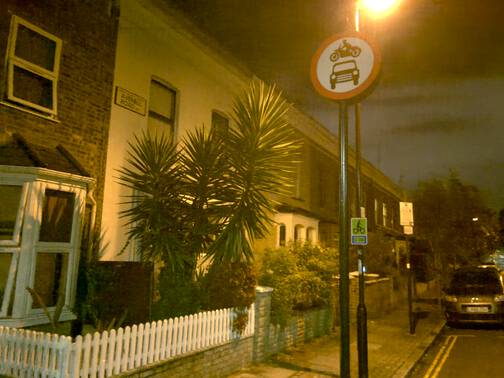} \\
    (d) Our LDR input & (e) LDR NeRF on (d) & (f) RawNeRF
    \end{tabular}
    }
    \caption{Comparison of training LDR NeRF using sRGB images either directly from the iPhone camera or from our simplified pipeline. (a) The JPEG image from the phone is extremely dark, so we brighten it for visualization (b). We also brighten the resulting LDR NeRF rendering (c), thereby revealing its pervasive color noise artifacts. When trained on the images from our LDR processing pipeline (d), LDR NeRF produces a more reasonable result (e), though the input images' biased noise distribution still results in muddy, low contrast dark regions and incorrectly muted colors. (f) Only RawNeRF accurately recovers the correct colors and details throughout the scene.}
    \label{fig:iphonejpeg}
\end{figure}

In all LDR NeRF comparisons in the main paper, we use our own simple postprocessing pipeline to generate LDR sRGB inputs from the raw data. However, a standard NeRF implementation would instead use JPEG images directly from the camera, which have a more sophisticated postprocessing pipeline that likely includes noise reduction and a more sophisticated nonlinear tonemap to better compress dynamic range. To satisfy the reader's potential curiosity, in Figure~\ref{fig:iphonejpeg} we provide an example of LDR NeRF trained on iPhone JPEGs versus our LDR images, as well as a RawNeRF result on the same scene.

\subsection{Bayer mosaic mask and sensor artifacts}

\begin{figure}
    \centering
\resizebox{\linewidth}{!}{
    \begin{tabular}{@{}c@{\,}c@{\,}c@{\,}c@{}}
    \includegraphics[height=4cm]{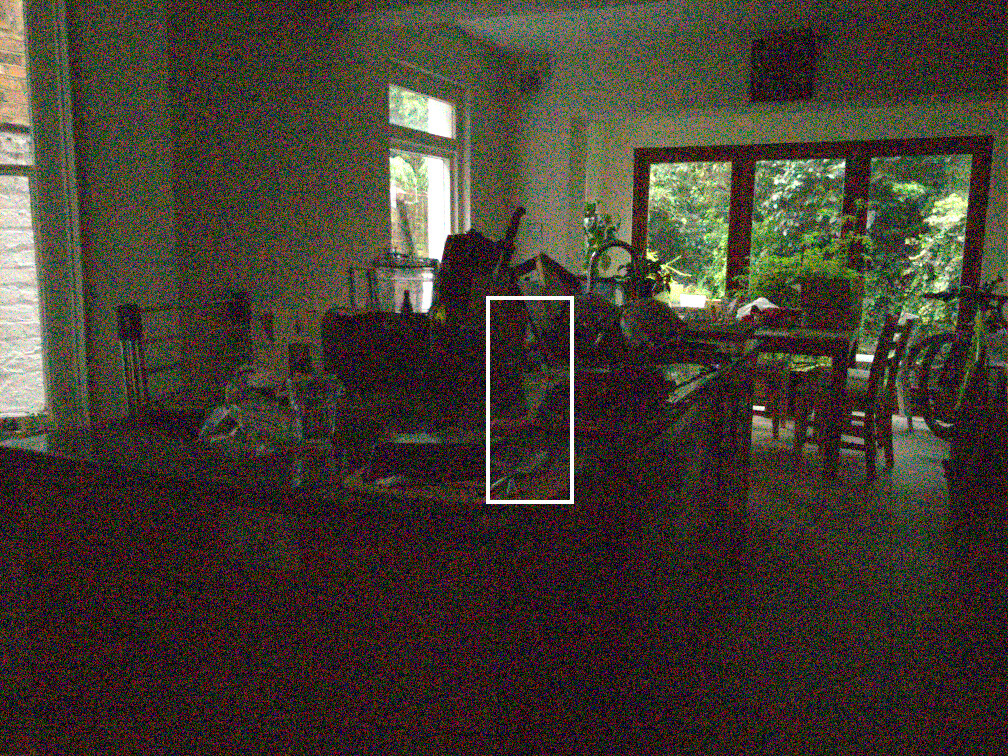} & 
    \includegraphics[height=4cm]{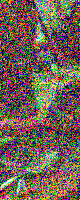} & 
    \includegraphics[height=4cm]{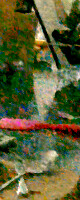} & 
    \includegraphics[height=4cm]{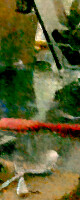} \\
    (a) Full noisy image & (b) Noisy & (c) No mask & (d) Ours
    \end{tabular}
    }
    \caption{Comparison of training with bilinear demosaicking and no Bayer masking (c), or with a Bayer mask that uses only the measured raw pixels (d). In image areas with extremely high noise, we observed unpleasant bright color noise emerge when training with bilinear demosaicked images in the raw domain.}
    \label{fig:demosaic}
\end{figure}

In the main text, we note that we only apply our loss function to the color channel measured by the Bayer filter for each ray. (In practice, we render all three colors for every training ray, then apply a one-hot mask to select the desired output color.) In Figure~\ref{fig:demosaic}, we show an example of the color noise that emerges when supervising all 3 color channels using bilinearly demosaicked raw images instead of masking the loss. Perhaps surprisingly, we noted that relatively clean regions of the scene seemed to benefit from using all 3 channels of a bilinear demosaicked image as supervision. However, we concluded that the distracting color artifacts induced by demosaicking outweighed this occasional benefit, and opted to use Bayer masking in all scenes. 

These artifacts may potentially be caused by broken ``hot'' pixels that are always fully saturated, in violation of our assumed noise distribution. Bilinear demosaicking would disperse the influence of a hot pixel to many neighboring pixels, potentially increasing its effect on the final trained NeRF. In preliminary experiments, we did not notice any benefit to additionally masking hot pixels when applying a Bayer mask. We did apply a second mask to remove a 4 pixel border from all training images, since many iPhone raw images contained 1 or 2 entire rows or columns of saturated pixels on one side, particularly in bright scenes.

\section{Synthetic defocus rendering model}

To render defocused images, we use a similar rendering model as prior work that has addressed this task~\cite{barron2015stereo,wadhwa2018defocus,zhang2019defocus}. To avoid prohibitively expensive rendering speeds, we first precompute a multiplane image~\cite{zhou18stereomag} representation from the trained RawNeRF model. This MPI consists of a series of fronto-parallel RGBA planes (with colors still in linear HDR space), sampled linearly in disparity within a camera frustum at a central camera pose. Given this MPI representation, our rendering algorithm for synthetic defocus (including lateral camera translation) is described in Algorithm~\ref{alg:defocus}.

\begin{algorithm}
\caption{Synthetic defocus rendering}\label{alg:defocus}
\begin{algorithmic}
\Procedure{Defocus}{$c_{\textrm{mpi}}, \alpha_{\textrm{mpi}}, i_\textrm{focus}, \Delta_r, \Delta_d$}
\State $C \gets 0$ %
\For{$i=0,\ldots,,N-1$}
    \State $r \gets \Delta_r \cdot |i - i_\textrm{focus}|$
    \State $k_\textrm{blur} \gets  \mathrm{blurkernel}(r)$
    \State $c_\textrm{blur} \gets \mathrm{convolve}(c_{\textrm{mpi}}^{(i)} \cdot \alpha_{\textrm{mpi}}^{(i)}, k_\textrm{blur})$
    \State $\alpha_\textrm{blur} \gets \mathrm{convolve}(\alpha_{\textrm{mpi}}^{(i)}, k_\textrm{blur})$
    \State $d \gets \Delta_d \cdot i$
    \State $c_\textrm{trans} \gets \mathrm{translate}(c_{\textrm{blur}}, d)$
    \State $\alpha_\textrm{trans} \gets \mathrm{translate}(\alpha_{\textrm{blur}}, d)$
    \State $C \gets c_\textrm{trans} + (1-\alpha_\textrm{trans}) C$
\EndFor
\State \Return $C$
\EndProcedure
\end{algorithmic}
\end{algorithm}
Here the input MPI planes are indexed from back to front. $i_\textrm{focus}$ controls the focal plane, $\Delta_r$ controls the simulated aperture size (defocus strength), and $\Delta_d$ (a 2D vector) controls the camera translation parallel to the image plane. $\mathrm{blurkernel}(r)$ returns a circular mask at the origin with radius $r$ pixels. $\mathrm{blurkernel}$ is implemented as a 2D Fourier space convolution, and $\mathrm{translate}$ is a continuous 2D image translation (using bilinear resampling). Note that the color is ``premultiplied'' by alpha before blurring, which is why alpha is not applied to $c_\textrm{trans}$ in the accumulation step for $C$.

\section{Scene index}

We provide various details about each scene shown in the paper and video in Table~\ref{tab:sceneindex}.

\begin{table*}[]
    \centering
    \begin{tabular}{r|l|cccccc}
    & Scene & Figures & Video & Images & Shutter speed ($\text{s}^{-1}$) & ISO & Time of day \\ \hline
 \parbox[t]{3mm}{\multirow{5}{*}{\rotatebox[origin=c]{90}{Main text}}}
 & \textit{candle}           &  1        &  0:00, 1:45     &  173  &               45, 119  &   2000  &  20:21   \\
 & \textit{livingroom}       &  2        &       &   50  &                  1429  &    800  &  15:14   \\
 & \textit{stove}            &  4        &  3:33     &  106  &        139, 258, 1621  &   2000  &  20:17   \\
 & \textit{windowlegovary}   &  5, 8d    &  4:28     &  104  &     432, 16129, 16393  &    500  &  10:29   \\
 & \textit{gardenlights}     &  8a-c     &  5:39     &   91  &                    50  &   1600  &  23:53   \\ \hline
 \parbox[t]{3mm}{\multirow{3}{*}{\rotatebox[origin=c]{90}{Test set}}} 
 & \textit{pianotest}        &  6, 8e    &  5:25     &  103  &              145, 207  &   2000  &  22:08   \\
 & \textit{officetest}       &  6        &       &  113  &              110, 249  &   2000  &  17:43   \\
 & \textit{yuccatest}        &  6, A4    &       &  102  &             181, 1104  &    800  &  13:09   \\ \hline
 \parbox[t]{3mm}{\multirow{14}{*}{\rotatebox[origin=c]{90}{Supplement and video}}} 
 & \textit{streetcorner}     &  A1a, A5  &  2:35     &   57  &                   123  &   2000  &  22:19   \\
 & \textit{candlefiat}       &  A1b      &  4:15     &   52  &                    97  &   2000  &  00:33   \\
 & \textit{nightstreet}      &  A1c      &       &   49  &                    82  &   2000  &  23:04   \\
 & \textit{parkstatue}       &  A1d      &  5:13     &   51  &                   124  &   2000  &  23:14   \\
 & \textit{bikes}            &  A1e      &  3:21     &   45  &                    62  &   2000  &  22:22   \\
 & \textit{twostatue}        &  A2a      &  4:52     &   86  &                   239  &     20  &  11:32   \\
 & \textit{choir}            &  A2b      &  4:03     &   28  &                   112  &     50  &  11:50   \\
 & \textit{stainedglass}     &  A2c      &  5:02     &   43  &                   155  &     32  &  11:46   \\
 & \textit{onestatue}        &  A2d      &  4:42     &   40  &                   112  &     25  &  11:51   \\
 & \textit{sharpshadow}      &  A2e      &       &   36  &                  8130  &     32  &  13:35   \\
 & \textit{morningkitchen}   &  A2f, A6  &       &   53  &                   110  &   2000  &  08:18   \\
 & \textit{scooter}          &           &  3:08     &   54  &                   107  &   2000  &  19:27   \\
 & \textit{notchbush}        &           &  3:44     &   63  &                    95  &   2000  &  22:15   
    \end{tabular}
    \caption{A summary of image metadata for our scenes. Figure from the supplement are indicated using the prefix ``A''.}
    \label{tab:sceneindex}
\end{table*}

\end{document}